\documentclass[preprint,12pt]{elsarticle}
\usepackage{hyperref}
\hypersetup{hypertex=true,
            colorlinks=true,
            linkcolor=blue,
            anchorcolor=blue,
            citecolor=blue}
\usepackage{amsthm}
\usepackage{mathrsfs}
\usepackage{subfigure}
\usepackage{amsmath}
\usepackage{esint}
\usepackage{float}
\usepackage{multirow}
\usepackage{color}
\usepackage{amssymb}
\usepackage[subfigure]{tocloft}
\usepackage{booktabs}
\usepackage{geometry}
\usepackage{graphicx} 
\usepackage{epstopdf}
\newtheorem{lemma}{Lemma}[section]
\newtheorem{definition}{Definition}%

\newtheorem{example}{Example}[section]

\begin{document}

\begin{frontmatter}




\title{Classification by estimating the cumulative distribution function for small data}




\author[1]{Meng-Xian Zhu }
\address[1]{School of Economics, Hainan University, Haikou, 570228, P.R.China}

\author[2]{Yuan-Hai Shao\corref{cor1}}
\address[2]{School of Management, Hainan University, Haikou, 570228, P.R.China} \cortext[cor1]{Corresponding author.} \ead{shaoyuanhai21@163.com}

\begin{abstract}
In this paper, we study the classification problem by estimating the conditional probability function of the given data. Different from the traditional expected risk estimation theory on empirical data, we calculate the probability via Fredholm equation, this leads to estimate the distribution of the data. Based on the Fredholm equation, a new expected risk estimation theory by estimating the cumulative distribution function is presented. The main characteristics of the new expected risk estimation is to measure the risk on the distribution of the input space. The corresponding empirical risk estimation is also presented, and an $\varepsilon$-insensitive $L_{1}$ cumulative support vector machines ($\varepsilon$-$L_{1}$VSVM) is proposed by introducing an insensitive loss. It is worth mentioning that the classification models and the classification evaluation indicators based on the new mechanism are different from the traditional one. Experimental results show the effectiveness of the proposed $\varepsilon$-$L_{1}$VSVM and the corresponding cumulative distribution function indicator on validity and interpretability of small data classification.
\end{abstract}

\begin{keyword}
Classification, cumulative distribution function, Fredholm equation, expected risk estimation, support vector machines.
\end{keyword}

\end{frontmatter}

\section{Introduction}\label{sec1}
Classification \citep{jain2000statistical,vapnik2000svm} is also known as pattern recognition. It is one of the main research problems of machine learning.  For binary classification, generally, we are given a set of training samples
$S=\{(x_{i},y_{i}),i=1,2,...,m\}$,
where $x_{i}\in R^{d}$ is the input and $y_{i}\in\{0,1\}$ is the
corresponding output with $i=1,\ldots,m$. Suppose the given data is generate from an independent and identically distributed ($i.i.d.$) distribution, our task is to deduce the output $y$ of any input $x$ from the given data set. In fact, the classification problem corresponds to the problem of conditional probability $P(y\vert x)$ estimation in statistical inference \citep{vapnik1999nature,vapnik2015v}, and it has been studied extensively in both machine learning and statistical inference \citep{theodoridis2006pattern,shao2011improvements,fukunaga2013introduction}.

A common way to estimate the conditional probability in classification is to suppose a function $f(x)\approx P(y\vert x)$. For the binary classification problem, since $P(y=1\vert x)+P(y=0\vert x)=1$, without loss of generality, we describe the problem of estimating the conditional conditional probability as estimating $P(y=1\vert x)$ from now on.
One of the most theory for classification is the expected risk minimization theory \citep{vapnik1991principles}, and the core idea of expected risk minimization theory is to minimize the error of estimation function $f(x)$ and real output in the expected space. The advantage of expected risk minimization theory is that it does not need to have too many assumptions about the estimation function. Most discriminant classifiers are based on the expected risk minimization theory, such as least squares classifiers (LSC) \citep{suykens1999least}, neural networks (NN) \citep{bishop1995neural}, support vector machines (SVM) \citep{cortes1995support,deng2012support}, ensemble learning (EL) \citep{zhou2012ensemble} et.al. In fact, when estimating the expected risk, it is necessary to estimate the distribution structure $F(x,y)$ of data. For simplicity of calculation, most of the models assumes that the data are uniformly distributed, thus there is no operation to estimate the data distribution.

Recently, Vapnik and Izmailov \citep{vapnik2019rethinking} further studied the classification problem by estimating the cumulative distribution function and conditional probability. And proposed a new way to estimate the conditional probability $P(y=1\vert x)$ by solving the integral equation based on the definition of conditional probability and the density function directly.
To solve the integral equation, \cite{vapnik2019rethinking, vapnik2020complete} presented a form of least squares implementation called VSVM
and compares it with the traditional least squares and support vector machine classifiers. The advantages of using the integral equation is that $f(x)$ can be solved by estimating the distribution function in input space, and the integral equation has wider convergence properties. Therefore, this method has attracted the attention of different machine learning fields \citep{cherkassky2019group,mottaghitalab2021prediction,lu2021nonlinear,zhang2022approach}.
However, the general paradigm of using integral equation inference $f(x)$ is not discussed, and the corresponding expected risk minimization theory and empirical estimation theory are not clear yet.

In this paper, different from the traditional expected risk estimation theory on empirical data, our paper studies to give the original theoretical model of the direct estimation of conditional probability from the framework of expectation risk via Fredholm equation, and presents the corresponding empirical model based on the
cumulative distribution function estimation.
One $\varepsilon$-insensitive empirical risk estimation is established and an $\varepsilon$-insensitive $L_{1}$ loss cumulative support vector machines ($\varepsilon$-$L_{1}$VSVM) is presented. It is worth mentioning that the classification evaluation indicator via estimating Fredholm equation  are different from the traditional one. Experimental results show the effectiveness of the proposed $\varepsilon$-$L_{1}$VSVM and the corresponding cumulative distribution function indicator on validity and interpretability of classification, especially when the training data is insufficient.
The main contributions are summarized as follows:
\begin{itemize}
\item[i)] For estimating $P(y=1\vert x)$ by using the cumulative distribution function, we perform an objective transformation of the Fredholm integral equation to estimated conditional probabilities from a probabilistic perspective. The main characteristic of using Fredholm integral equation is the function to be evaluated and the real output are equal under the cumulative distribution function.
\item[ii)] For estimating $P(y=1\vert x)$ by using statistical estimation, the expected risk minimization and empirical risk minimization theory based on the Fredholm integral equation is presented. Due to the cumulative distribution function should be estimate, the loss function and the expected risk defined by Fredholm integral equation is different from the traditional one.
\item[iii)] Different from traditional 0-1 loss, an insensitive way of estimating the empirical risk using $\varepsilon$-insensitive loss is proposed, it has sparsity and with higher efficiency than least squares loss. And an $\varepsilon$-$L_{1}$VSVM is proposed based on $\varepsilon$-insensitive loss. A new evaluation criterion with distribution information is proposed, an out-of-sample estimation of the $v$ value of testing sample.
\item[iv)] Experimental results show the effectiveness of the proposed $\varepsilon$-$L_{1}$VSVM and the corresponding cumulative distribution function indicator on validity and interpretability of classification, especially when the training data is insufficient.
\end{itemize}

The rest of the paper is organized as follows. In section 2, we introduce the implementation path of classical model for classification and the brief overview of the method for direct estimation of conditional probabilities described in VSVM. In section 3, we give a theoretical framework for estimating conditional probabilities with empirical distribution function, which includes a rigorous derivation and transformation of the integral equation from a probabilistic point of view, and a theoretical study of estimating the equation in a statistical sense. In section 4, specific estimation models based on the proposed theoretical framework and an improved evaluation indicator is presented.
In section 5, we show experimental results on artificial datasets and a number of publicly available datasets.
In section 6, we give some concluding remarks.

\section{Preliminaries}\label{sec2}

\subsection{Risk minimization in classification}
In the following, we consider binary classification model of learning from training samples through three components \citep{vapnik1999nature}
\begin{enumerate}[1)]
\item A generator  of random vectors $x\in \mathcal{X}$, drawn independently from a fixed but unknown probability distribution function $F(x)$.
\item A supervisor  who returns a binary output value $y_i\in \mathcal{Y}:=\{0,1\}$ to every input vector $x_i$, according to a conditional probability function $P\left(y=1\vert x\right)$, where $P(y=0\vert x)=1-P(y=1\vert x)$, also fixed but unknown.
\item A learning machine capable of implementing a set of indicator functions (functions which take only two values: zero and one) $\mathcal{F}$.
\end{enumerate}
The problem of learning is that of choosing from the given set of indicator functions $\mathcal{F}$, the one that best approximates the supervisor's response.
The selection of the desired function is based on a training set of $m$ $i.i.d.$ observations
\begin{equation}\label{Data}
(x_1,y_1),(x_2,y_2)...,(x_m,y_m), x_i\in \mathcal{X}, y_i\in \mathcal{Y},
\end{equation}
drawn according to $F(x, y) = F(x)P(y\vert x)$.
To find the function $y=f(x)$ (the mapping $f: \mathcal{X}\to \mathcal{Y}$) based on a training set that satisfies the above basic assumptions, we need to have some measure of how good a function as a decision function. To this end, we introduce the concept of loss function. This is a function $L(y,f(x))$ which tells us the cost of classifying instance $x\in \mathcal{X}$ as $y\in \mathcal{Y}$.
For example, the natural loss function is the 0-1 loss
\begin{equation}\label{01loss}
\begin{split}
L_{0/1}(y,f(x))=I(y,f(x)),
\end{split}
\end{equation}
where
$$I(u,v) =\left\{ \begin{array}{l}
	0\text{,\,\,}if\,\,u=v\text{,}\\
	1\text{,\,\,}if\,\,u\ne v.\\
\end{array} \right.$$
The function \eqref{01loss} determines the probability of different answers given by the indicator function $f(x)$ and label $y$. We call the case of different answers a classification error.
While the loss function measures the error of a function on some individual data sample, the risk of a function is the average loss over data samples generated according to the underlying distribution $F(x,y)$,
\begin{equation}\label{exprisk}
\begin{split}
R(f)&=\mathbb{E}_{X,Y}[L(y,f(x))]=\int_{\mathcal{X}\times\mathcal{Y}}{L(y,f(x))dF(x,y)}.
\end{split}
\end{equation}
Therefore, the goal of learning is to find the function which minimizes the risk function \eqref{exprisk} in the given set of indicator function $\mathcal{F}$ when the probability measure $F(x,y)$ is unknown but training data \eqref{Data} are given.

Since the training data \eqref{Data} sampled from the underlying probability distribution $F(x,y)$, we can infer a function $f(x)$ from \eqref{Data} whose risk is close to expect risk. In practical, we approximate the unknown cumulative distribution function $F(x,y)$ by its joint empirical cumulative distribution function
\begin{equation}\label{P(x,y)}
\begin{split}
F_m(x,y)=\sum_{i=1}^{m}\theta(x-x_i)\theta(y-y_i),
\end{split}
\end{equation}
where the one-dimensional step function is defined as
\begin{equation}\label{theta}
\begin{split}
\theta \left( x-x_i \right) =\left\{ \begin{array}{l}
	1,\ if\ x\ge x_i\\
	0,\ otherwise\\
\end{array} \right.
\end{split}
\end{equation}
and the multi-dimensional step function for $x=\left( x^1,...,x^d \right)$ is defined as
$$
\theta \left( x-x_i \right) =\prod_{k=1}^d{\theta \left( x^k-x_i^k \right)}.
$$
Using approximation \eqref{P(x,y)} in the expected loss function \eqref{exprisk}, we obtain the following empirical loss function
\begin{equation}\label{emploss}
\begin{split}
R_m(f)=\frac{1}{m}\sum_{i=1}^{m}L(y_i,f(x_i)).
\end{split}
\end{equation}
That is, given some training data \eqref{Data}, a loss function $L$ and a function space $\mathcal {F}$ to work with, we define the classifier $f_m$ as
$$
f_m:= \underset{f\in \mathcal{F}}{\arg\min}\ R_{emp}(f).
$$
This approach is called the empirical risk minimization (ERM) induction principle. The ERM principle is conducted with the law of large numbers well. However, when the sample size $m$ is relatively small, ERM may be inconsistent with the expected risk minimization.
To solve this problem, Vapnik \citep{vapnik1999nature} considerers the effect of the set of functions $\mathcal{F}$ on the expected risk and restrict the set of admissible functions to render empirical risk minimization consistent, and gets the following two lemmas.
\begin{lemma}
Let $h$ be the Vapnik$-$Chervonenkis (VC) dimension of $\mathcal{F}$, if $m>h$,and $m\varepsilon^2\ge 2$, where $\varepsilon$ is any positive number, then we have
\begin{equation}\label{prob}
P\bigg\{\underset{f\in \mathcal{F}}{sup}\left( R\left( f \right) -R_{emp}\left( f \right) >\varepsilon \right)\bigg\} \le 4\exp \left( h\left( \ln \frac{2m}{h}+1 \right) -\frac{m\varepsilon ^2}{8} \right).
\end{equation}
\label{lem-2}
\end{lemma}
\begin{lemma}
Let $h$ be the VC dimension of $\mathcal{F}$, and if $$m > h~\text{and}~h(ln(\frac{2m}{h}+1)+ln\frac{4}{\delta}\geq\frac{1}{4},$$
then for any probability distribution $F(x,y)$, any $\delta\in(0,1]$ and any function $f$ in $\mathcal{F}$ holds with probability at least $1-\delta$ of the inequality
\begin{equation}\label{bound}
\begin{split}
R(f)\le R_{emp}(f)+\sqrt{\frac{8}{m}\left(h\left(ln\frac{2l}{h}+m\right)+ln\frac{4}{\delta}\right)}.
\end{split}
\end{equation}
where the right-hand side of the equation \label{prob} be equal to $\delta$, and
$$
\varepsilon=\sqrt{\frac{8}{m}\left(h\left(ln\frac{2m}{h}+m\right)+ln\frac{4}{\delta}\right)}.
$$
\label{lem-3}
\end{lemma}
Based on the above lemma, Vapnik$-$Chervonenkis theory \citep{vapnik1971uniform} introduces a generalization of the empirical risk minimization method which called structural risk minimization (SRM) induction principle \citep{vapnik1999overview,von2011statistical}.
The principle of SRM for solving classification problem is to choose a set of decision functions $\mathcal{F}(t)$ that depends on parameter $t$, which satisfies the following nested structure that increases as $t$ increases
 $$
 \mathcal{F}(t_1)\subset \mathcal{F}(t_2)\ \ \forall t_1<t_2.
 $$
For each $t$, a function $f_t$ is found in the set $\mathcal{F}(t)$ that minimizes its empirical risk, and at the same time, there is a value of structural value for $f_t$. We want to choose the best $\hat{t}$ that minimizes the structural risk and use the corresponding function $f_{\hat{t}}$ as the decision function.

An implicit way of dealing with nested function spaces is the principle of regularization. Instead of minimizing the empirical risk $R_{emp}(f)$ and then expressing the generalization ability of the resulting classifier $f_t$ using some capacity measure of the underlying set of functions $\mathcal{F}$, a more direct approach is preferred to minimize the regularization risk
$$
R_{reg}(f) = R_{emp}(f) + \lambda \Omega(f),
$$
here $\Omega(f)$ is the so-called structure term or regularizer, and this regularizer is supposed to punish overly complex functions.

\subsection{Conditional probability estimation via CDF}\label{subsec24}
In the classical approach to the learning machines (analyzed by the above VC theory), we ignored the fact that the desired decision rule is related to the conditional probability function $P(y=1\vert x)$ used by supervisor and the different distribution structure of the data $F(x)$, which will have an impact on the learned mapping $f: \mathcal{X}\to \{0,1\} $.
\cite{vapnik2019rethinking} proposes to consider the problem of directly estimating the conditional probability function $P(y = 1\vert x)$ using training data \eqref{Data} as the main problem of learning. And further \citep{vapnik2019rethinking,vapnik2020complete} gives a direct setting of conditional probability estimation problem based on the standard definitions of a density and a conditional probability.

From the perspective of probability, to predict the label $y$, we consider the special (but common) case $P(y=1\vert x)=1-P(y=0\vert x)$.
Thus, we describe the classification problem as the estimation problem of conditional probability function, i.e., $f_0(x)=P(y=1\vert x)$,
here $f_0(x)$ is the function to be estimated and is a single valued deterministic function that at every $x\in R^d$ specifies the probability $y$.

Consider the definition of cumulative distribution function (CDF), the probability $F(x)=P\{\omega <x\}$ is CDF, where $\omega\in R^d$ be a random vector.
The density $p(x)$ is defined from CDF if there exits a nonnegative function $p(x)$ such that for all $x$ the following equality is valid
\begin{equation}\label{pdf}
\begin{split}
\int_{-\infty}^{x}{p(\omega)}d\omega=F(x).
\end{split}
\end{equation}
The density $p(y=1,\omega)$ is defined as the solution of the integral equation $$\int_{-\infty}^{x}{p(y=1,\omega)d\omega}=F(y=1,x)$$.
The definitions of conditional probability function $f_0(x)=P(y=1 \vert x)$ for continuous $x$ is as follows
\begin{equation}\label{defcon}
\begin{split}
f_0\left( x \right) p\left( x \right) =p\left( y=1,x \right)~~~~ \text{or} ~~~~ f_0\left( x \right) = \frac{p\left( y=1,x \right)}{p\left( x \right)}, ~p(x)>0
\end{split}
\end{equation}
where $p\left( y=1,x \right)$ is the joint density function of $x$ and $y=1$, $p(x)$ is the probability density function (PDF) of $x$, $f_0(x)$ is the ratio of the two density functions. Different from using \eqref{defcon} and Bayes theory \citep{berger2013statistical} to calculate $f_0(x)$, we consider another direct definition of the conditional probability function based on \eqref{pdf} and \eqref{defcon}, which is defined as the solution of the Fredholm equation
\begin{equation}\label{FredEq}
\begin{split}
\int{G\left( x-x' \right)f_0(x')}dF(x') =\int{G\left( x-x' \right)}dF\left( y=1,x' \right),
\end{split}
\end{equation}
where $G\left( x-x' \right)$ is any kernel from $L_2$ space,
$F(x)$ and $F(y=1,x)$ are the CDF of $x$ and joint distribution function of $x$ and $y=1$, respectively. The solution of the Fredholm equation \eqref{FredEq} directly defines the conditional probability function $f_0(x)=P(y=1\vert x)$ when the kernel $G\left( x-x' \right)$ is given and $F(x)$ and $F(x,y=1)$ are known.

Compared to \eqref{defcon}, there is no density function in the integral equation, but only distribution functions in \eqref{FredEq}. There are at least two main advantages of using \eqref{FredEq} to calculate $f_0(x)$.\\
i) The distribution function is a monotonically non-decreasing function with a value range of $[0,1]$, which is easier to be estimated than the ratio of two probability density functions, for example, it can be approximated by the empirical cumulative distribution function (ECDF).\\
ii) The conditional probability function directly controls many different statistical properties that are expressed in the training data, such as the distribution of the data. Keeping these invariants in the rule is equivalent to incorporating some prior knowledge and allowing the learning machine to extract additional information from the data that cannot be directly extracted by classical methods.

In order to solve \eqref{FredEq}, \cite{vapnik2019rethinking} estimates $f(x)$ using the empirical distribution approximations of $F(x)$ and $F(y=1,x)$ which are given by $F_{m}\left( y=1,x \right) =\frac{1}{m}\sum_{i=1}^m{y_i\theta \left( x-x_i \right)}$ and $F_{m}\left( x \right) =\frac{1}{m}\sum_{i=1}^m{\theta \left( x-x_i \right)}$. At this time, the strict equality relation \eqref{FredEq} is hard to establish.
It solves Eq.\eqref{FredEq} in the set of function $\mathcal{F}$ by minimizing the distance of the $L_2$ metric
\begin{equation}\label{L2Loss}
\begin{split}
\rho^2=\frac{1}{m^2}\int{\left(\sum_{i=1}^{m}{G(x-x_i)f_0(x_i)}-\sum_{i=1}^{m}{y_iG(x-x_i)}\right)^2}d\mu(x),
\end{split}
\end{equation}
where $\mu(x)$ is a probability measure defined on domain $D$ consisting of $x\in R^n$.  

Under the given training samples, Eq.\eqref{L2Loss} simplifies to
\begin{equation}\label{L2Loss1}
\begin{split}
\rho^2=\frac{1}{m^2}\sum_{i,j=1}^{m}{(f_0(x_i)-y_i)(f_0(x_j)-y_j)v(x_i,x_j)},
\end{split}
\end{equation}
here $v$ is an $m\times m$-dimensional symmetric non-negative matrix and its specific form is as follow
\begin{equation}\label{Vmatrix}
\begin{split}
v(x_i,x_j)=\int{G(x-x_i)G(x-x_j)d\mu(x)}.
\end{split}
\end{equation}

Obviously, considering the characteristics of the data, we can choose the corresponding appropriate kernel functions $G(\cdot)$ and $\mu(x)$, so as to get a more accurate location information of samples.
And the value of $v$ dose not depend on labels $y_i$ but shows the location of each sample in the distribution. It's an objective property of the input while it's ignored in most researches, such as the classical least square method only consider the residual error $\Delta _i=f_0\left( x_i \right) -y_i,i=1,...,m$ of each $x_i$ and take less account of the distribution of input data.

As the approximation \eqref{L2Loss1} of Fredholm equation may be an ill-posed problem, \cite{vapnik2019rethinking} gives the solution of Tikhonov's regularization method as
\begin{equation}\label{aim}
\begin{split}
R(f)=\frac{1}{m^2}\sum_{i,j=1}^{m}{(f_0(x_i)-y_i)(f_0(x_j)-y_j)v(x_i,x_j)}+\gamma\Vert f(x) \Vert_2
\end{split}
\end{equation}
where $\gamma>0$ is the regularization constant.

By defining the following notations:
$A=(\alpha_1,...,\alpha_m)^T$, $\mathcal{K}(x)=(K(x_1,x),K(x_2,x),\\...,K(x_m,x))^T$, $m\times m$-dimensional matrix $K$ with $K_{ij}=K(x_i,x_j)$ being the $ij^{th}$ entry, $Y=(y_1,y_2,...,y_m)^T$, $1_m=(1,1,...,1)^T$, $I$ is the $m\times m$-dimensional identity matrix.
Then $f(x)=A^T\mathcal{K}(x)+c$ and $R(x)$ can be rewritten as
$$
R(A)=(KA+cl_m)^TV(KA+cl_m)-2(KA+cl_m)^TVY+Y^TVY+\gamma A^TKA
$$
and the closed form solution for $A$ and $c$ are given by $A=A_b-cA_c$, where $A_b=(VK+\gamma I)^{-1}VY$, $A_c=(VK+\gamma I)^{-1}V1_m$ and $c=\frac{1_m^TV(KA_bY)}{1_m^TV(KA_c-1_m)}$.

Although this model looks very similar to the weighted support vector machines \citep{lin2002fuzzy,vapnik2009new,lapin2014learning} and the weighted least squares support vector machines \citep{suykens2002weighted,yang2018study}, their starting points and functions are very different. In the following, we will explain their differences by further analyzing the estimated distribution function.

\section{Classification by cumulative distribution estimation theory}\label{sec3}

\subsection{Probability calculation via Fredholm equation}\label{subsec31}
In this subsection, we consider the conditional probability $P(y=1\vert x)$ calculation from the Fredholm equation theory.
According to the definition of the density function $p(x)$ and $p(y=1,x)$ in \eqref{pdf}, we have
\begin{equation}\label{derivatives}
\begin{split}
P(y=1\vert x)=\frac{p(x,y=1)}{p(x)}=\frac{\partial\left(F\left(x,y=1\right)\right)}{\partial\left(F\left( x\right)\right)},
\end{split}
\end{equation}
when the derivatives of $F(x)$ and $F(x,y=1)$ exist and non-zero. Therefore, \eqref{FredEq} can be simplified to
\begin{equation}\label{Fredshift}
\begin{split}
\int{G\left( x-x' \right)}f_0(x') dF\left( x' \right)=\int{G\left( x-x' \right) \left( \frac{\partial (F\left( x',y=1 \right))}{\partial(F\left( x' \right))} \right)}dF\left( x' \right),
\end{split}
\end{equation}
where $\partial(v)$ represents the gradient of $v$.

Instead of treating each random variable $X$  as an independent individual and aiming to let $f_0(x)=P(y=1\vert x)$, where $P(y=1\vert x)$ denotes the true and known conditional probability of $x$. We see that there are different ways to compute the conditional probability $f_0(x)$ from \eqref{FredEq}, \eqref{derivatives} and \eqref{Fredshift} by using the cumulative distribution function.  Here each random variable $X$ is in the entire distribution, and the method of directly estimating conditional probability using \eqref{FredEq} considers the distribution. For each sample $x$, a relationship with the relative position of other samples in the distribution is established, which contains the prior information of the data.

In real-world problem, the classical setting of output is hard output in classification, that is the label of the sample is known instead of the conditional probability of the sample. At this time, we take into account the fact that the transformation rule for classification.
For binary classification problem, we have
$$
r(x)=\theta\left(P(y=1\vert x)-0.5\right)=\theta\left(f_0(x)-0.5\right),
$$
where $\theta(z)$ is the step function. Thus, the rule $r (x)$ classifies vector $x$ to the
first class ($y = 1$) if the conditional probability $f_0(x)$ exceeds 0.5.

Different from estimating the conditional probability function $f(x)$, we compute the hard output classification rule $r(x)$. Similarly, the standard definitions of $r(x)$ for continuous $x$ is as follows
\begin{equation}\label{r(x)y}
\begin{split}
r(x)=\theta\left(\frac{p(y=1,x)}{p(x)}-0.5\right)=y,\ y\in\{0,1\}.
\end{split}
\end{equation}
Then corresponding to \eqref{FredEq}, we have the direct definition of decision rule $r(x)$
\begin{equation}\label{FredEqy}
\begin{split}
\int{G\left( x-x' \right)}r(x') dF\left( x',y' \right) 
= \int{ G( x-x')y'}dF( x',y').
\end{split}
\end{equation}

Compare the two pairs of expressions \eqref{defcon} and \eqref{FredEq} or \eqref{r(x)y} and \eqref{FredEqy}, we have\\
 i), when the kernel function is chosen as \eqref{theta} (This kernel function was mentioned and used in \citep{vapnik2020complete}, which covers the setting of cumulative distribution function), \eqref{FredEqy} or \eqref{FredEq} is an equation about the distribution function, while \eqref{r(x)y} or \eqref{defcon} is an equation about the density function. \eqref{FredEqy} or \eqref{FredEq} is easier to compute with respect to \eqref{r(x)y} or \eqref{defcon} because in probability theory, the distribution function is determined first while the density function is defined by the distribution function. And the distribution function is unique, the density function is not necessarily unique.\\
ii), when the choice of kernel function is extended to any kernel function from the $L_2$ space, this involves two different modes of convergence in Hilbert space \citep{vapnik2020complete}, which describes the relationship between a sequence of functions $r_m(x)\in L_2$ and the desired function $r_0(x)$.
The strong mode of convergence (convergence in metrics) is expressed as
$$
\underset{m\to \infty}{lim}\Vert f_m(x)-f_{0}(x)\Vert =0, \ \ \forall x.
$$
The weak mode of convergence (convergence in inner products) is
$$
\underset{m\to \infty}{lim}\langle G(x),\{f_m(x)-f_{0}(x)\}\rangle=0, \ \ \forall G(x)\in L_2.
$$
Obviously, \eqref{FredEqy} or \eqref{FredEq} is an implementation of weak convergence, which is more common in real life than strong convergence mode. From this point of view, probability calculation via Fredholm equation is more general than the traditional one due to the weak convergence is also strong convergence.

The above discussion is a description of the outcomes of random experiments for a theoretical model from the perspective of probability theory, without considering the specific data and hypothesis space of functions. The next part will discuss the classification from a statistical perspective.

\subsection{Expected risk of Fredholm equation}\label{subsec32}

Consider the statistical estimation problem with an input space $\mathcal{X}$ and the output set $\mathcal{Y}=\{0,1\}$ and a set of training samples \eqref{Data}.
If we estimate \eqref{FredEq} under the above assumptions, the integral equation does not always strictly hold.
In order to find the optimal decision function satisfied the above assumptions, we firstly introduce the concept of loss function to evaluate the pros and cons of decision function $f(x)$.
\begin{definition}
	\label{label}
    \textbf{(Loss function based on Fredholm equation)} For quadruple $(y,f(x),F(x,y),G(\cdot))\in \mathcal{Y}\times \mathcal{Y}\times \mathcal{P}\times \mathcal{P}$, where $\mathcal{Y}=\{0,1\}$, $\mathcal{P}= [0,1]$, $y$ is an observation corresponding to the input $x$, and $r(x)$ is the conditional probability value of the decision function $r$ at $X$. For a given kernel function $G(\cdot)\in L_2$ and any $x\in \mathcal{X},y\in \mathcal{Y}$, if the mapping $L_{Fred}:\mathcal{Y}\times \mathcal{Y}\times \mathcal{P}\times \mathcal{P}\to [0,\infty)$, there are $L_{Fred}(y,y,F(x,y),G(\cdot))=0$, then it is called $L_{Fred}$ a loss function based on the Fredholm equation.
\end{definition}

Now, the classification problem becomes to estimate Eq.\eqref{FredEq} with minimized losses.
For a given kernel function $G(x-x')$, the loss function based on the Fredholm equation corresponding to the decision function $f(x)$ is expressed as
\begin{equation}\label{LossFred}
\begin{split}
&L_{Fred}(y,f(x),F(x,y),G(\cdot))\\&=L\left(\int{G(x-x')r(x')dF(x',y')},\int{G(x-x')y'dF(x',y')}\right).
\end{split}
\end{equation}

The value of loss function \eqref{LossFred} is just a quantitative measure of how well a decision function $f(x)$ performs at any input $x$. For the population, the expected risk is introduced below to assess the overall effect.
The expected loss of \eqref{LossFred} is written as
\begin{equation}\label{ExpRisk}
\begin{split}
R_{Fred}(f)&=\mathbb{E}_{X}\left[L \left(\int_{\mathcal{X}\times \mathcal{Y}}{G(x-x')r(x')dF(x',y')},\int_{\mathcal{X}\times \mathcal{Y}}{G(x-x')y'dF(x',y')}\right) \right]\\
&=\int_{\mathcal{X}}L\left(\int_{\mathcal{X}\times \mathcal{Y}}{G(x-x')r(x')dF(x',y')},\int_{\mathcal{X}\times \mathcal{Y}}{G(x-x')y'dF(x',y')}\right)dF(x),
\end{split}
\end{equation}
where $F(x)$ is a distribution function with respect to random variables $X$.

Compared to the loss in \eqref{exprisk}, the loss function \eqref{LossFred} measures the difference between the two sides of the integral equation \eqref{FredEq}, it is the difference between a series of $y$ and $f(x)$ under the concept of distribution. While for \eqref{exprisk} and \eqref{ExpRisk}, we have
\newtheorem{thm}{\bf Theorem}[section]
\begin{thm}\label{thm1}
Given $R(x)$ in \eqref{exprisk} and $R_{Fred}(x)$ in \eqref{ExpRisk}, assuming that the distribution structure of the data is not considered, i.e., $F(x, y)=\theta(x-x_0)\theta(y-y_0)$, $R_{Fred}(x)$ degenerates to classical expect risk $R(x)$, where $$
F(x,y)=\theta(x-x_0)\theta(y-y_0)=\left\{ \begin{array}{l}
	1,\ x\ge x_0,y\ge y_0\\
	0,\ otherwise.\\
\end{array} \right.
$$ 
\end{thm}
\vspace{-0.5cm}

\begin{proof}
We first consider the distributions defined by the Dirac function and the commonly used distributions (with the normal distribution as an example). The Dirac function has a value at only one point and the integral over the domain of definition is 1. Its definition is as follow
$$
\delta(x-x_0)=\left\{ \begin{array}{l}
	+\infty,\ x= x_0\\
	0,\ \ \ \ \ x\neq x_0\\
\end{array} \right.
$$
and its corresponding cumulative distribution function is
$$
F(x)=\theta(x-x_0)=\left\{ \begin{array}{l}
	1,\ x\ge x_0\\
	0,\ otherwise.\\
\end{array} \right.
$$
The image of the functions are shown in Figure \ref{Fig1:a} and \ref{Fig1:b}. Obviously, the function only considers a certain point, while the commonly used density function such as the normal distribution shown in Figure \ref{Fig1:c} and \ref{Fig1:d}, each $x$ has a corresponding non-zero value, reflecting the position of each point in the entire distribution. Thus when we do not consider the distribution structure between the data, the Dirac function $\delta(\cdot)$ \citep{dirac1981principles,federer2014geometric} or its cumulative distribution function $\theta(\cdot)$ is used.
\begin{figure*}[h]
\setlength{\abovecaptionskip}{0.cm}
\setlength{\belowcaptionskip}{-0.cm}
\centering
    \subfigure[Dirac Function]{\includegraphics[width=0.32\textheight]{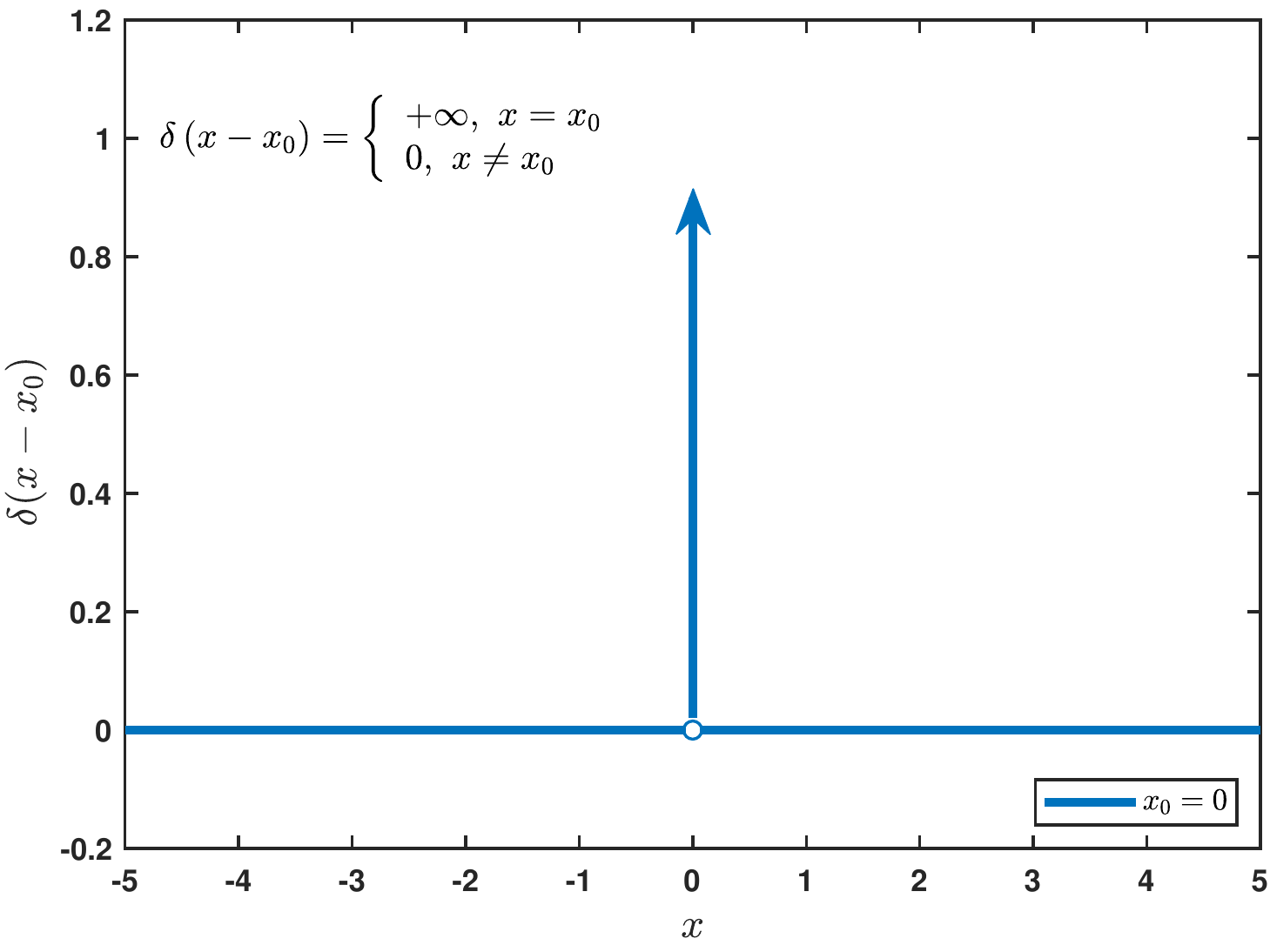}\label{Fig1:a}}
    \subfigure[Step Function]{\includegraphics[width=0.32\textheight]{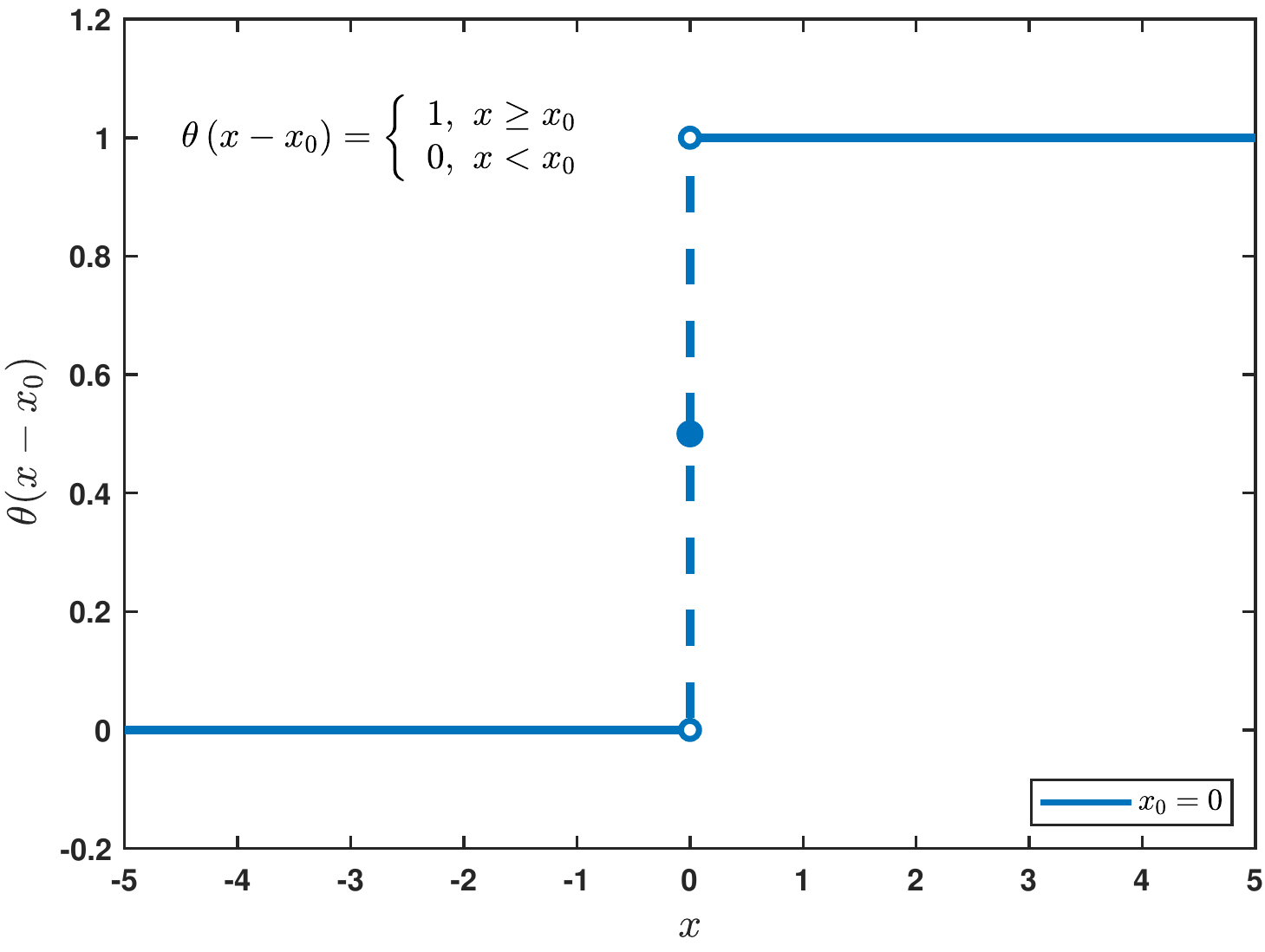}\label{Fig1:b}}
    \subfigure[Normal PDF]{\includegraphics[width=0.32\textheight]{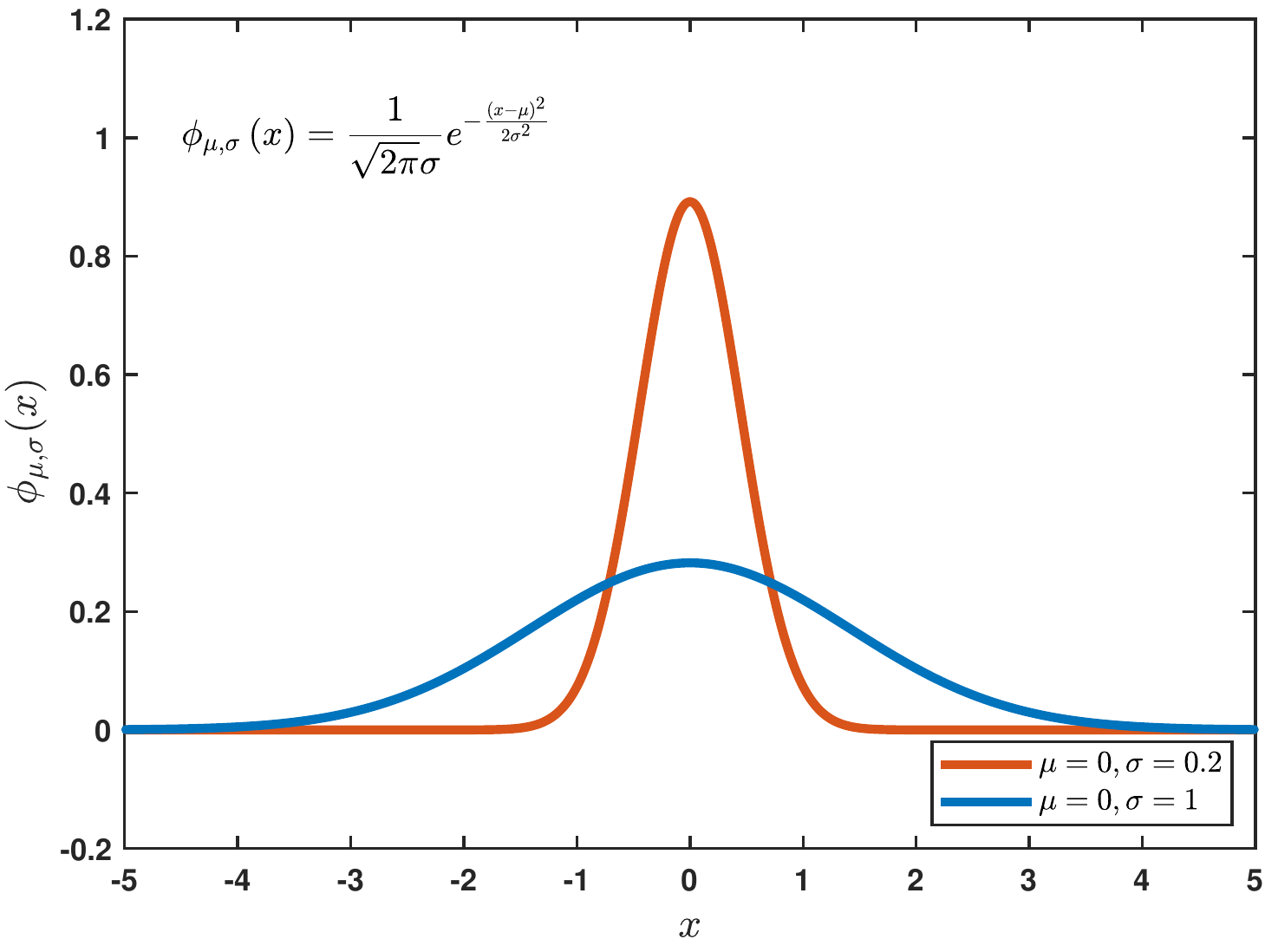}\label{Fig1:c}}
    \subfigure[Normal CDF]{\includegraphics[width=0.32\textheight]{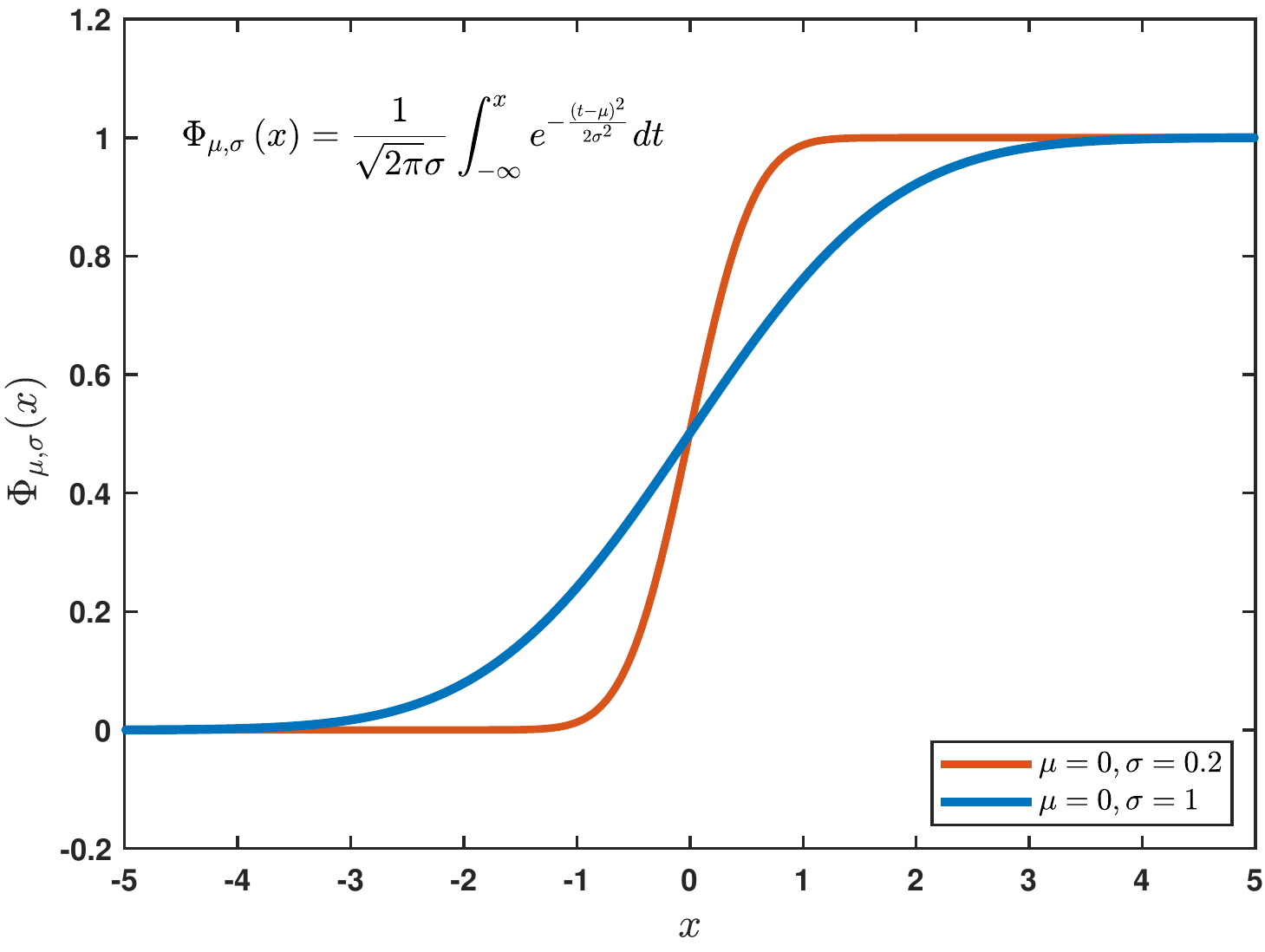}\label{Fig1:d}}
\caption{The PDF and CDF of Dirac and Normal distribution.}
\label{fig.1}
\end{figure*}

Denote $L_{left}=\int_{\mathcal{X}\times \mathcal{Y}}{G(x-x')r(x')dF(x',y')}$.
Since the integrated function is independent of $y$, $F(x,\infty) = F_{X}(x)=\theta (x-x_0)$ and we have
\begin{equation}\label{left}
\begin{split}
L_{left}=\int_{\mathcal{X}}{G(x-x')r(x')dF(x')}.
\end{split}
\end{equation}
Put $F(x)=\theta (x-x_0)$ into \eqref{left}, then
\begin{equation}\label{proof1}
\begin{split}
L_{left}&=\int{G(x-x')r(x')d\theta (x'-x)}\\
&=\int{G(x-x')\delta (x'-x)r(x')dx'}\\
&=G(x-x)r(x)\\
&=r(x).
\end{split}
\end{equation}

Similarly, for $F(x, y)=\theta(x-x_0)\theta(y-y_0)$, $L_{right}=\int_{\mathcal{X}\times \mathcal{Y}}{G(x-x')y'dF(x',y')}=y$. Therefore, $R(x)$ in \eqref{exprisk} is a special case of \eqref{ExpRisk} and $R_{Fred} (x)$ is a more general expression of expected risk.
\end{proof}

In short, the statistical formulation of the original classification problem can be expressed as: in a given set of functions $\mathcal{F}$, estimate the one that minimizes the expected risk functional \eqref{ExpRisk}, under the assumption that probability measure $F(x)$ and $F(x,y)$ are unknown but $m$ $i.i.d.$ pairs $(x_i,y_i)$, where $i=1,...,m$, satisfied this distribution are given.

Theoretically, the proposed expected risk is an implementation of weak convergence approximating strong convergence, while the traditional one only considers strong convergence mode.
Further, from the structure of the above expect risk, the inner layer in equation \eqref{ExpRisk} over the $\mathcal{X}\times \mathcal{Y}$ space describes the difference between the estimated and true values of a series of pairs $(x'_i,y'_i)$, and the outer layer over the $\mathcal{X}$ space and the kernel function $G(x-x')$ jointly construct information about the structure of the distribution of $x'$. Thus, from the perspective of empirical estimation, the outer layer structure captures information about the distribution of the data, which in turn improves the performance of the estimated classifier. The details of implementation will be elaborated in Section 4.

\subsection{Empirical risk of Fredholm equation}\label{subsec32}

Now, we express the integral equation \eqref{FredEqy} by the form
\begin{equation}\label{oper}
Af=F,
\end{equation}
here $A$ is a linear operator which maps the elements $f$ of a metric space $E_1$ into the elements $F$ of a metric space $E_2$.
Then the empirical risk of the given sample $\{(x_i,y_i)\}_{i=1}^{m}$ is expressed as the distance on both sides of the equation \eqref{FredEqy} in $E_2$ space. We mark it as $\rho_{E_2}(A_mf,F_m)$.

We say that the problem of solving operator equation \eqref{oper}
in the set $\mathcal{F}$ is well-posed if the solution exits, is unique, and is continuous. The problem is called ill-posed if at least one of the three above conditions is violated.
The solution of ill-posed problems is based on the following lemma.
\begin{lemma}
(Lemma about inverse operator.) If A is a continuous one-to-one operator defined on compact set $\mathcal{M}$ of functions {f}, then the inverse operator $A^{-1}$ is continuous on the set $\mathcal{N} = A\mathcal{M}$.
Consider a continuous non-negative functional W(f) and the set of functions
$$
\mathcal{M_C} = \{f : W(f)\leq C\},
$$
defined by a constant $C > 0$. Let the set of functions $\mathcal{M_C}$ be convex and compact for any
C. Suppose that the solution of operator equation belongs to compact sets $\mathcal{M_C}$.
\label{lem-1}
\end{lemma}
The idea of solving ill-posed problems is to choose an appropriate compact set (i.e., to choose a constant $C^*$) and then solve Eq.\eqref{oper} on the compact set of functions defined by
$C^*$. In other words, to minimize the distance on both sides of the equation \eqref{FredEqy} in $E_2$ space
$$
\rho=\rho_{E_2}(A_mf,F_m)
$$
over functions $f$ subject to the constraint
$$
W(f)\leq C^*.
$$
This is similar to that of Tikhonov's regularization method. In this way, the function
\begin{equation}\label{R(f)}
\begin{split}
R(f)=\rho_{E_2}(A_mf,F_m)+\gamma W(f),
\end{split}
\end{equation}
is minimized, where $\gamma > 0$ is a regularization parameter.
Obviously, this idea of solving ill-posed problem is the same as structural risk minimization in VC theory. In both cases, a structure is defined on the set of functions. When solving well-posed problems, elements of structure should have finite VC-dimension. When solving ill-posed problems, elements of structure should be compact sets.

In practical, $F(x)$ in expected risk \eqref{ExpRisk} can be estimated with given $i.i.d.$ data. A common method is to approximate the unknown cumulative distribution functions by their empirical cumulative distribution functions \eqref{P(x,y)}.
Using approximation \eqref{P(x,y)} in the expected loss function \eqref{ExpRisk}, we obtain the empirical loss function and the expression of the specific empirical risk will be described in detail in the next section.

\section{Classification by cumulative distribution estimation}\label{sec4}

\subsection{Model estimation with loss function}\label{subsec33}
Now, we consider the empirical estimation of the expected risk \eqref{ExpRisk} with the 0-1 loss function \citep{cortes1995support,wang2021support}.
As is discussed above, we approximate the unknown cumulative distribution function $F(x',y')$ by using empirical cumulative distribution function \eqref{P(x,y)}.
Putting \eqref{P(x,y)} into \eqref{ExpRisk} instead of unknown elements $F(x',y')$, we obtain the following approximation to \eqref{ExpRisk} as
\begin{equation}\label{EmpRisk}
\begin{split}
\hat{R}(f) =\int{I\left(\sum_{i=1}^{m}{G(\hat{x}-x_i)r(x_i)},\sum_{i=1}^{m}{G(\hat{x}-x_i)y_i} \right)} d\mu(\hat{x}).
\end{split}
\end{equation}
Note that $\mu(\hat{x})$ is a cumulative distribution function which define the concept of closeness of $\hat{x}$.

\textbf{Remark}: Random variables $x$ and $x'$ in \eqref{ExpRisk} are both from the fixed unknown distribution $P$ on $\mathcal{X}$.
By using the empirical cumulative distribution function in \eqref{EmpRisk},
$x'$ is corresponding to the labeled sample $x_{i}$ from training set, and $x$ is corresponding to $\hat{x}$ (here $\hat{x}$ is an arbitrary sample from the distribution, whether it's labeled or not, continuous or not). We point out that $\mu(\hat{x})$ could be estimated by the empirical data or other different ways.

For each $\hat{x}$, we substitute training data $(x_i,y_i)$ instead of a continuous variable $x'$ in \eqref{ExpRisk}. At this time, \eqref{FredEqy} is hard to satisfied because the empirical estimation is hard to be exactly equal and we have
\begin{equation}\label{Af=F}
\begin{split}
\sum_{i=1}^{m}{G(\hat{x}-x_i)r(x_i)}\approx \sum_{i=1}^{m}{G(\hat{x}-x_i)y_i}.
\end{split}
\end{equation}
Therefore, the indicator function $I(\cdot,\cdot)$ in the empirical approximation expression \eqref{EmpRisk} obtained from the expected risk takes a high probability of 1. There will be a large bias in evaluating the performance of $f(x)$ by using 0-1 loss function above. That is to say, the traditional 0-1 loss is not suitable for this type of problem. In this case, for each $\hat{x}$, we give a small tolerance to empirical estimation
$$
\Vert \sum_{i=1}^{m}{G(\hat{x}-x_i)r(x_i)} -\sum_{i=1}^{m}{G(\hat{x}-x_i)y_i} \Vert \le \varepsilon,
$$
where $\varepsilon>0$ is a small scalar. Then, the 0-1 loss converts to the 0-1 loss with $\varepsilon$-insensitive zone, and \eqref{EmpRisk} is transformed to the following unconstrained optimization problem
\begin{equation}\label{eps01}
\begin{split}
\underset{f}{\min}\int{I_{\varepsilon}\left( \sum_{i=1}^m{G\left( \hat{x}-x_i \right) r\left( x_i \right)}, \sum_{i=1}^m{G\left( \hat{x}-x_i \right) y_i } \right) d\mu \left( \hat{x} \right)}+\lambda W(f),
\end{split}
\end{equation}
where 0-1 loss function with $\varepsilon$-insensitive zone is
$$
I_{\varepsilon}(u,v)= \left\{ \begin{array}{l}
	0\text{,\,\,}if\,\,\vert u-v\vert\leq \varepsilon\text{,}\\
	1\text{,\,\,}otherwise.\\
\end{array} \right.
$$
$W(f)$ is the regularization term to avoid the ill-posed problem and
$\lambda$ is a positive regular parameter.

Noticing that both $I_{\varepsilon}(u,v)$ and $r(\cdot)$ are nonconvex discontinuous, arises from the fact that the gradient of the objective function of \eqref{eps01} is either not defined or is equal to zero, and \eqref{eps01} is difficult to optimize.
In order to solve this problem, we use common alternative to utilize a convex surrogate to substitute the non-convex 0-1 type function as
\begin{equation}\label{eps-L1}
\begin{split}
\underset{f}{\min}\int{\ell_{\varepsilon}\left(\sum_{i=1}^m{G\left( \hat{x}-x_i \right) \left( f\left( x_i \right) -y_i \right)} \right) d\mu \left( \hat{x} \right)}+\lambda W(f),
\end{split}
\end{equation}
where $\varepsilon$-insensitive loss function is
$$
\ell_{\varepsilon}(z)= \left\{ \begin{array}{l}
	0\text{,\,\,\,\,\,\,\,\,\,\,\,\,\,\,\,\,}if\,\,\vert z\vert\leq \varepsilon\text{,}\\
	\vert z\vert-\varepsilon\text{,\,\,}otherwise.\\
\end{array} \right.
$$

We now discuss the specific form of estimating $\mu(\hat{x})$.
Specially, two ways to calculate the $\mu(\hat{x})$ are given:
\begin{description}
\item[\textbf{1)(Parameter estimation of $\mu(\hat{x})$)}]
Assumes that the probability distribution $\mu(\hat{x})$ belongs to a certain family of distributions, such as obeying a uniform distribution. Then, we just need to estimate the parameter of the distribution by maximum likelihood estimation.
\begin{example}
 Consider one-dimensional case of continuous variable $\hat{x}\in [-a,a]$, where $\hat{x}$ is uniformly distributed on $[-a,a]$(i.e.,$\mu(\hat{x})=\frac{\hat{x}+a}{2a})$, we have
 $$\rho_{E_2}(A_mf,F_m)=\frac{1}{2a}\int_{-a}^{a}{\ell_{\varepsilon}\left(\sum_{i=1}^m{G\left( \hat{x}-x_i \right) \left( f\left( x_i \right) -y_i \right)} \right)d\hat{x} }.$$
For multidimensional case of $\hat{x}=(\hat{x}^1,\hat{x}^2,...,\hat{x}^d)\in [-a_1,a_1]\times ...\times[-a_d,a_d]$, we have
\begin{equation}\label{para}
\begin{split}
\rho_{E_2}(A_mf,F_l)&=\prod_{k=1}^d{\frac{1}{2a}\int_{-a}^{a}{\ell_{\varepsilon}\left(\sum_{i=1}^m{G\left( \hat{x}^k-x_i^k \right)\left( f\left( x_i \right) -y_i \right)} \right) d\hat{x}^k}}.
\end{split}
\end{equation}

\label{eg1}
\end{example}
 \item[\textbf{2)(Non-parameter estimation of $\mu(\hat{x})$)}]
 If the certain family of distributions is unknown, $\mu(\hat{x})$ can be approximated with an empirical distribution function.
\begin{example}
Given $i.i.d.$ data $\hat{x_1},\hat{x_2},...,\hat{x_N}$ (whether they are labeled or not) from the unknown distribution $P$ on $\mathcal{X}$, consider the case $\mu(\hat{x})=F_N(\hat{x})=\frac{1}{N}\sum_{s=1}^{N}{\theta(\hat{x}-x_s)}$. For this measure, we obtain the following estimation
\begin{equation}\label{nonpara}
\begin{split}
\rho_{E_2}(A_lf,F_l)=\frac{1}{N}\sum_{s=1}^{N}{\ell_{\varepsilon}\left(\sum_{i=1}^m{G\left( \hat{x}_s-x_i \right)\left( f\left( x_i \right) -y_i \right)}\right)}.
\end{split}
\end{equation}
\label{eg2}
\end{example}
\end{description}
\vspace{-0.5cm}

Now, we have known all other statistical variables except for $f(x)$. In the following, we are looking for solutions of our inference problem in the set of functions $\mathcal{F}$ that belong to Reproducing Kernel Hilbert Space (RKHS) associated with kernel $K(x,x')$, where $K(x,x')$ is a continuous positive semi-definite function of variables $x,x'\in X\subset R^n$.
For the solution in RKHS with bias, we use the representation
\begin{equation}\label{Lgarage}
\begin{split}
f\left( x \right) =\sum_{i=1}^m{a_iK\left( x_i,x \right) +b},
\end{split}
\end{equation}
where $x_i$, $i = 1,...,m$ are vectors from the training set. For the regularization term, we have
\begin{equation}\label{Wf}
\begin{split}
W(f)=\lVert f\left( x \right) \rVert ^2=\sum_{i,j=1}^m{a_ia_jK\left( x_i,x_j \right)}.
\end{split}
\end{equation}
Therefore, finding the solution of \eqref{ExpRisk} in a given set of functions $\mathcal{F}$ requires the minimization of the functional
\begin{equation}\label{L1VSVM}
\begin{split}
\underset{f}{\min}\int{\ell_{\varepsilon}\left(\sum_{i=1}^m{G\left( \hat{x}-x_i \right) \left( f\left( x_i \right) -y_i \right)} \right)d\mu \left( \hat{x} \right)}+\lambda \lVert f\left( x \right) \rVert ^2.
\end{split}
\end{equation}

In fact, VSVM \citep{vapnik2019rethinking} is also could be considered as an approximation to \eqref{ExpRisk} by using the $L_2$ metric to transform the goal into minimizing the following unconstrained convex optimization problem
\begin{equation}\label{VSVM}
\begin{split}
\underset{f}{\min}\int{\left(\sum_{i=1}^{m}{G(\hat{x}-x_i)(f (x_i)-y_i)}\right)^2d\mu(\hat{x})}+\lambda \lVert f\left( x \right) \rVert ^2.
\end{split}
\end{equation}

\subsection{$\varepsilon$-$L_1$VSVM}\label{subsec34}
In this subsection, we discuss the solution of the optimization objective function Eq.\eqref{L1VSVM}. Since $G(\cdot)$ in \eqref{L1VSVM} is a non-negative function. The following inequality hold
\begin{equation}\label{AimLeq}
\begin{split}
& \int{\ell_{\varepsilon}\left(\sum_{i=1}^m{G\left( \hat{x}-x_i \right) \left( f\left( x_i \right) -y_i \right)} \right)d\mu \left( \hat{x} \right)}\\
& = \int{\ell_{\varepsilon}\left(\mid \sum_{i=1}^m{G\left( \hat{x}-x_i \right) \left( f\left( x_i \right) -y_i \right)} \mid \right)d\mu \left( \hat{x} \right)}\\
&\leq \int{\ell_{\varepsilon}\left( \sum_{i=1}^m {G\left( \hat{x}-x_i \right) \mid \left( f\left( x_i \right) -y_i \right)}\mid \right)d\mu \left( \hat{x} \right)}\\
& \underset{\varepsilon\rightarrow 0}{\leq}  \int{\sum_{i=1}^m G\left( \hat{x}-x_i \right)\ell_{\varepsilon}\left(  \mid  f\left( x_i \right) -y_i  \mid  \right)d\mu \left( \hat{x} \right)}\\
& =\sum_{i=1}^m{ \ell_{\varepsilon}\left( f\left( x_i \right) -y_i \right)\int{ G\left( \hat{x}-x_i \right) d\mu \left( \hat{x} \right)}}
\end{split}
\end{equation}

Therefore, under this inequality, minimize \eqref{L1VSVM} turns into
\begin{equation}\label{Aim}
\begin{split}
\underset{f}{\min} \sum_{i=1}^m{( \ell_{\varepsilon}\left( f\left( x_i \right) -y_i \right)\int{G\left( \hat{x}-x_i \right) d\mu \left( \hat{x} \right))}}+\lambda \lVert f\left( x \right) \rVert ^2.
\end{split}
\end{equation}
The advantage of minimizing \eqref{Aim} is that the variables $\hat{x}$ and $x_i$ are separable.
Thus, in practice, minimization \eqref{Aim} can be performed in two steps:
\begin{description}
\item[\textbf{Step 1:}]
In the space of $\hat{x}$, for each training sample $x_i$, compute
\begin{equation}\label{vi}
\begin{split}v_i=\int{ G\left( \hat{x}-x_i \right) }d\mu \left( \hat{x} \right).
\end{split}
\end{equation}

\item[\textbf{Step 2:}]
In the set of $\{f(x)\}$, minimize
\begin{equation}\label{Aim3}
\begin{split}
\underset{f}{\min}\sum_{i=1}^m{ \ell_{\varepsilon}\left( f\left( x_i \right) -y_i \right)v_i}+\lambda \lVert f\left( x \right) \rVert ^2.
\end{split}
\end{equation}
\end{description}

Note that $v_i$ is the integral over $\hat{x}$ for the training sample $x_i$ in the distribution, so $v_i$ can be obtained by the parameter or non-parameter estimation above, such as Example \ref{eg1} and Example \ref{eg2}.
In \eqref{vi}, $x_i$ is the training sample which has the corresponding label $y_i$, whereas $\hat{x}$ can be any $\hat{x}$ from the unknown distribution $P(x)$. From this point of view, use the priori probability or unlabeled samples to estimate a more accurate location information $v_i$ about the sample $x_i$ than only using training set is promising, and a relationship between the training set (labeled samples) and the test set (unlabeled samples) through $v$ method could be established.

The model \eqref{Aim3} is called $\varepsilon$-$L_1$VSVM as it is similar to a weighted SVR model, where the weights are $v$-values describing the location information of the training samples. Unlike traditional weighted models, the weights in \eqref{Aim3} is a kind of knowledge from distribution, adding $v$ to the model is similar to adding the geometric properties attributes of the data to the model. As discussed above, $v$ can construct a link between the training samples and the overall distribution to improve the model performance. In the experiments, we first use all the samples (including the test set) to construct $v$-values to manifest the location information of training set, and then use it for modeling.

To solve $\varepsilon$-$L_1$VSVM effectively, we rewrite \eqref{Aim3} as
\begin{equation}\label{L1VSVMeps}
\begin{split}
\underset{f}{\min}\frac{1}{2}\lVert f\left( x \right) \rVert ^2+\gamma \sum_{i=1}^m{\ell_{\varepsilon}(f\left( x_i \right) -y_i)v_i},
\end{split}
\end{equation}
where the specific form of $v_i$ can be chosen according to the data characteristics.
By introducing the slack variables, \eqref{L1VSVMeps} can be equivalently written as
\begin{equation}\label{QP}
\begin{split}
&\underset{f,\xi ^*,\xi}{\min}\gamma \sum_{i=1}^m{\left( \xi _i^*+\xi _i \right) v_i+\frac{1}{2}\sum_{i,j=1}^m{a_ia_j}K\left( x_i,x_j \right)},\\
&s.t\ -\varepsilon -\xi _i^*\le f\left( x_i \right) -y_i\le \varepsilon +\xi _i,~ \xi _i^*\ge 0,~ \xi _i\ge 0.\\
\end{split}
\end{equation}
By defining the following notations:
$a=(a_1,...,a_m)^T$,
$\alpha=(\alpha_1,...,\alpha_m)^T$,
$\alpha^*=(\alpha^*_1,...,\alpha^*_m)^T$,
$\xi=(\xi_1,...,\xi_m)^T$,
$\xi^*=(\xi^*_1,...,\xi^*_m)^T$,
$v=(v_1,...,v_m)^T$,
$l\times l$-dimensional matrix $K$ with $K_{ij}=K(x_i,x_j)$ being the $ij^{th}$ entry, $y=(y_1,y_2,...,y_m)^T$, $1_m=(1,1,...,1)^T$,
we consider the dual formulation of \eqref{QP}. The Lagrangian with $\mu_i\geq 0, \mu_i^*\geq 0, \alpha_i\geq 0, \alpha_i^*\geq 0$ of \eqref{QP} is
\begin{equation}\label{Lgarage}
\begin{split}
L( a,b,\xi ,\xi ^*,\alpha ,\mu ,\mu ^* ) =\frac{1}{2}a^TKa+\gamma (\xi^*+\xi) ^Tv-\mu ^T\xi -\mu ^{*T}\xi ^*+\alpha ^T\\
( Ka+b1_m-Y-\varepsilon1_m -\xi) +\alpha ^{*T}( Y-Ka-b1_m-\varepsilon1_m -\xi ^*).
\end{split}
\end{equation}
According to
\begin{equation}\label{Lgarage1}
\begin{split}
&\frac{\partial L}{\partial a}=Ka+K\alpha -K\alpha ^*=0,\\
&\frac{\partial L}{\partial b}=\alpha ^T1_m-\alpha ^{*T}1_m=0,\\
&\frac{\partial L}{\partial \xi}=-\mu +\gamma v-\alpha =0,\\
&\frac{\partial L}{\partial \xi ^*}=-\mu ^*+\gamma v-\alpha ^*=0,\\
\end{split}
\end{equation}
we obtain the dual problem of \eqref{QP} as follows
\begin{equation}\label{Lgarage2}
\begin{split}
&\underset{\alpha ,\alpha ^*}{\max}\ \left( \alpha ^*-\alpha \right) ^TY-\varepsilon \left( \alpha ^*+\alpha \right) ^T1_m -\frac{1}{2}\left( \alpha ^*-\alpha \right) ^TK\left( \alpha ^*-\alpha \right)\\
&s.t.\ \left( \alpha ^*-\alpha \right) ^T1_m=0,\\
&\ \ \ \ \ \ 0\le \alpha ,\alpha ^*\le \gamma v.
\end{split}
\end{equation}

Since the $\varepsilon$-insensitive loss is sparse and the dual problem \eqref{Lgarage2} has sparse solutions, we have the following lemma.
\begin{lemma}
(Lemma about inverse operator.) $\alpha_i$ takes a non-zero value only if $f(x_i) -y_i-\varepsilon -\xi _i =0$ and $\alpha_i^*$ takes a non-zero value only if $f(x_i) -y_i-\varepsilon -\xi _i^* =0$. In addition, constraints $f(x_i) -y_i-\varepsilon -\xi _i =0$ and $f(x_i) -y_i-\varepsilon -\xi _i^* =0$ cannot be held at the same time, so at least one of $\alpha_i$ and $\alpha_i^*$ is zero.
\label{lem-1}
\end{lemma}
From the Karush-Kuhn-Tucker \citep{mangasarian1994nonlinear} conditions
$$
\left\{ \begin{array}{l}
	\alpha _i\left( f\left( x_i \right) -y_i-\varepsilon -\xi _i \right) =0,\\
	\alpha _{i}^{*}\left( f\left( x_i \right) -y_i-\varepsilon -\xi _i^* \right) =0,\\
	\alpha _i\alpha _i^*=0,\xi _i\xi _{i}^{*}=0,\\
	\left( \gamma v_i-\alpha _i \right) \xi _i=0,\left( \gamma v_i-\alpha _{i}^{*} \right) \xi _{i}^{*}=0,\\
\end{array} \right.
$$
we have the conclusion that $\alpha_i$ and $\alpha_i^*$ takes a non-zero value only if the sample $(x_i, y_i)$ does not fall into the $\varepsilon$-insensitive zone, and at least one of $\alpha_i$ and $\alpha_i^*$ is zero. That is to say, the solutions of $\varepsilon$-$L_1$VSVM is sparse.

\subsection{Evaluation}\label{subsec35}
In this subsection, we discuss the indicator and the evaluation of the classification. Typically, the generalisation error of a classifier cannot be evaluated directly but estimated by some testing error \citep{theodoridis2006pattern}.
The 'testing error' on the test set is then used as an approximation to the generalisation error. Assuming that the probability of test samples is equal or unknown, we usually use the error rate or accuracy rate to evaluate the classifier.
Given a testing set $\{(x_t,y_t\}_{t=1}^{T}$, the accuracy rate is defined as follows
\begin{equation}\label{Acc}
\begin{split}
 Acc_{test}=\frac{1}{T}\sum_{t=1}^{T}{\mathbb{I}(y_t= r(x_t))},
 \end{split}
\end{equation}
where
$$\mathbb{I}(y_t= r(x_t)) =\left\{ \begin{array}{l}
	1\text{,\,\,}if\,\,y_t= r(x_t)\text{,}\\
	0\text{,\,\,\,}otherwise.\\
\end{array} \right.$$
As for the expected risk \eqref{ExpRisk}, it estimates the relative position and distribution of samples based on the Fredholm equation, which requires that for each sample $x_t$, satisfying $r(x_t) = y_t$ in the sense of distribution (That is, the integral equation is satisfied). Therefore, it is reasonable to construct the corresponding evaluation indicator considering the distribution than to use the classical one (the probability of test samples is equal or unknown).

Based on the above proposed $v$-vector, for testing samples, we define the testing accuracy by considering location information of testing samples as
\begin{equation}\label{Vac}
\begin{split}
 Vac_{test}&=\frac{1}{T}\sum_{t=1}^{T}{\mathbb{I}(y_t= r(x_t))}\int_{\hat{\mathcal {X}}}{G(\hat{x}-x_t)d\mu(\hat{x})}
 \\&=\frac{1}{T}\sum_{t=1}^{T}{\mathbb{I}(y_t= r(x_t))}v_t.
 \end{split}
\end{equation}
Clearly, when the data distribution is uniform or not considered, we have
\begin{equation}\label{mux}
\begin{split}
\mu(\hat{x})=\theta(\hat{x}-x_t),
\end{split}
\end{equation}
and
$$
v_t=\int_{\hat{\mathcal {X}}}{G(\hat{x}-x_t)d\mu(\hat{x})}=G(x_t-x_t)=1.
$$
At this time, \eqref{Vac} degenerates into classical accuracy \eqref{Acc}.

Traditionally, the data distribution is non-uniform and we need to estimate the $v_t$ out-of-sample.  As discussed above,
$v_t$ describes the position of each sample in the distribution, corresponding to the samples one by one. Now, we show that $v_t$ also could be estimate out-of-sample.
Similarly, two types of $v_t$ are given, depending on whether the expression contains parameters.
\begin{description}
\item[\textbf{1)($v_t$ without parameter)}]
\begin{example}
Let $G(\hat{x}-x_t)=\theta(\hat{x}-x_t)=\left\{ \begin{array}{l}
	1,\ if\ \hat{x}\ge x_t\\
	0,\ otherwise\\
\end{array} \right.$. Consider one-dimensional case of $\hat{x}\in (-\infty,\infty)$, where $\hat{x}$ follows a normal distribution $(a,b^2)$(i.e.,$\mu(\hat{x})=\frac{1}{\sqrt{2\pi}b}exp\{-\frac{(x-a)^2}{2b^2}\}$. We have
\begin{equation}\label{v1}
\begin{split}
v_t=\int_{-\infty}^{x_t}{\frac{1}{\sqrt{2\pi}b}exp\{-\frac{(t-a)^2}{2b^2}\}dt},
\end{split}
\end{equation}
where $C$ is a constant.
For multidimensional case of $\hat{x}$, the elements $v_t$ of v-vector have the form
$$
v_t=\prod_{k=1}^d\left(\int_{-\infty}^{x^k_t}{\frac{1}{\sqrt{2\pi}b}exp\{-\frac{(t-a)^2}{2b^2}\}dt}\right).
$$
\label{eg1}
\end{example}

\item[\textbf{2)($v_t$ with parameter)}]
\begin{example}
Let $G(\hat{x}-x_t)=exp\{-\frac{(\hat{x}-x_t)^2}{2\sigma^2}\}$. Consider one-dimensional case of $\hat{x}\in (-a,a)$, where $\hat{x}$ is uniformly distributed on $(-a,a)$(i.e.,$\mu(\hat{x})=\frac{\hat{x}-a}{2a})$. We have
$$
v_t=-\frac{1}{2a}\int_{-a}^{a}{exp\{-\frac{(\hat{x}-x_t)^2}{2\sigma^2}\}d(\hat{x})}.
$$
After integration, we obtain
\begin{equation}\label{v2}
\begin{split}
v_t=C\left(erf\left(\frac{a+x_t}{\sigma}\right)+erf\left(\frac{a-x_t}{\sigma}\right)\right),
\end{split}
\end{equation}
where $C$ is a constant.
For multidimensional case of $\hat{x}=(\hat{x}^1,\hat{x}^2,...,\hat{x}^d)\in [-a_1,a_1]\times ...\times[-a_d,a_d]$, the elements $v_t$ of v-vector have the form
$$
v_t=C \prod_{k=1}^d{\left(erf\left(\frac{a_k+x_t^k}{\sigma}\right)+erf\left(\frac{a_k-x_t^k}{\sigma}\right)\right)}.
$$
\label{eg1}
\end{example}
\end{description}
\textbf{Algorithmic Remark}:
In high dimensions of the v-vector stated in \eqref{vi} and V-matrix stated in \eqref{Vmatrix} may be ill-conditioned \citep{vapnik2019rethinking,mazaheri2020robust}. So, it suggests using an addictive version of the v-vector. This version derives from considering each dimension of the empirical distribution function separately and then minimizing the sum of the $L_1$ or $L_2$ errors. Thus, v-vector becomes
$$
\hat{v_i}=\int \left( \sum_{k=1}^{d}G(\hat{x}^k-x_i^k) \right)d\mu(\hat{x}).
$$
It is worth noting that $x_t$ in the expression of $v$ represents the test samples, and $\hat{x}$ can be broader in scope to include all samples (e.g., training set and testing set). This leads the position of the samples in the whole dataset, including the training samples and test samples. Obviously, for a given dataset, if $v$-vector without parameter is used, then the calculated $v$ is fixed, and the results of evaluation indicator can be compared directly between different models.
However, if a parameterized $v$-vector is used, the value of $v$ vary with the selection of parameter $\sigma$, and the values of evaluation indexes of different models are not comparable.
Therefore, we only use the above evaluation indicator in the cross-validation selection process in the experiments.
In addition, from the perspective of knowledge, for other models that do not consider data distribution information, using the above  proposed evaluation indicator is another way to adding the priori knowledge of data distribution information to the modeling (model selection), thus improving the performance of the model.

Compared to the expression \eqref{Acc} and \eqref{Vac}, in \eqref{Acc}, the weight given to each test sample in the testing set is 1, i.e., each test sample is treated equally, without considering the location information of the samples. Whereas, in \eqref{Vac}, it computes not only the residuals of the true and estimated values, but also information about the distribution of each sample.

\section{Numerical Experiments}\label{sec4}
In this section, we carry out all our numerical experiments to compare the proposed method and improved evaluation indicator with other five related SVM classifiers. Firstly, we show the details of the experimental setup. Then, we present an empirical analysis based on two synthetic datasets, where the optimal Bayes classifier is known. At last, we compare all classifiers over various widely investigated real-life small datasets. All experiments are implemented by using Matlab R2020a on a desktop with 8 $Intel^{^\circledR}Core^{TM}$ i7-7700K processors (4.20 GHz)and 32GB RAM.

\subsection{Experimental setup}\label{sec41}
Five related SVM-type classifiers are used for experimental comparison with the proposed $\varepsilon$-$L_{1}$VSVM
$\footnote{codes available at: \url{http://www.optimal-group.org/Resources/Code/L1VSVM.html}}$.
\begin{itemize}
    \item CSVM\citep{chang2011libsvm}: CSVM is the classical SVM with hinge loss which maximizes the margin between the nearest points of different classes.
    \item LSSVM\citep{pelckmans2002ls}: LSSVM is a least squares version of the support vector machine, which derives its solution from solving a set of linear equations.
    \item IDLSSVM\citep{hazarika2021density}: IDLSSVM (improved density weighted LSSVM) is an improved 2-norm-based density-weighted least squares SVM for binary classification which considers the density of each data sample in its class.
    \item VSVM\citep{vapnik2019rethinking}: VSVM is a model constructed based on the Fredholm integral equation and the least squares loss that takes into account mutual positions of the data.
    \item $\varepsilon$-$L_1$SVM: When $v-$vector is an all-1 vector, the proposed $\varepsilon$-$L_{1}$VSVM recover to  this model, which can be used for classification problems and is denoted as $\varepsilon$-$L_1$SVM. Both VSVM and $\varepsilon$-$L_1$SVM can be used to compare the performance of v-vectors of $\varepsilon$-$L_1$SVM.
\end{itemize}

In experiment, the standard 10-fold cross-validation with grid search strategy is conducted to tune the parameters associated with each classifier.
The tradeoff parameters $C$ is selected from $\{2^{-8},2^{-7},...,2^8\}$. For both VSVM and $\varepsilon$-$L_1$VSVM, the distribution information $v$ value is calculated, the parameter $\sigma$ in Gaussian kernel function $G(\cdot)$ is choosing from $\{2^{-4},2^{-3},...,2^4\}$ and the parameter $\sigma$ = 0 while $G(\cdot)$ is \eqref{theta}. For $\varepsilon$-$L_1$SVM and $\varepsilon$-$L_1$VSVM, the parameter $\varepsilon$ is selected from $\{2^{-4},2^{-3},2^{-2}\}$. For IDLSSVM, the parameter k is fixed to 5 as mentioned in \citep{hazarika2021density}. As for kernel setting, the radius basic function (rbf) kernel
  $$
  K_{rbf}=exp(-\frac{\Vert x-x'\Vert^2}{2\delta^2})
  $$
is used and the parameter is selected from $\Delta :=\{2^{-4},2^{-3},...,2^4\}$.

To evaluate the classification performance, two commonly employed metrics are applied: accuracy and G-means. 
\begin{itemize}
    \item Accuracy represents the proportion of correctly classified instances, i.e., Accuracy(Acc for simplicity) = $\frac{TP+TN}{TP+FN+FP+TN}$.
    \item Sensitivity (also referred as recall) explains the prediction accuracy among positive instances, i.e., Sensitivity = Recall=$\frac{TP}{TP+FN}$.
    \item Specificity represents the prediction accuracy among negative instances, i.e., Specificity = $\frac{TP}{TP+FN}$.
    \item G-mean is defined as the geometric mean of sensitivity and specificity, i.e., G-mean= $\sqrt{Sensitivity\times Specificity}$.
\end{itemize}
Since we need to compare Acc and Vac, Bayes decision is the main indicator of artificial data experiments and G-mean is the main indicator of benchmark data experiments.

\subsection{Experiments on synthetic datasets}\label{subsec41}
\textbf{Experiment 1: Validity of model and indicator.}
For visualization, we first consider a two-dimensional example, for which the features are simulated based on a Gaussian bivariate model and labels are generated with equal probability.
 We generate $N=\{100,200,300\}$ random variate from Gaussian distribution, where $N$ is the total number of the examples from the two classes $y_i \in \{0,1\}$. Therefore, the training data $\{(x_i,y_i),i=1,...,N\}$ are simulated as follows:
\begin{equation}\label{Gaussian}
\begin{split}
X_i \vert y=1\sim N\left( \mu ,\Sigma \right) \ and \ X_i \vert y=-1\sim N\left( -\mu ,\Sigma \right),
\end{split}
\end{equation}
where $\mu=[1,-2]^T$ and $\Sigma=diag(0.5,2)$. The data generated in this way has centralized features of each class, and the theoretical optimal linear classification boundary (Bayes classifier) of this example is known as
\begin{equation}\label{Bayes}
\begin{split}
x_2=k_0x_1+q_0,\ where\ k_0=2\ and \ q_0=0.
\end{split}
\end{equation}
Our goal is to estimate $k$ and $q$ for each classifier and compare them with the Bayes classifier \eqref{Bayes}.
The six classifiers are compared via repeating the sampling and training 100 times of sample size $N$ for which the linear decision rule is computed, i.e., $(k_i,q_i)$ for all $1\le i\le 100$.

Each classifier is fairly compared against the Bayes classifier via the mean and variance of the slope $\bar k$ and the intercept $\bar q$. Further, to show the differences in the models more clearly, we additionally calculated the distance
\begin{equation}\label{Dist}
\begin{split}
Dist_j=\vert \bar k_j-k_0\vert \hat{\sigma}_{k_j}+\vert \bar q_j-k_0\vert \hat{\sigma}_{q_j}
\end{split}
\end{equation}
for $j\in\{$C-SVM, LSSVM, VSVM, IDLSSVM, $\varepsilon$-$L_{1}$SVM, $\varepsilon$-$L_{1}$VSVM$\}$ to Bayes classifier, where $\bar k_j(\bar q_j)$ and $\hat{\sigma}_{k_j}(\hat{\sigma}_{q_j})$ are the mean and standard deviation estimates of $k(q)$ based on the 100 estimates of $k(q)$. We set v-vector in Table \ref{thetanorm} to version \eqref{v1} and in Table \ref{gussuni} to version \eqref{v2} and the V-matrix is set the same way, the purpose of which is to show the characteristics and advantages of different versions of v-vector and V-matrix.

The results are presented in Tables \ref{thetanorm} and Table \ref{gussuni}, where Bayesian classifier is the baseline and is noted on the caption of the table.
\begin{table}[h]
\centering
\caption{Classification boundaries and distance based on Acc \eqref{Acc} and Vac \eqref{Vac} as evaluation indicators and v-vector in \eqref{v1} for synthetic data. The first row of values in the table is \eqref{Dist}, and the result of Bayes is 0; the second row slope $k$, and the result of Bayes is 2; the third row is the intercept $p$, and the result of Bayes is 0.}
\label{thetanorm}
\resizebox{\textwidth}{!}{
\setlength{\tabcolsep}{0.9mm}{
\begin{tabular}{l|l|l|l|l|l|l}
\toprule
N         & \multicolumn{1}{l}{100} &                 & \multicolumn{1}{l}{200} &                 & \multicolumn{1}{l}{300} &                  \\
\hline
Indicator  & Acc                      & Vac            & Acc                      & Vac            & Acc                     & Vac              \\
\hline
CSVM       & 0.4560                    & 0.2677          & 0.1960                   & 0.1759          & 0.1806              & 0.1191           \\
           & 1.43$\pm$0.80          & 1.70$\pm$0.85  & 1.66$\pm$0.57          & 1.76$\pm$0.69          & 1.65$\pm$0.51          & 1.78$\pm$0.52   \\
           & -0.0001$\pm$0.32        & -0.03$\pm$0.35 & 0.004$\pm$0.25          & -0.04$\pm$0.26     & 0.002$\pm$0.21          & -0.21$\pm$0.12  \\
\hline
LSSVM      & 0.3548                  & 0.1839          & 0.1958                  & 0.1253          & 0.1562                  & 0.0946           \\
           & 1.35$\pm$0.54          & 1.73$\pm$0.65  & 1.53$\pm$0.41          & 1.70$\pm$0.41  & 1.63$\pm$0.42          & 1.76$\pm$0.40   \\
          & 0.02$\pm$0.17         & 0.03$\pm$0.19 & 0.026$\pm$0.13          & 0.005$\pm$0.13  & -0.01$\pm$0.11         & 0.009$\pm$0.12   \\
\hline
VSVM      & 0.0892                  & 0.0836          & 0.0759                  & 0.0433     & \textbf{0.0089}                  & \textbf{0.0057}          \\
           & 1.86$\pm$0.64         & 1.87$\pm$0.58
           & 1.82$\pm$0.40          & 1.90$\pm$0.42
           & \textbf{2.03$\pm$0.30}          & \textbf{2.02$\pm$0.32}   \\
         & 0.01$\pm$0.21         & -0.05$\pm$0.20
           & -0.02$\pm$0.13         & 0.02$\pm$0.15
           & \textbf{-0.007$\pm$0.12}         & \textbf{0.008$\pm$0.12}   \\
\hline
IDLSSVM  & 0.2741        & 0.1398        & 0.1302         & 0.0982        & 0.0890          & 0.0479         \\
        & 1.60$\pm$0.68 & 1.73$\pm$0.48 & 1.73$\pm$0.48  & 1.78$\pm$0.44 & 1.7510$\pm$0.35  & 1.85$\pm$0.3097  \\
        & 0.01$\pm$0.21 & 0.04$\pm$0.21 & -0.009$\pm$0.22 & 0.02$\pm$0.17 & -0.01$\pm$0.12 & 0.002$\pm$0.12\\
\hline
$\varepsilon$-$L_{1}$SVM  & 0.2812                  & 0.0944          & 0.1140                   & 0.0429          & 0.0735                  & 0.0312         \\
          & 1.57$\pm$0.65          & 1.90$\pm$0.89
           & 1.84$\pm$0.69          & 2.04$\pm$0.93
           & 1.87$\pm$0.55          & 1.94$\pm$0.46   \\
       & 0.01$\pm$0.20          & 0.03$\pm$0.22
           & 0.01$\pm$0.18          & -0.04$\pm$0.16
           & -0.01$\pm$0.14         & -0.02$\pm$0.13  \\
\hline
$\varepsilon$-$L_{1}$VSVM & \textbf{0.0872}                  & \textbf{0.0645}          & \textbf{0.0509}                  & \textbf{0.0294}          & 0.0480~                 & 0.0263           \\
         & \textbf{1.88$\pm$0.72}          & \textbf{1.93$\pm$0.81}  & \textbf{1.93$\pm$0.70}          & \textbf{1.95$\pm$0.59}  & 1.90$\pm$0.48          & 1.96$\pm$0.68   \\
          & \textbf{0.02$\pm$0.20}          & \textbf{-0.02$\pm$0.21} & \textbf{0.006$\pm$0.15}          & \textbf{0.01$\pm$0.18}  & -0.01$\pm$0.12         & -0.014$\pm$0.12   \\
\bottomrule
\end{tabular}
}}
\end{table}

\begin{center}
\begin{table}[h]
\centering
\caption{Classification boundaries and distance based on Acc \eqref{Acc} and Vac \eqref{Vac} as evaluation indicators and v-vector in \eqref{v2} for synthetic data. The first row of values in the table is \eqref{Dist}, and the result of Bayes is 0; the second row slope $k$, and the result of Bayes is 2; the third row is the intercept $p$, and the result of Bayes is 0.}
\label{gussuni}
\resizebox{\textwidth}{!}{
\setlength{\tabcolsep}{0.9mm}{
\begin{tabular}{l|l|l|l|l|l|l}
\toprule
N        & \multicolumn{1}{l}{100} &                 & \multicolumn{1}{l}{200} &                 & \multicolumn{1}{l}{300} &                  \\
\hline
Indicator & Acc             & Vac            & Acc              & Vac            & Acc              & Vac             \\
\hline
CSVM      & 0.4560         & 0.2717         & 0.1960~         & 0.1807          & 0.1806          & 0.1408          \\
          & 1.43$\pm$0.80  & 1.72$\pm$0.95 & 1.6583$\pm$0.57  & 1.76$\pm$0.72    & 1.65$\pm$0.51  & 1.75$\pm$0.55  \\
          & -0.0001$\pm$0.32 & 0.005$\pm$0.33 & 0.004$\pm$0.25  & -0.02$\pm$0.21   & 0.002$\pm$0.21  & 0.01$\pm$0.24  \\
\hline
LSSVM     & 0.3548          & 0.2420       & 0.1958          & 0.1753          & 0.1562          & 0.1036          \\
          & 1.35$\pm$0.54  & 1.64$\pm$0.66& 1.53$\pm$460.41  & 1.65$\pm$0.50  & 1.63$\pm$0.42  & 1.73$\pm$0.38   \\
          & 0.02$\pm$0.17  & 0.02$\pm$0.22 & 0.03$\pm$0.13  & -0.01$\pm$0. 14 & -0.01$\pm$0.11 & -0.0001$\pm$0.11  \\
\hline
VSVM      & 0.3236          & 0.2593         & 0.2138          & 0.2275          & 0.1238          & 0.1956         \\
          & 1.54$\pm$0.70  & 1.47$\pm$0.48 & 1.67$\pm$0.64  & 1.23$\pm$0.30  & 1.75$\pm$0.49  & 1.11$\pm$0.22  \\
          & -0.008$\pm$0.19 & -0.02$\pm$0.1753 & -0.0004$\pm$0.14 & 0.01$\pm$0. 23  & -0.01$\pm$0.11 & -0.01$\pm$0.11   \\
\hline
IDLSSVM   &  0.2741       & 0.1726         & 0.1302         & 0.0842         & 0.0890          & 0.0612         \\
        & 1.60$\pm$0.68 & 1.64$\pm$0.48  & 1.73$\pm$0.48  & 1.76$\pm$0.35 & 1.75$\pm$0.35  & 1.83$\pm$0.35  \\
        & 0.01$\pm$0.21 & -0.009$\pm$0.16 & -0.009$\pm$0.22 & -0.01$\pm$0.12 & -0.01$\pm$0.12 & 0.007$\pm$0.16 \\
\hline
$\varepsilon$-$L_{1}$SVM     & 0.2812          & 0.1647        & 0.1140           & 0.1060           & 0.0735          & 0.0462          \\
          & 1.57$\pm$0.65  & 1.74$\pm$0.63 & 1.84$\pm$0.69  & 1.83$\pm$0.60  & 1.87$\pm$0.55  & 1.92$\pm$0.57  \\
          & 0.01$\pm$0.20  & 0.01$\pm$0.14 & 0.01$\pm$0.18  & -0.01$\pm$0.15 & -0.01$\pm$0.14 & 0.02$\pm$0.13  \\
\hline
$\varepsilon$-$L_{1}$VSVM    & \textbf{0.1293}          & \textbf{0.1100}         & \textbf{0.0945}           & \textbf{0.0576}          & \textbf{0.0370}           & \textbf{0.0035}          \\
          & \textbf{1.73$\pm$0.45}  & \textbf{1.85$\pm$0.72} & \textbf{1.82$\pm$0.79}  & \textbf{1.92$\pm$0.71}  & \textbf{1.94$\pm$0.66}  & \textbf{2.00$\pm$0.61}  \\
          & \textbf{-0.03$\pm$0.26} & \textbf{-0.007$\pm$0.15} & \textbf{-0.02$\pm$0.15} & \textbf{0.0001$\pm$0.17}  & \textbf{-0.004$\pm$0.12} & \textbf{0.009$\pm$0.12}  \\
\bottomrule
\end{tabular}
}}
\end{table}
\end{center}

From the experimental results in Table \ref{thetanorm} and \ref{gussuni}, we yield the following observations:
\begin{itemize}
\item[i)]  We compare the performance of these six classifiers, using Bayesian decision as a baseline, and it is clear that with 12 results in these two tables, $\varepsilon$-$L_{1}$VSVM has the best results in 10 out of 12 and is shown in bold. When comparing LSSVM with VSVM, VSVM is almost always better than LSSVM, and for $\varepsilon$-$L_{1}$SVM and $\varepsilon$-$L_{1}$VSVM, $\varepsilon$-$L_{1}$VSVM is also closer to the optimal decision.
\item[ii)] The amount of artificial data in the two tables are increasing in the order of \{100,200,300\}, and it can be found that the performance of each model is improving as the amount of data increases. It is worth noting that when N=100 (small datasets), the performance of $\varepsilon$-$L_{1}$VSVM and VSVM with v-value is more prominent than others.
\item[iii)] Taking Table \ref{thetanorm} as an example to evaluate the performance difference between Vac and Acc, we find that the results of columns 2, 4, 6 are better than those of columns 1, 3, 5, which verifies the validity of Vac. Especially, the performance improvement for CSVM, LSSVM, IDLSSVM and $\varepsilon$-$L_{1}$SVM are particularly significant compared to VSVM and $\varepsilon$-$L_{1}$VSVM.
\item[iv)]We compare the two tables and find that the results in Table \ref{thetanorm} are better compared to Table \ref{gussuni}, especially the VSVM and our model $\varepsilon$-$L_{1}$VSVM have outstanding performance. Since the difference between the two tables is that the kernel functions $G(\cdot)$ and $\mu(x)$ used for the calculation of v-vector and V-matrix are not the same and therefore capture the distribution information in different ways. It is obvious that \eqref{v1} is more suitable for the artificial dataset of Gaussian distribution. The VSVM is better for N=300 in Table \ref{thetanorm} can be attributed to the fact that the estimation of v value is more accurate at this time and the V-matrix is a matrix compared to the v-vector in our model to get more comprehensive information.
\end{itemize}

\begin{figure*}[h]
\setlength{\abovecaptionskip}{0.cm}
\setlength{\belowcaptionskip}{-0.cm}
\centering
    \subfigure[C-SVM]{\includegraphics[width=0.25\textheight]{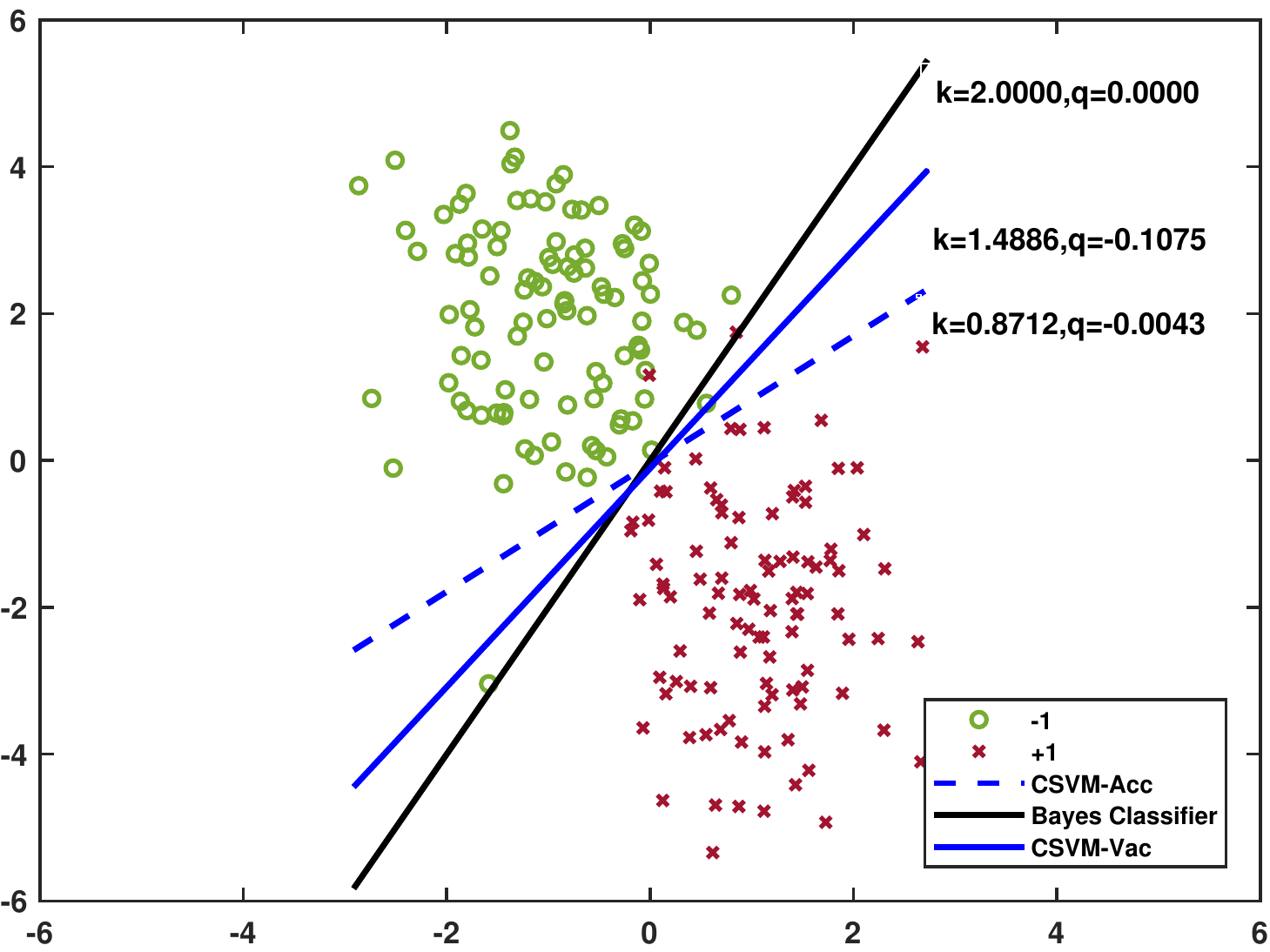}}
    \subfigure[LSSVM]{\includegraphics[width=0.25\textheight]{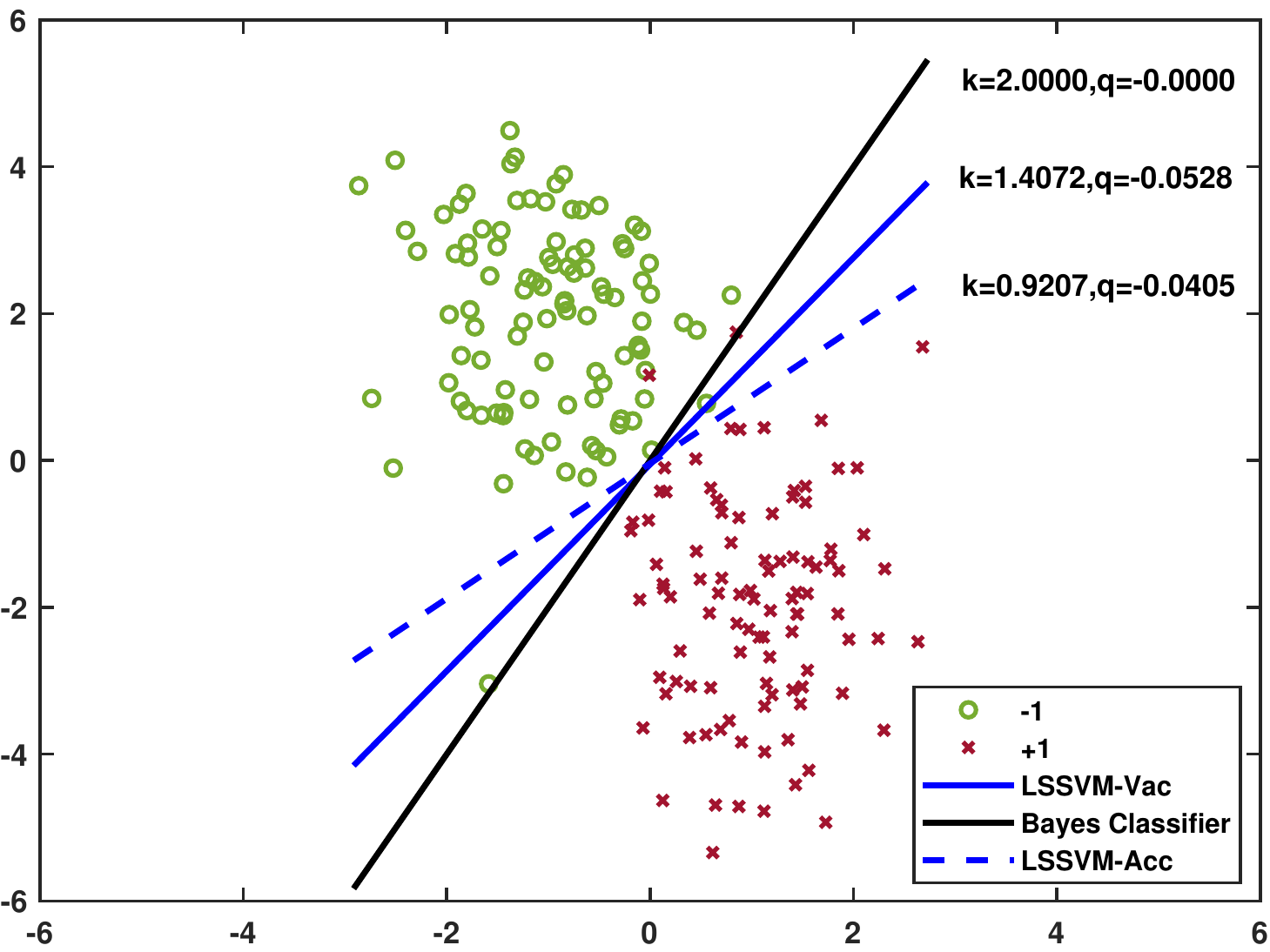}}
    \subfigure[VSVM]{\includegraphics[width=0.25\textheight]{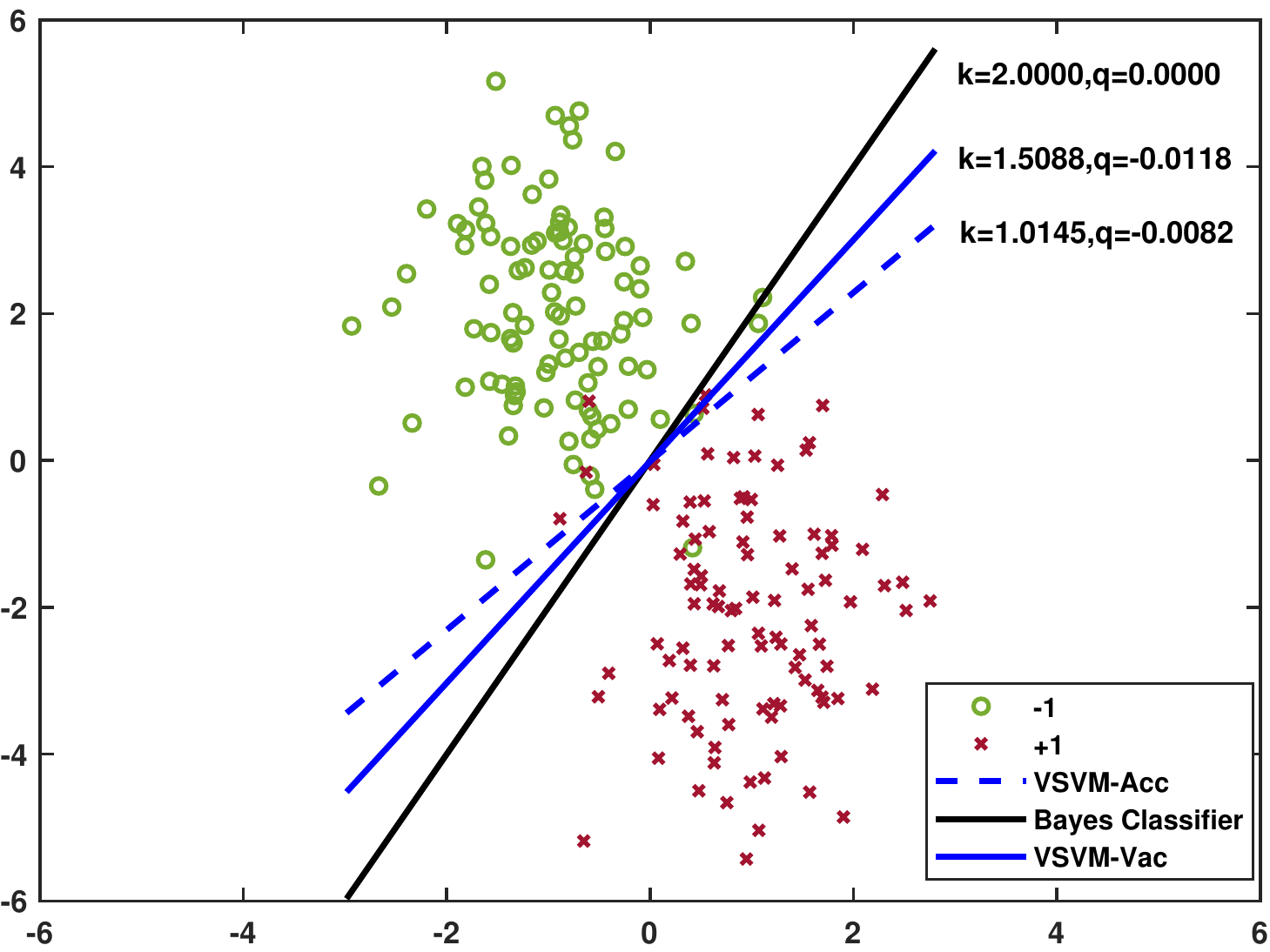}}
    \subfigure[IDLSSVM]{\includegraphics[width=0.25\textheight]{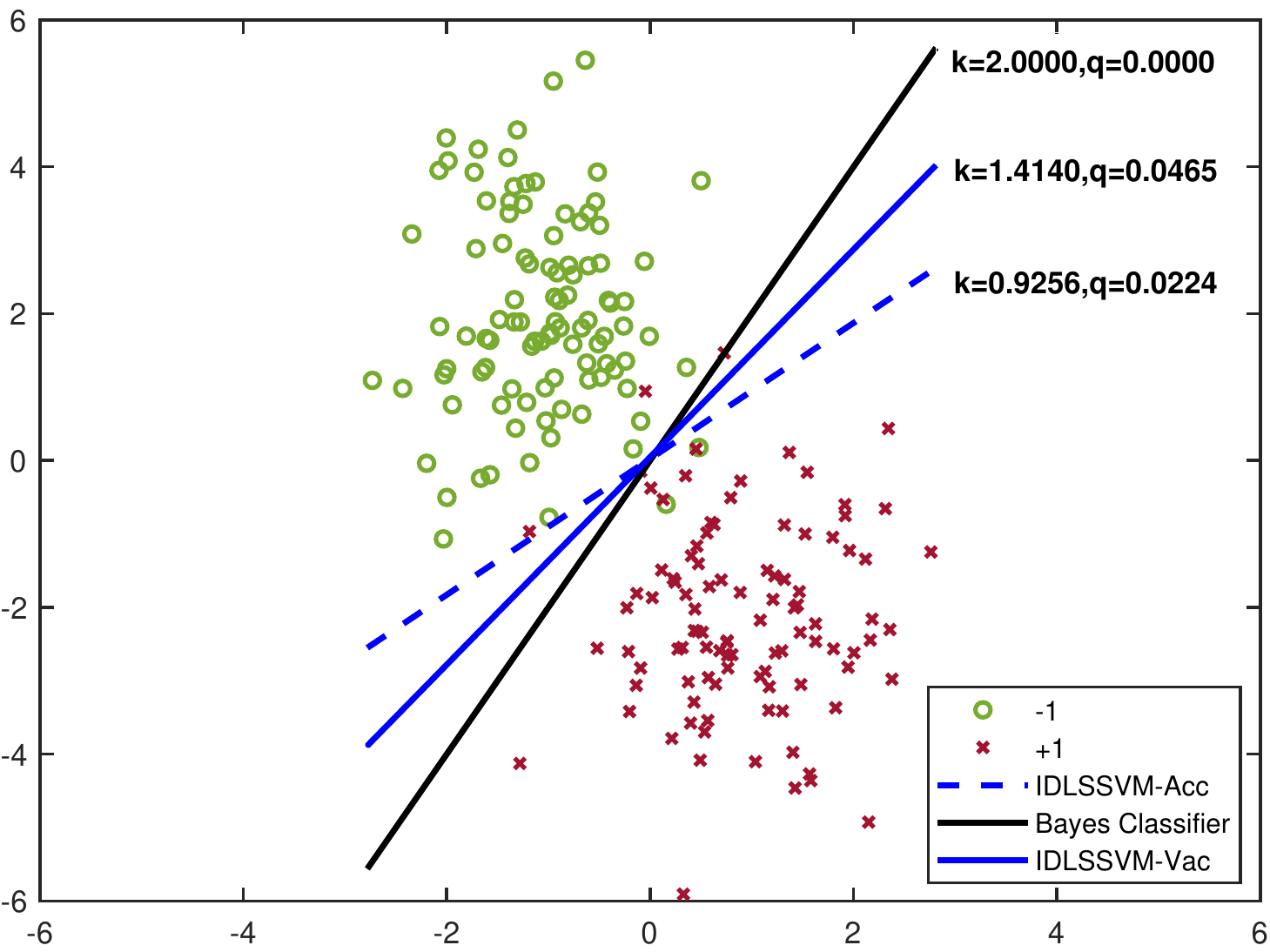}}
    \subfigure[$\varepsilon$-$L_1$SVM]{\includegraphics[width=0.25\textheight]{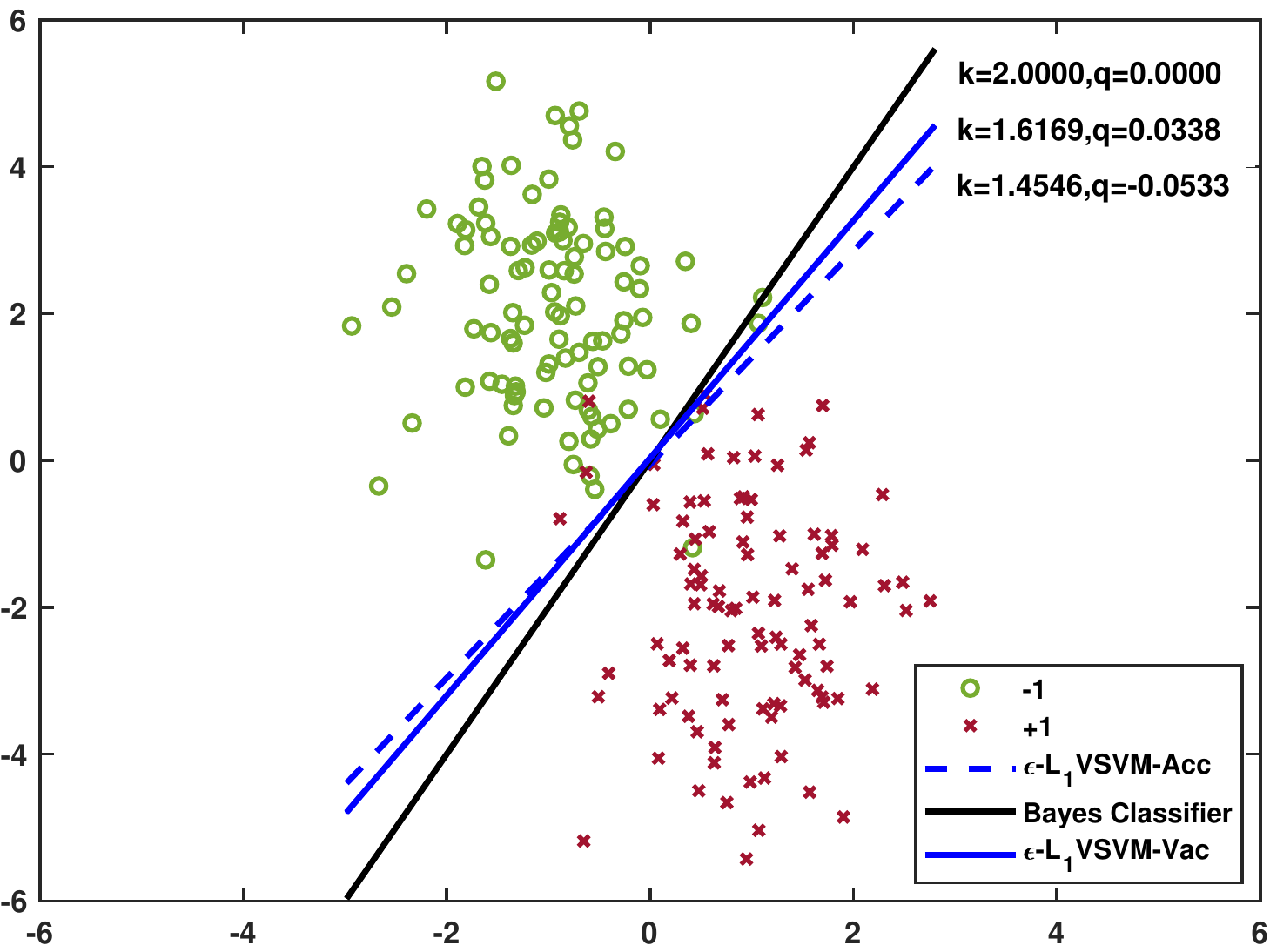}}
    \subfigure[$\varepsilon$-$L_1$VSVM]{\includegraphics[width=0.25\textheight]{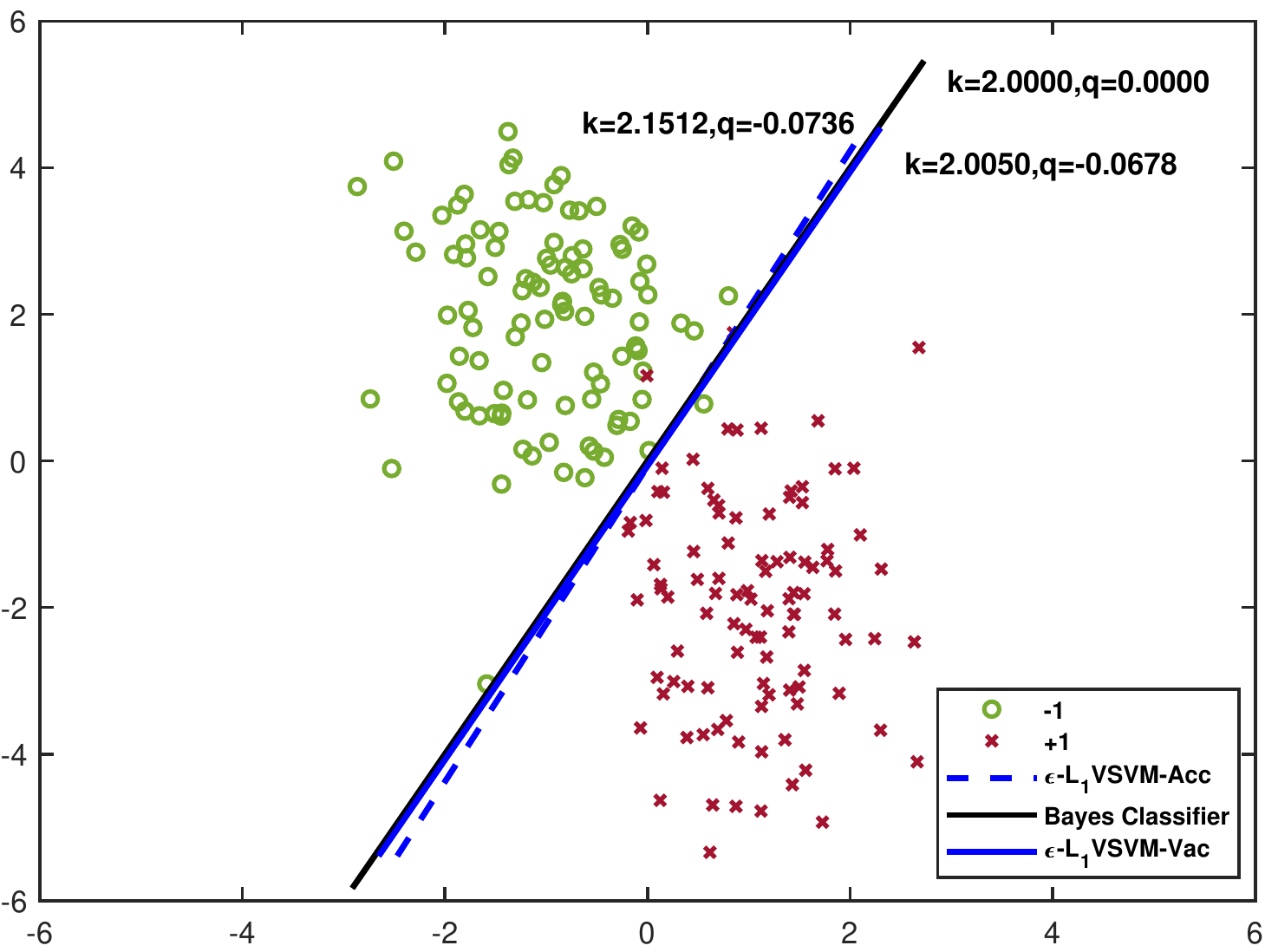}}
\caption{Classification boundaries based on Acc \eqref{Acc} and Vac \eqref{Vac} as evaluation indicators for six classifiers with the v-vector of \eqref{v2}.}
\label{bayes}
\end{figure*}

To show more clearly the effect of Vac compared to Acc, Figure \ref{bayes} shows the linear decisions based on Acc and Vac for the above six models as well as the Bayesian optimal decisions for Table \ref{gussuni}. The green and red dots in the figure are the positive and negative class samples, the black line is the Bayesian optimal decision, the blue solid line is the decision using Vac as the evaluation indicator, and the blue dashed line is for Acc. Clearly, the model with Vac as the indicator performs better and our proposed model is the closest to the Bayesian decision among the six models in both cases.


\begin{figure*}[h]
\setlength{\abovecaptionskip}{0.cm}
\setlength{\belowcaptionskip}{-0.cm}
\centering
    \subfigure[C-SVM]{\includegraphics[width=0.25\textheight]{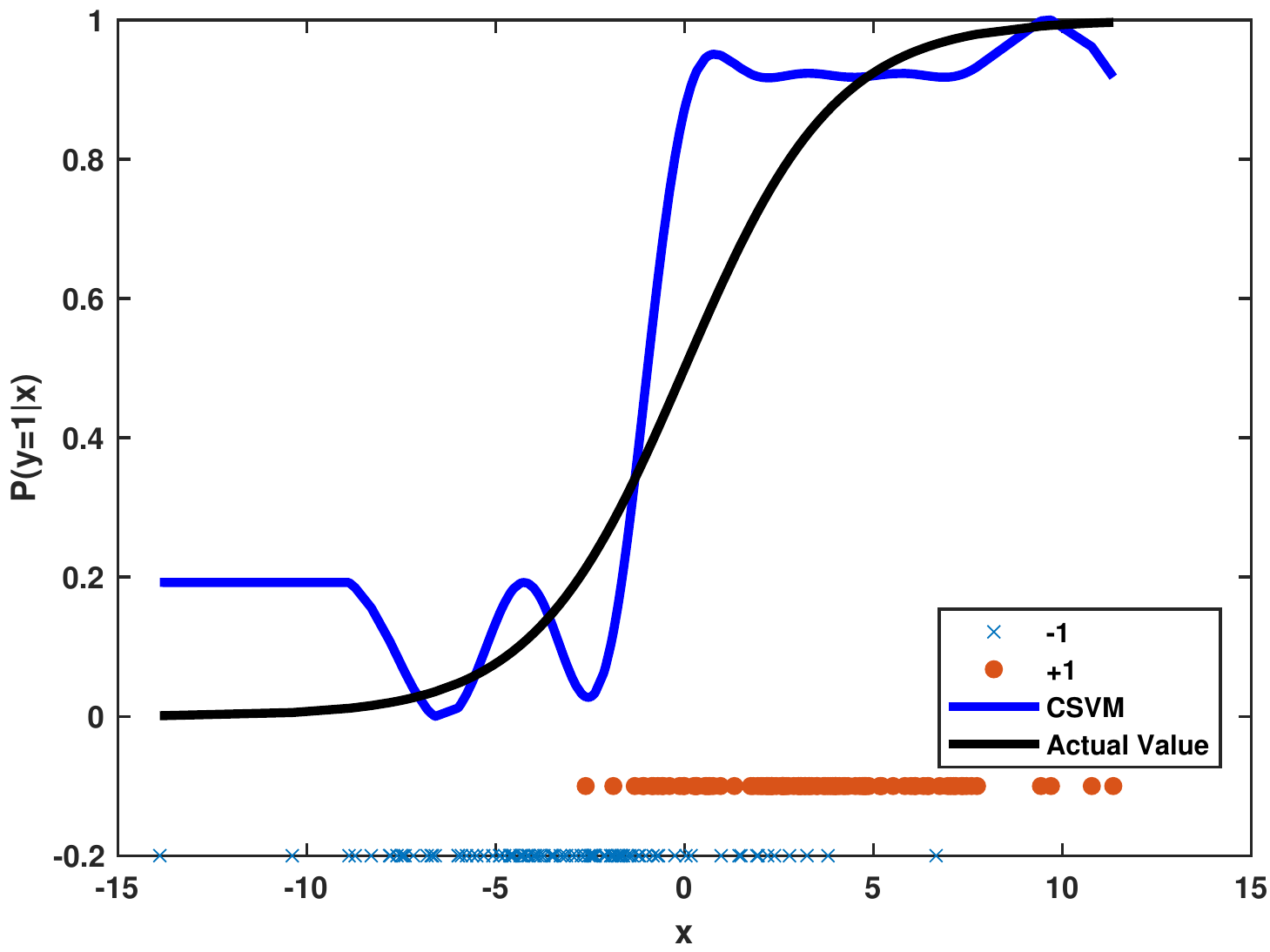}}
    \subfigure[LSSVM]{\includegraphics[width=0.25\textheight]{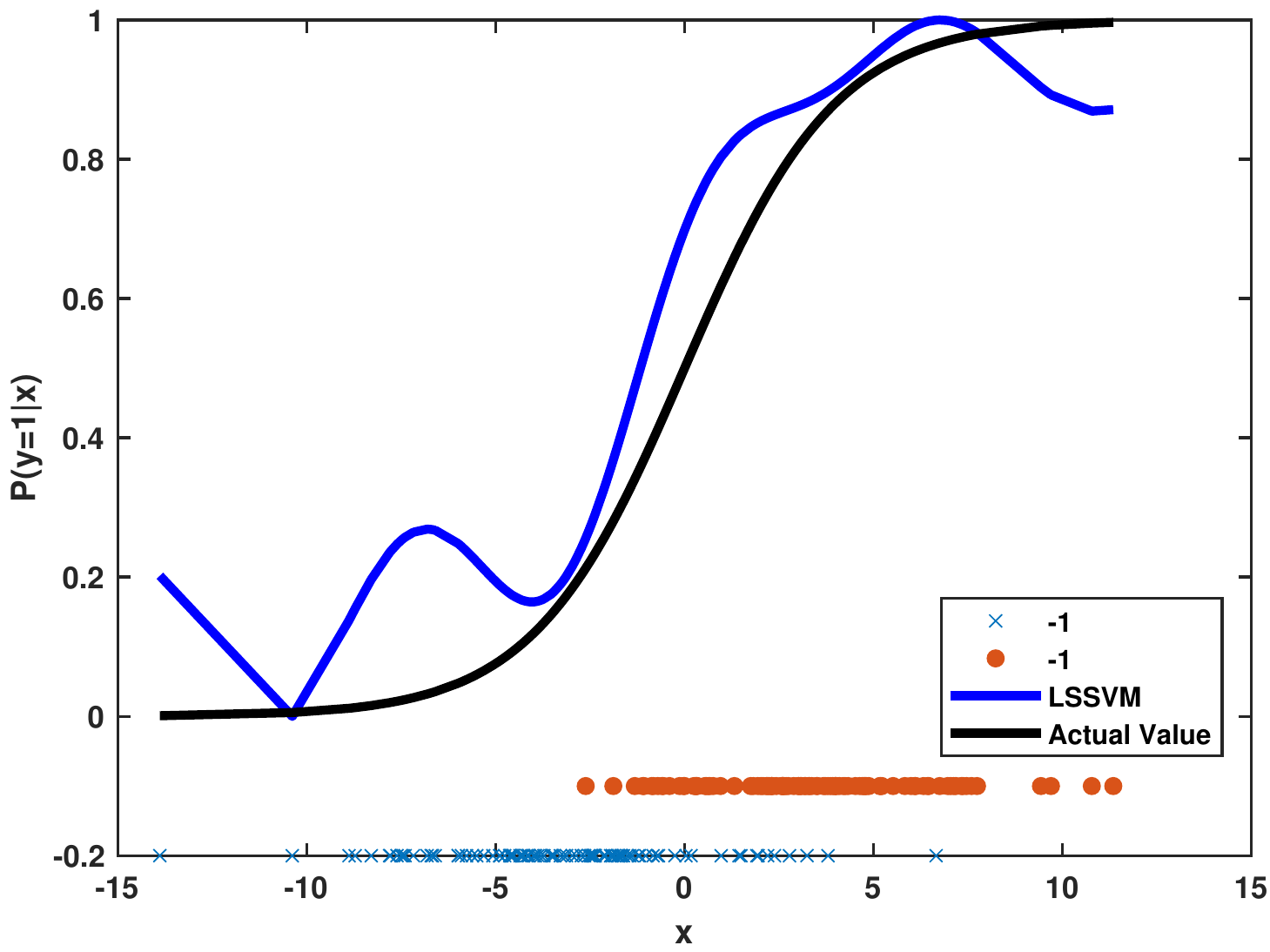}}
    \subfigure[VSVM]{\includegraphics[width=0.25\textheight]{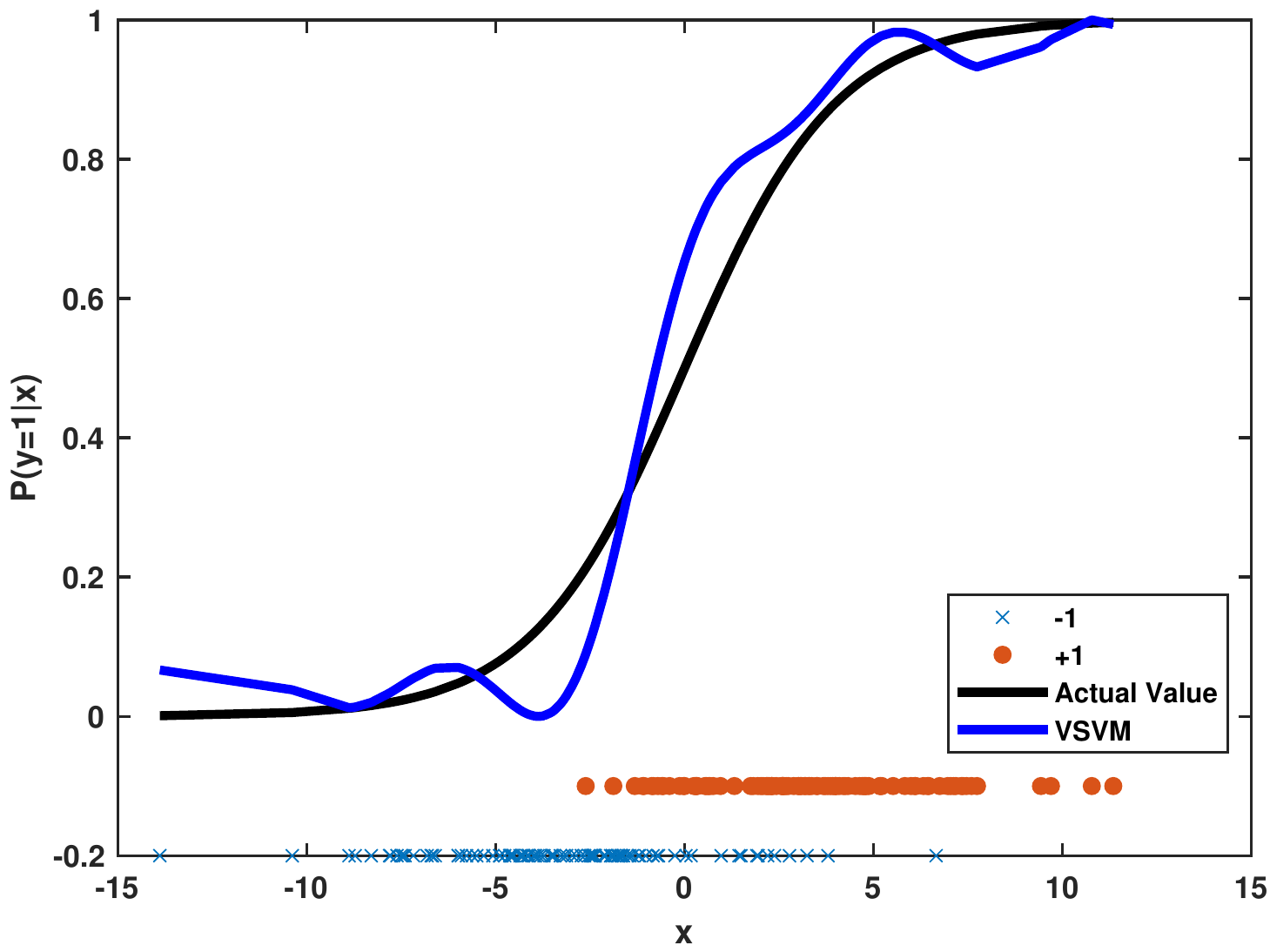}}
   \subfigure[IDLSSVM]{\includegraphics[width=0.25\textheight]{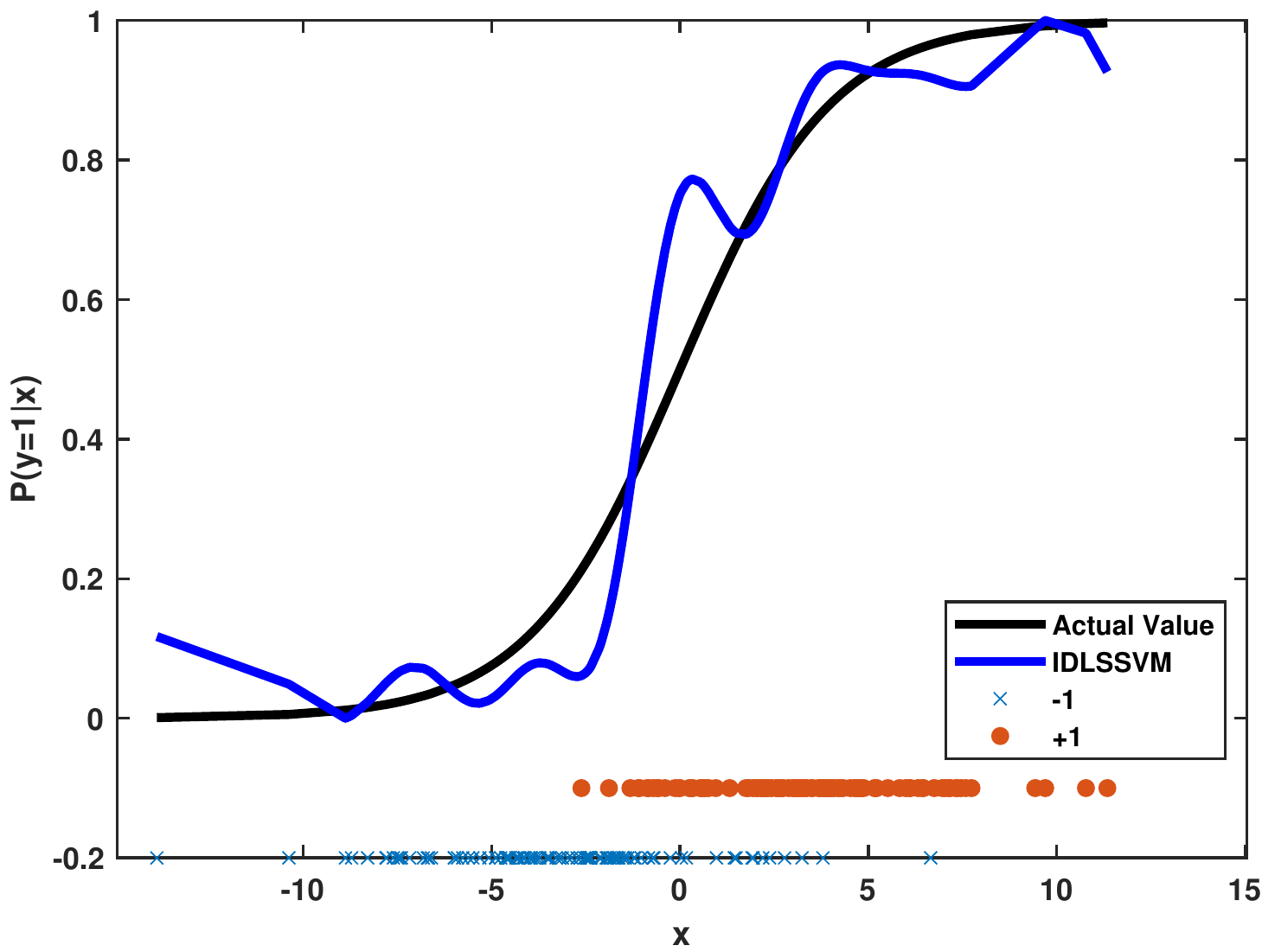}}
    \subfigure[$\varepsilon$-$L_1$SVM]{\includegraphics[width=0.25\textheight]{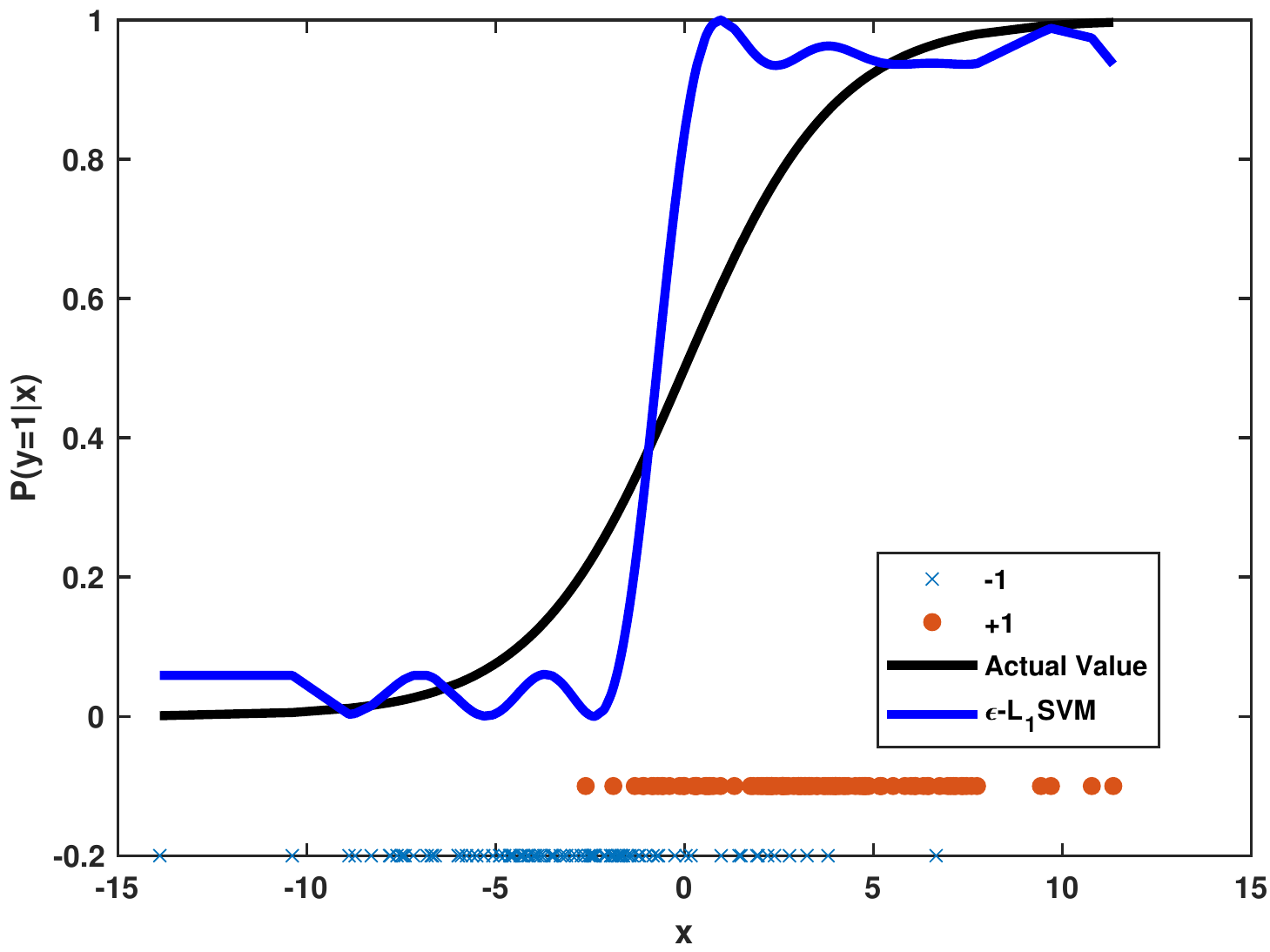}}
    \subfigure[$\varepsilon$-$L_1$VSVM]{\includegraphics[width=0.25\textheight]{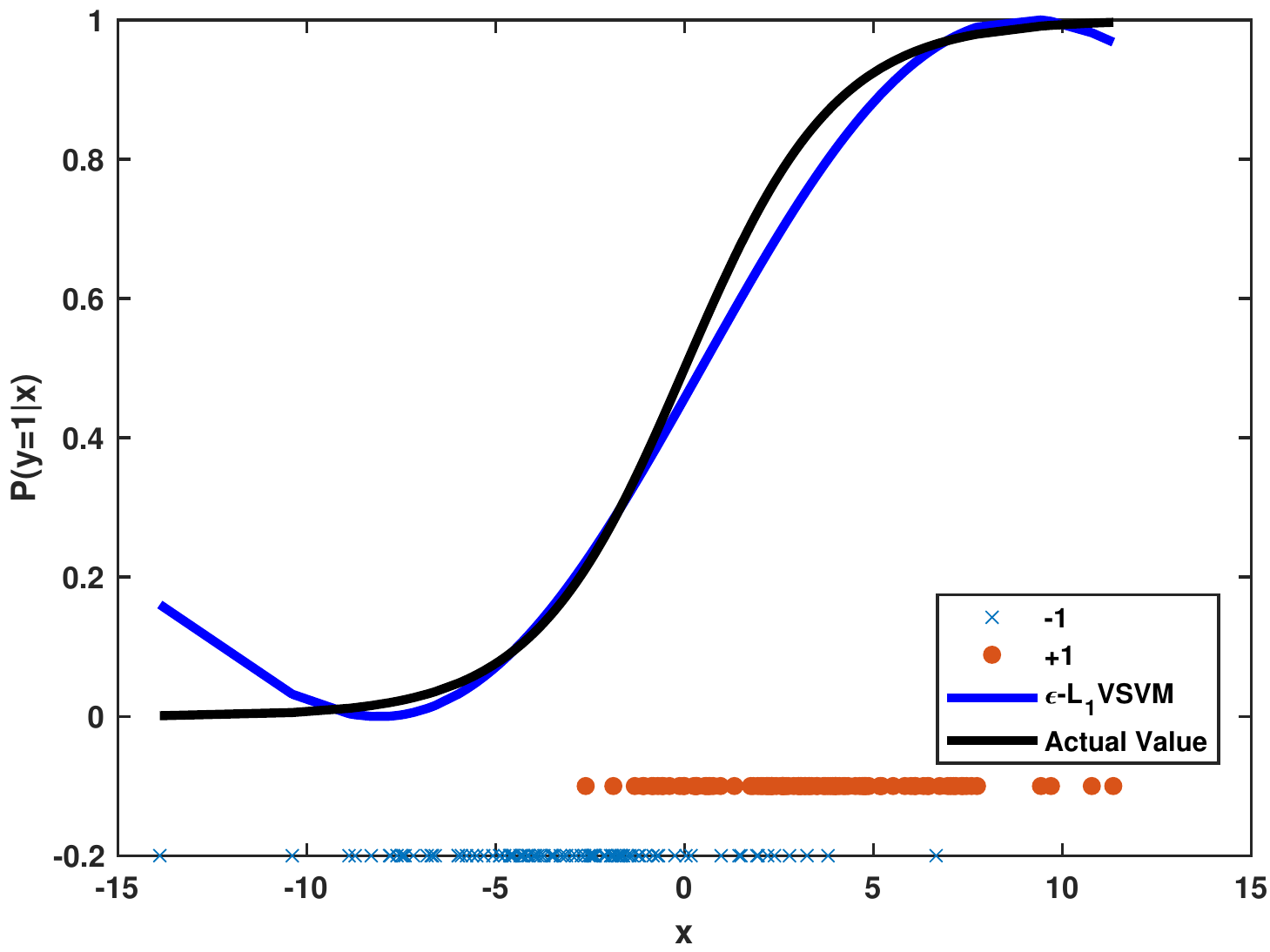}}
\caption{Robustness test of six classifiers with 200 one-dimensional training samples.}
\label{robust}
\end{figure*}

\textbf{Experiment 2 : Robustness test of the classifiers.}
Since VSVM and $\varepsilon$-$L_1$VSVM are based on the Fredholm integral equation and approximate the conditional probability by estimating the distribution function. In the following, we show the stability of the solutions of the different classifiers as mentioned in \citep{mazaheri2020robust,vapnik2020complete}.
Similar to the way of constructing the data above, this experiment also constructs two normally distributed data with a total number of N=200, but with one dimension. Therefore, the training data $\{(x_i,y_i),i=1,...,N\}$ are simulated as follows:
\begin{equation}\label{Gaussian}
\begin{split}
X_i \vert y=1\sim N\left( -3 ,3 \right) \ and \ X_i \vert y=-1\sim N\left( 3 ,3 \right).
\end{split}
\end{equation}

In order to reflect the stability of the output results of each classifier, we use the rbf kernel function for nonlinear experiments. The experimental results are shown in Figure \ref{robust}, where the black lines are $P(y=1\vert x)=\frac{1}{1+e^{-0.5x}}$. The horizontal coordinates are the values of the one-dimensional data and the data distribution is shown at the bottom of the figure, where red is the positive class and blue is the negative class; the vertical coordinates represent the probability values estimated by the classifier, and the blue image is a two-dimensional curve of the estimated results versus input variable for each nonlinear classifier. In addition, the v-values are used in version \eqref{v2}.

From Figure \ref{robust}, we see that the classifiers based on Fredholm integral equation, like VSVM and $\varepsilon$-$L_1$VSVM, they are more smooth and stable in their prediction. In contrast, the predictions by C-SVM, LSSVM, IDLSSVM, $\varepsilon$-$L_1$SVM are less smooth. That is to say, the estimation of the v-value is more robust the traditional ones.

\subsection{Experiments on benchmark datasets}\label{subsec42}
In this subsection, we conduct numerical experiments to show the effectiveness of the proposed $\varepsilon$-$L_{1}$VSVM for some small benchmark datasets.
We apply six classifiers to seventeen UCI datasets summarized in Table \ref{datasets}, which could retrieve the data from UCI depository of Machine Learning Dataset $\footnote{\url{https://archive.ics.uci.edu/ml/index.php}}$.

In all the experiments, we normalize all feature vectors to be in [0,1] and the standard 10-fold cross validation is performed to select the optimal parameters
according to selected evaluation indicator. And, considering that the real data distribution is unknown, the v-vector or v-matrix in the version of equation \eqref{v2} is used for all the following experiments.
\begin{table}[h]
\caption{Descriptions of 17 real datasets}
\label{datasets}
\centering
\begin{tabular}{c|ccc}
\toprule
~~~~  Datasets   ~~~~~  &~~~ID~~~  &    \#of samples & \#of features  ~~~~\\
\midrule
AustralianCAS~    &D1& 690          & 14             \\
Cleveland\_HeartS &D2& 270          & 13             \\
Cleveland0v2      &D3& 201          & 13             \\
Cleveland0v3      &D4& 200          & 13             \\
Cleveland0v4      &D5& 178          & 13             \\
Cleverland0vr     &D6& 303          & 13             \\
Echo              &D7& 131          & 10             \\
Ecoli             &D8& 336          & 7              \\
Ecoli0vr          &D9& 336          & 7              \\
German            &D10& 1000         & 20             \\
Led7digit         &D11& 443          & 7              \\
Monks3            &D12& 121          & 6              \\
Monkst            &D13& 432          & 6              \\
WOBC              &D14& 699          & 9              \\
Yeast4vr          &D15& 1484         & 8              \\
Yeast456vr        &D16& 1484         & 8              \\
Abalone3vr	      &D17& 4177         & 8              \\

\bottomrule
\end{tabular}
\end{table}

\textbf{Linear results:} Table \ref{linear} summarizes the linear classification results for all 17 datasets and the best G-means is shown in boldface. For each measure, the last two rows show the average G-means and the number of wins, draws and losses between our $\varepsilon$-$L_{1}$VSVM and other classifiers. Specially, we also give the corresponding Acc values to ensure the reasonableness of the G-means. It can be clearly seen that $\varepsilon$-$L_{1}$VSVM achieves the best performance in 14 out of 17.
Grouping the results in the table for comparison, it is worth noting that $\varepsilon$-$L_{1}$VSVM has higher G-means compared to $\varepsilon$-$L_{1}$SVM in all seventeen datasets. Similarly, VSVM performs better than LSSVM on most of the datasets.
Overall, $\varepsilon$-$L_{1}$VSVM exhibit competitive advantage towards their competitor. Further, VSVM and $\varepsilon$-$L_{1}$VSVM are the same type of model containing data location information. But $\varepsilon$-$L_{1}$VSVM using $\varepsilon$-insensitive loss, which is sparse and robust, and it perform better than VSVM.

\begin{table}[h]
\centering
\renewcommand{\arraystretch}{1.4}
\caption{ Testing results on benchmark datasets for linear classifiers.}
\label{linear}
\resizebox{\textwidth}{!}{
\begin{tabular}{c|cccccc}
\toprule
ID               & CSVM                & LSSVM             & VSVM
                       &IDLSSVM              & $\varepsilon$-$L_{1}$SVM             & $\varepsilon$-$L_{1}$VSVM                   \\
                       & G-means(Acc)(\%)           & G-means(Acc)(\%)  & G-means(Acc)(\%)           & G-means(Acc)(\%)  & G-means(Acc)(\%)            & G-means(Acc)(\%) \\
\midrule
D1         & 85.97$\pm$0.00(85.51)          & 86.33$\pm$0.16(85.90)
                       & \textbf{86.86$\pm$0.31(86.74)} &86.50$\pm$0.46(86.12)
                       & 86.16$\pm$0.43(86.25)
                       & 86.38$\pm$0.59(86.40)           \\
D2      & 82.84$\pm$0.42(83.33)          & 83.04$\pm$0.61(83.67)
                       & 83.39$\pm$0.85(84.00)          &82.81$\pm$0.70(83.11)
                       & 82.63$\pm$0.66(83.03)
                       & \textbf{83.76$\pm$0.53(84.15)}  \\
D3            & 83.47$\pm$1.70(91.39)          & 79.25$\pm$1.92(90.15)
                       & 82.24$\pm$1.81(91.34)          &81.95$\pm$1.15(91.84)
                       & 81.20$\pm$1.60(91.74)
                       & \textbf{86.61$\pm$1.87(92.44)}  \\
D4          & 88.53$\pm$1.22(93.45)          & 84.52$\pm$2.06(92.90)
                       & 86.14$\pm$2.08(93.40)          &85.43$\pm$1.56(92.55)
                       & 88.48$\pm$1.79(93.60)
                       & \textbf{89.44$\pm$1.64(93.95)}  \\
D5           & 73.68$\pm$5.51(95.39)          & 64.89$\pm$7.89(95.00)
                       & 64.04$\pm$5.19(95.11)          &60.50$\pm$2.74(94.78)
                       & 76.13$\pm$2.37(95.11)
                       & \textbf{77.79$\pm$2.32(95.73)}  \\
D6          & 83.65$\pm$0.56(84.12)          & 83.98$\pm$0.50(84.55)
                       & 84.11$\pm$0.53(84.65)          &83.17$\pm$0.92(83.33)
                       & 82.72$\pm$0.42(82.90)
                       & \textbf{84.37$\pm$0.61(85.02)}  \\
D7                   & 85.17$\pm$0.78(89.54)          & 82.51$\pm$2.34(88.77)
                       & 82.82$\pm$1.53(89.69)          &85.70$\pm$0.68(90.61)
                       & 86.81$\pm$0.89(90.45)
                       & \textbf{87.12$\pm$0.75(91.15)}  \\
D8                  & 75.39$\pm$2.28(92.32)          & 42.12$\pm$2.79(90.54)
                       & 57.24$\pm$2.20(91.36)          &43.79$\pm$2.04(90.63)
                       & 73.35$\pm$2.01(91.78)
                       & \textbf{76.64$\pm$1.88(92.86)}  \\
D9               & 96.51$\pm$0.46(96.61)          & 96.65$\pm$0.20(96.67)
                       & 96.65$\pm$0.20(96.08)          &96.37$\pm$0.29(96.49)
                       & 96.36$\pm$0.27(96.83)
                       & \textbf{97.05$\pm$0.18(97.02)}  \\
D10                & 63.07$\pm$0.60(75.74)          & 62.11$\pm$0.54(75.28)
                       & 62.10$\pm$0.52(75.61)          &63.54$\pm$0.72(75.52)
                       & 69.60$\pm$0.30(76.40)
                       & \textbf{69.78$\pm$0.59(76.53)}  \\
D11             & \textbf{89.81$\pm$1.03(96.30)} & 00.00$\pm$0.00(91.65)
                       & 00.00$\pm$0.00(91.65)          &43.31$\pm$5.03(92.33)
                       & 86.96$\pm$1.65(96.45)
                       & 88.32$\pm$1.02(96.61)           \\
D12                & 76.87$\pm$1.70(77.02)          & 75.94$\pm$1.83(76.12)
                       & 76.45$\pm$1.98(76.45)          &76.26$\pm$2.00(76.36)
                       & 77.39$\pm$1.76(77.44)
                       & \textbf{78.55$\pm$1.39(78.60)}  \\
D13                & 82.73$\pm$0.57(82.78)          & 78.09$\pm$0.54(79.88)
                       & 81.40$\pm$0.91(82.22)          &81.69$\pm$0.66(81.69)
                       & 87.04$\pm$0.93(87.57)
                       & \textbf{87.60$\pm$1.05(88.08)}  \\
D14                  & 96.38$\pm$0.33(96.67)          & 94.71$\pm$0.19(95.77)
                       & 94.90$\pm$0.23(95.91)          &94.41 0.16(95.57)
                       & 97.10$\pm$0.17(96.97)
                       & \textbf{97.32$\pm$0.23(97.14)}  \\
D15               & \textbf{82.86$\pm$0.48(94.59)} & 58.58$\pm$0.33(92.38)
                       & 75.53$\pm$0.62(93.97)          &64.00$\pm$0.85(92.97)
                       & 75.61$\pm$3.35(94.06)
                       & 81.13$\pm$0.62(94.60)           \\
D16          & 84.10$\pm$0.20(92.86)          & 77.61$\pm$0.58(91.37)
                       & 81.24$\pm$0.24(92.06)          &80.61$\pm$0.44(92.08)
                       & 84.36$\pm$0.71(92.82)
                        &\textbf{84.85$\pm$0.33(92.97)}  \\
D17                    &70.95$\pm$0.17(78.01)                & 69.06$\pm$0.17(77.31)
                       & 69.05$\pm$0,17(77.31) & 70.56$\pm$0.10(77.55) & 73.26$\pm$0.18(77.67) & \textbf{74.01$\pm$0.15(78.39) }   \\
\midrule
\textbf{Avg.}   &82.47             & 71.73             & 74.36             & 75.33             & 82.66             & \textbf{84.16}             \\
\textbf{W/T/L} & 2/0/15                     & 0/0/17            & 1/0/16                     & 0/0/17            &0/0/17     &--                \\
\bottomrule
\end{tabular}
}
\end{table}

\textbf{Nonlinear results:} The nonlinear classification results on the benchmark datasets are recorded in Table \ref{nonlinear}. Clearly, the overall performances of nonlinear $\varepsilon$-$L_{1}$VSVM are superior
to all that of other nonlinear classifiers. Moreover, $\varepsilon$-$L_{1}$VSVM gains higher G-means than $\varepsilon$-$L_{1}$SVM on 16 datasets and VSVM performs better than LSSVM on 12 datasets, therefore confirms the conclusion above further. From another point of view, our model is similar to a weighted model, which has a v-value weight and represents the location information of the data, while the contrasted IDLSSVM is a density-weighted model that also tries to reflect the distribution of the data. However, IDLSSVM uses KNN to calculate the samples in the domain and thus converts the corresponding density, which focuses on the situation within the local area, while $\varepsilon$-$L_{1}$VSVM considers a relative position of each sample in the whole data and highlights a global concept more.

\begin{table}[h]
\centering
\renewcommand{\arraystretch}{1.5}
\caption{ Testing results on benchmark datasets for nonlinear(rbf) classifiers.}
\resizebox{\textwidth}{!}{
\label{nonlinear}
\begin{tabular}{c|cccccc}
\toprule
ID               & CSVM                & LSSVM             & VSVM
                       &IDLSSVM                      & $\varepsilon$-$L_{1}$SVM             & $\varepsilon$-$L_{1}$VSVM                   \\
                       & G-means(Acc)(\%)            & G-means(Acc)(\%)
                       & G-means(Acc)(\%)            & G-means(Acc)(\%)
                       & G-means(Acc)(\%)            & G-means(Acc)(\%) \\
\midrule
D1          & 85.67$\pm$0.28(85.67)          & 85.61$\pm$0.60(85.70)
                       & 85.68$\pm$0.38(85.52)          & 86.54$\pm$0.41(86.39)
                       & 84.17$\pm$2.43(84.74)          & \textbf{86.30$\pm$0.82(86.30)}  \\
D2      & 82.45$\pm$0.74(82.93)          & 82.27$\pm$0.58(83.11)
                       & 83.40$\pm$0.66(83.89)          & 82.81$\pm$0.70(83.11)
                       & 82.63$\pm$0.60(82.96)          & \textbf{83.66$\pm$0.54(84.00)}  \\
D3           & 82.55$\pm$3.32(91.49)          & 72.75$\pm$1.75(88.56)
                       & 83.36$\pm$0.90(92.19)          &81.58$\pm$1.15(91.84)
                       & 81.93$\pm$1.12(91.79 )         & \textbf{84.32$\pm$1.73(92.24)}  \\
D4           & 87.60$\pm$1.83(93.40)          & 79.82$\pm$1.45(91.50)
                       & 85.73$\pm$1.62(93.35)          &85.43$\pm$1.56(92.55)
                       & 85.31$\pm$2.19(92.35)          & \textbf{90.86$\pm$1.51(94.70)}  \\
D5          & 75.12$\pm$3.61(95.05)          & 60.29$\pm$3.99(94.33)
                       & 78.39$\pm$1.52(96.01)          &64.83$\pm$5.06(95.67)
                       & 65.18$\pm$5.01(95.06)          & \textbf{83.26$\pm$2.81(96.07)}  \\
D6          & 83.71$\pm$0.54(84.22)          & 82.33$\pm$0.69(82.94)
                       & 84.22$\pm$0.56(84.68)          &82.85$\pm$0.70(82.97)
                       & 82.55$\pm$0.78(83.17)          & \textbf{84.28$\pm$0.45(84.55)}  \\
D7                   & 86.46$\pm$1.36(89.47)          & 83.63$\pm$1.42(88.86)
                       & 85.88$\pm$0.77(90.00)          &84.46$\pm$.31(90.15)
                       & 84.27$\pm$1.23(89.85)          & \textbf{86.72$\pm$0.78(91.15)}  \\
D8                 & 78.47$\pm$0.73(93.57)          & 75.91$\pm$1.70(93.09)
                       & 73.53$\pm$1.34(93.18)          &77.29$\pm$1.51(93.54)
                       & 73.59$\pm$1.34(92.92)          & \textbf{78.98$\pm$1.69(93.66)}  \\
D9               & 96.73$\pm$0.30(96.75)          & 96.83$\pm$0.18(96.84)
                       & 97.26$\pm$0.24(97.20)          &96.47$\pm$0.51(96.55)
                       & 97.10 0.25(97.14)          & \textbf{97.27$\pm$0.26(97.26)}  \\
D10                 & 62.23$\pm$0.58(76.46)          & 66.38$\pm$0.81(76.98)
                       & 64.31$\pm$ 0.82(76.39)         &64.02 0.98(76.42)
                       & 64.85$\pm$1.00(76.43)          & \textbf{69.83$\pm$0.44(76.67)}  \\
D11              & 88.52$\pm$0.79(96.73)          & 87.90$\pm$1.67(96.64)
                       & 71.04$\pm$ 0.00(94.36)         &\textbf{90.65$\pm$0.09(96.88)}
                       & 88.35$\pm$1.27(96.68)          & 89.71$\pm$1.00(96.91)  \\
D12               & 87.31$\pm$1.82(87.36)          & 88.70$\pm$1.32(88.76)
                       & 84.62$\pm$0.59(85.18)          &88.69$\pm$1.04(88.68)
                       & 90.24$\pm$1.72(90.33)          & \textbf{92.17$\pm$0.98(92.15)}  \\
D13                & 99.10$\pm$0.82(99.12)          & 96.57$\pm$0.51(96.50)
                       & 97.40$\pm$0.46(97.53)          &97.12 0.26(97.01)
                       & 97.31$\pm$0.07(97.20)          & \textbf{99.71$\pm$0.29(99.72)}  \\
D14                   & 96.94$\pm$0.16(96.82)          & 96.63$\pm$0.16(96.70)
                       & 96.88$\pm$0.36(96.87)          &96.33$\pm$0.27(96.55)
                       & 96.79$\pm$0.19(96.82)          & \textbf{97.30$\pm$0.18(97.05)}  \\
D15               & 85.09$\pm$0.27(95.51)          & 82.31$\pm$0.46(95.10)
                       & 82.20$\pm$0.34(95.24)          &84.09 0.52(92.92)
                       & \textbf{86.82$\pm$0.47(95.80)} & 86.61$\pm$0.26(95.75)           \\
D16             & \textbf{86.77$\pm$0.51(93.07)} & 84.19$\pm$0.54(92.67)
                       & 84.62$\pm$0.29(92.83)          &82.32$\pm$0.57(95.05)
                       & 85.10$\pm$0.26(93.09)          & 86.62$\pm$0.46(93.23)        \\
D17 & 74.59$\pm$0.14(79.37) & 74.90$\pm$0.18(79.35) & 73.27$\pm$0.14(78.71) & 73.62$\pm$0.10(79.13) & 75.63$\pm$0.17(78.96) & \textbf{76.16$\pm$0.1479.43)}  \\
\midrule
\textbf{Avg.}   & 84.67             & 82.18             & 83.05             & 83.48             & 83.64             & \textbf{86.69}
              \\
\textbf{W/D/L} & 1/0/16                     & 0/0/17            & 0/0/17            & 1/0/16                     & 1/0/16    &--                      \\
\bottomrule
\end{tabular}
}
\end{table}

\begin{figure*}[h]
\setlength{\abovecaptionskip}{0.cm}
\setlength{\belowcaptionskip}{-0.cm}
\centering
    \subfigure[Cleveland-HeartS]{\includegraphics[width=0.225\textheight]{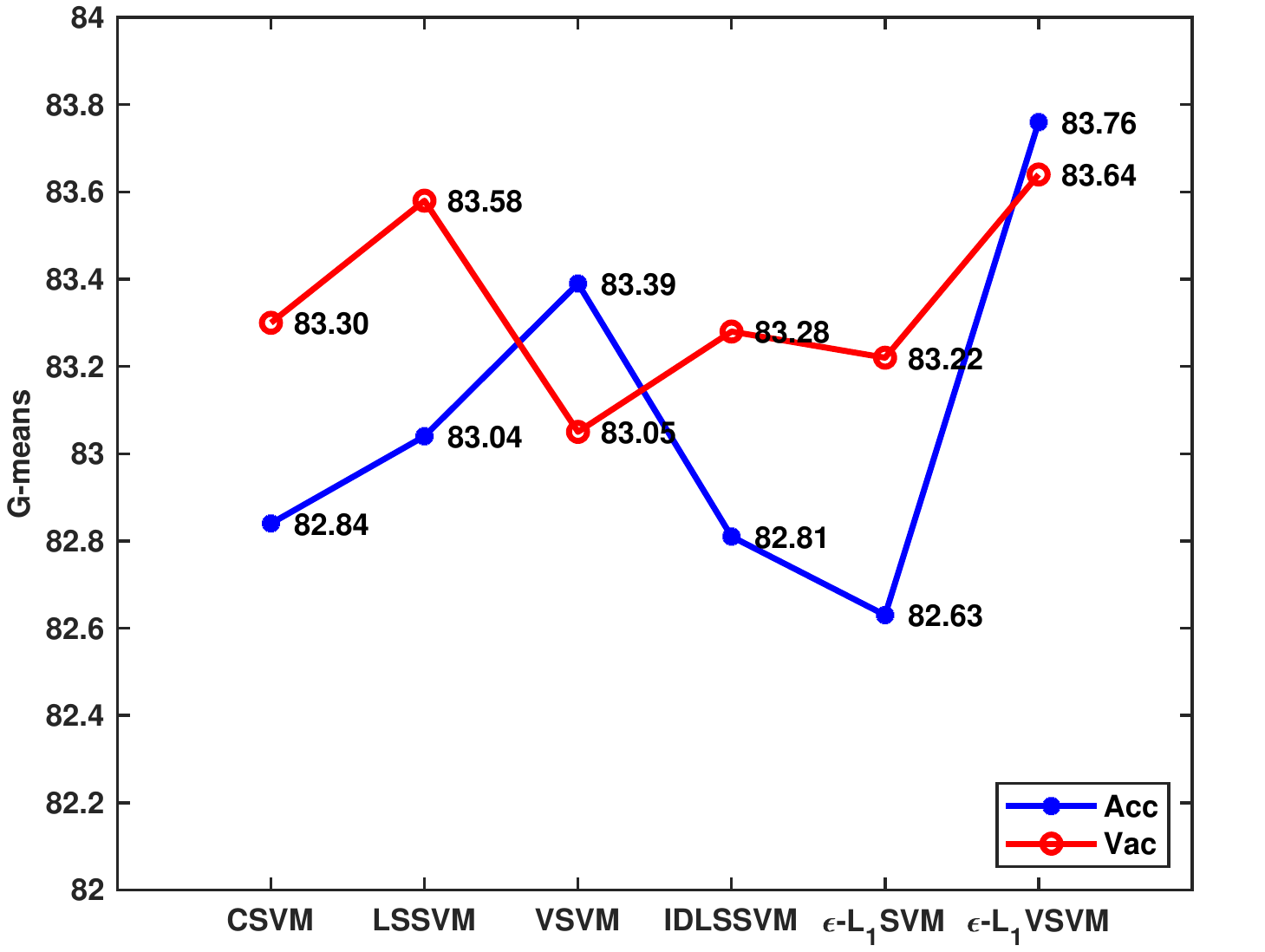}}
    \subfigure[Cleveland0v2]{\includegraphics[width=0.225\textheight]{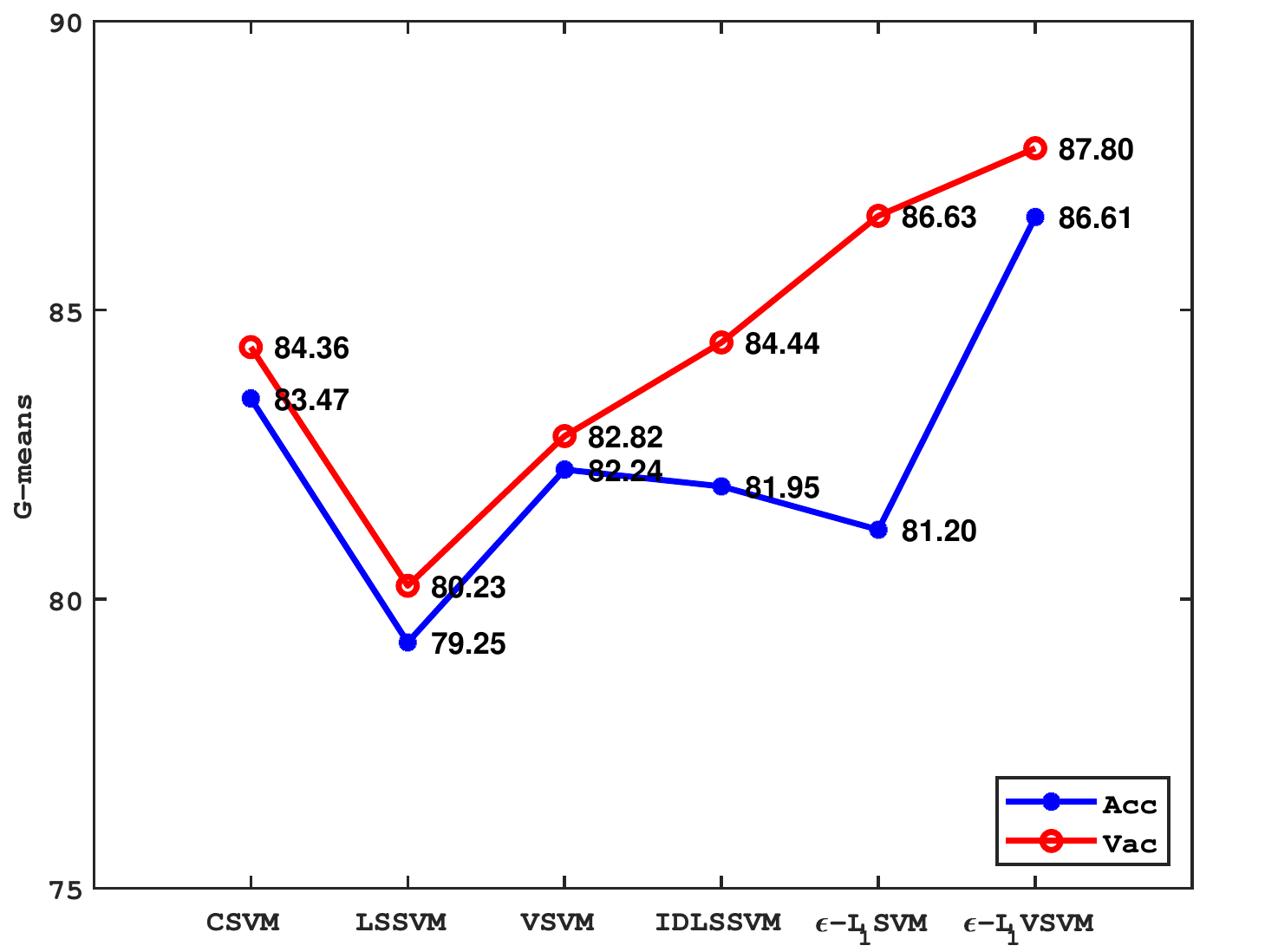}}
    \subfigure[Cleveland0v3]{\includegraphics[width=0.225\textheight]{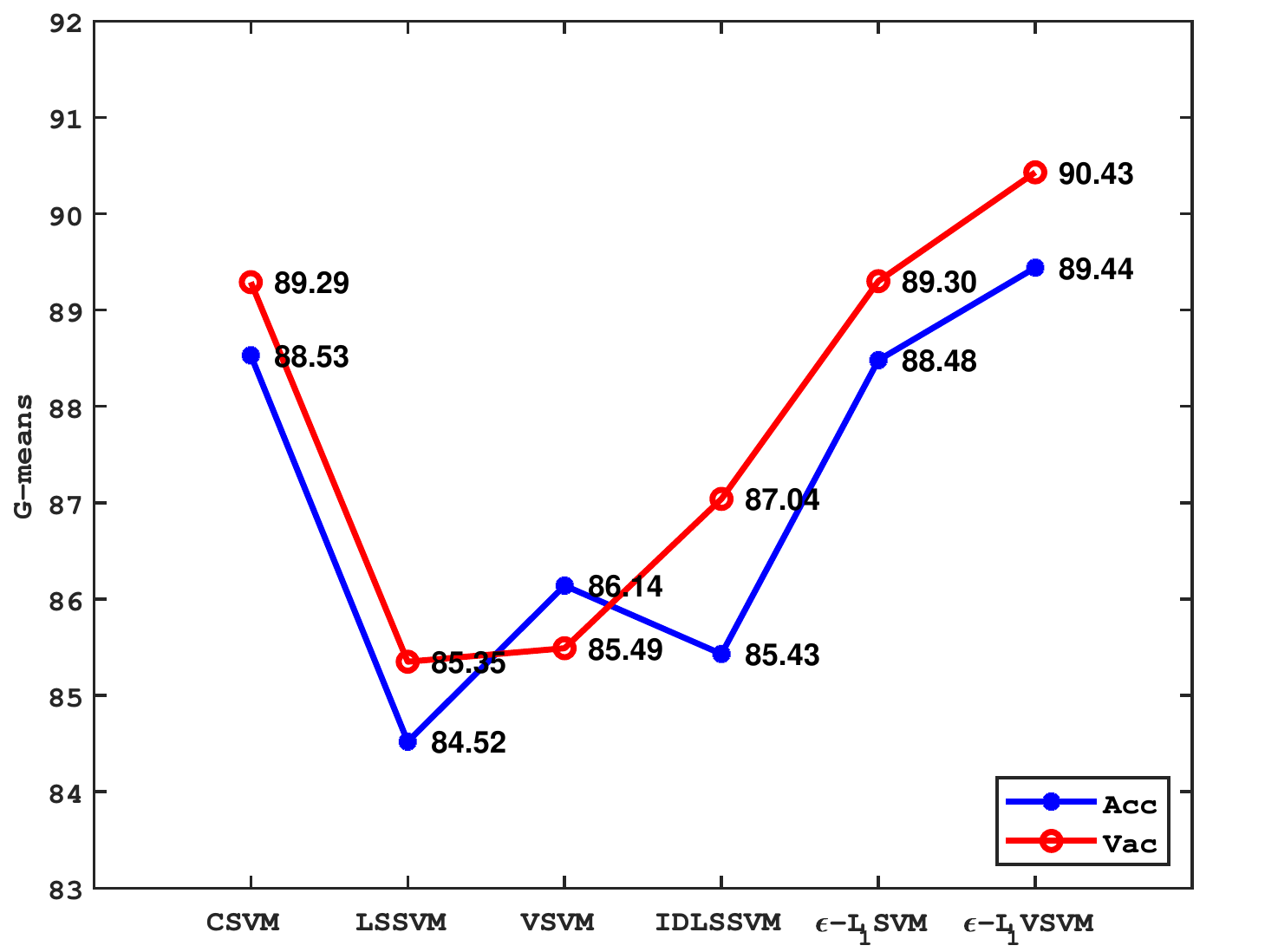}}
    \subfigure[Cleveland0v4]{\includegraphics[width=0.225\textheight]{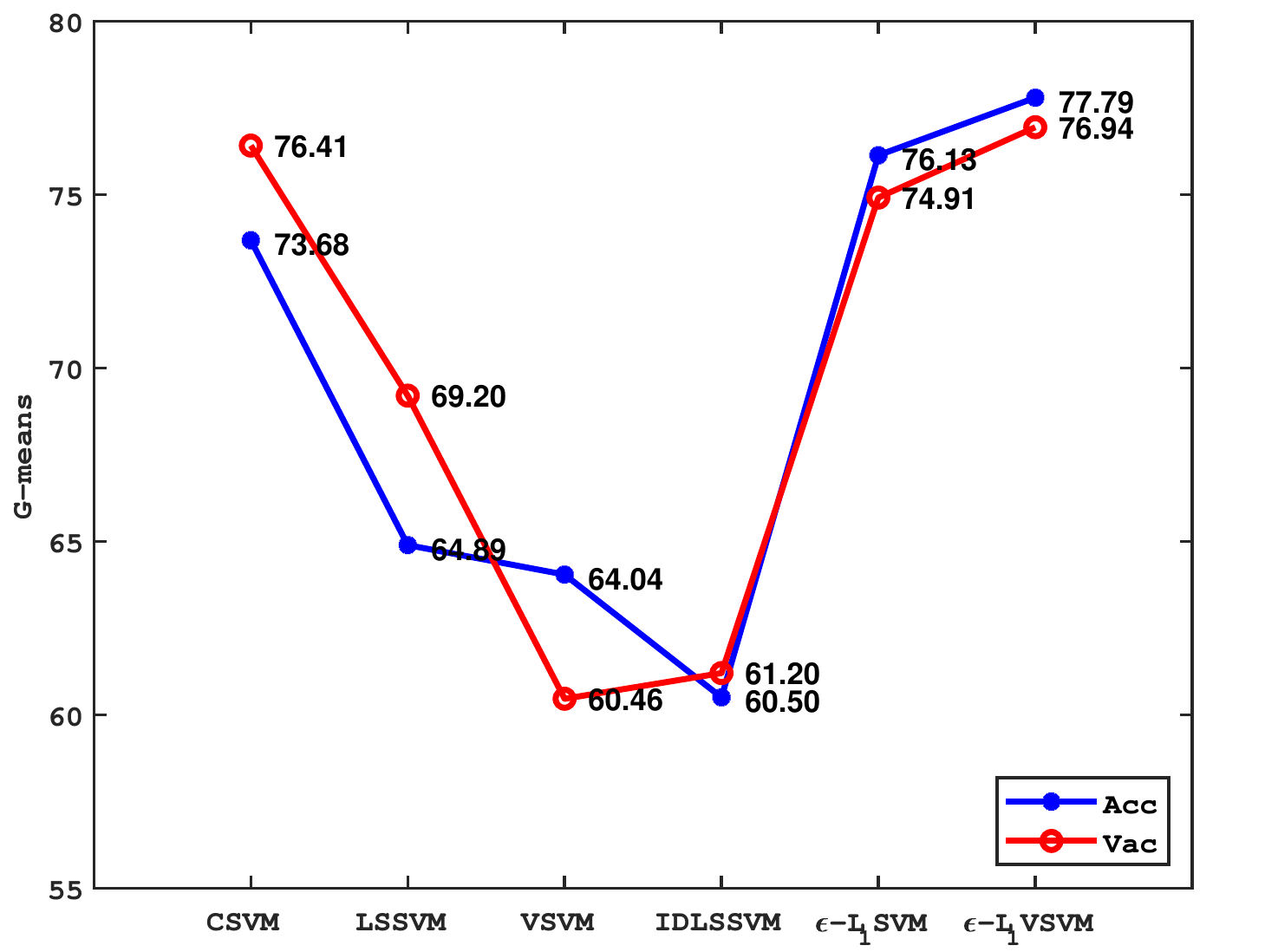}}
    \subfigure[Cleverland0vr]{\includegraphics[width=0.225\textheight]{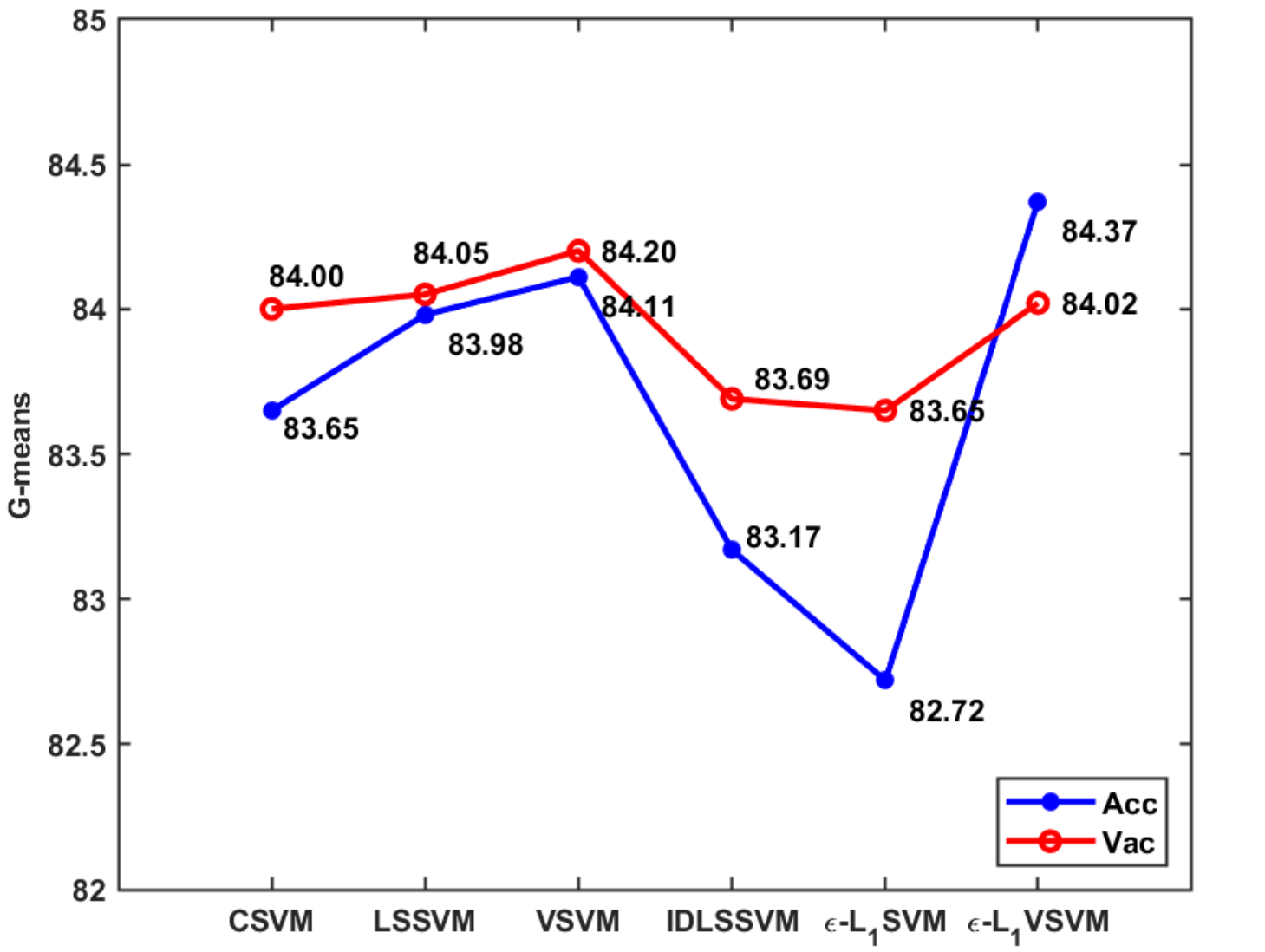}}
    \subfigure[Echo]{\includegraphics[width=0.225\textheight]{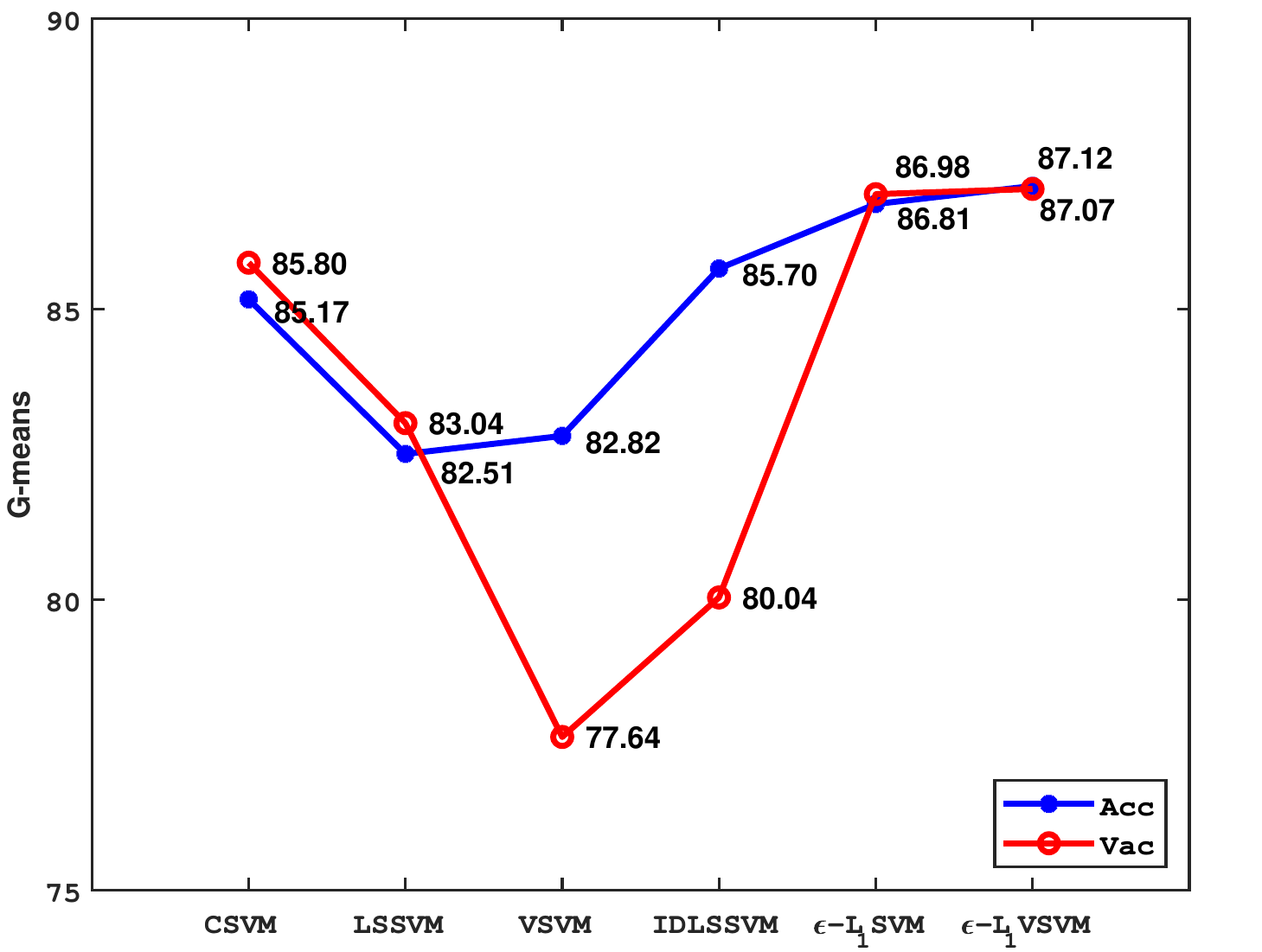}}
    \subfigure[Monks3]{\includegraphics[width=0.225\textheight]{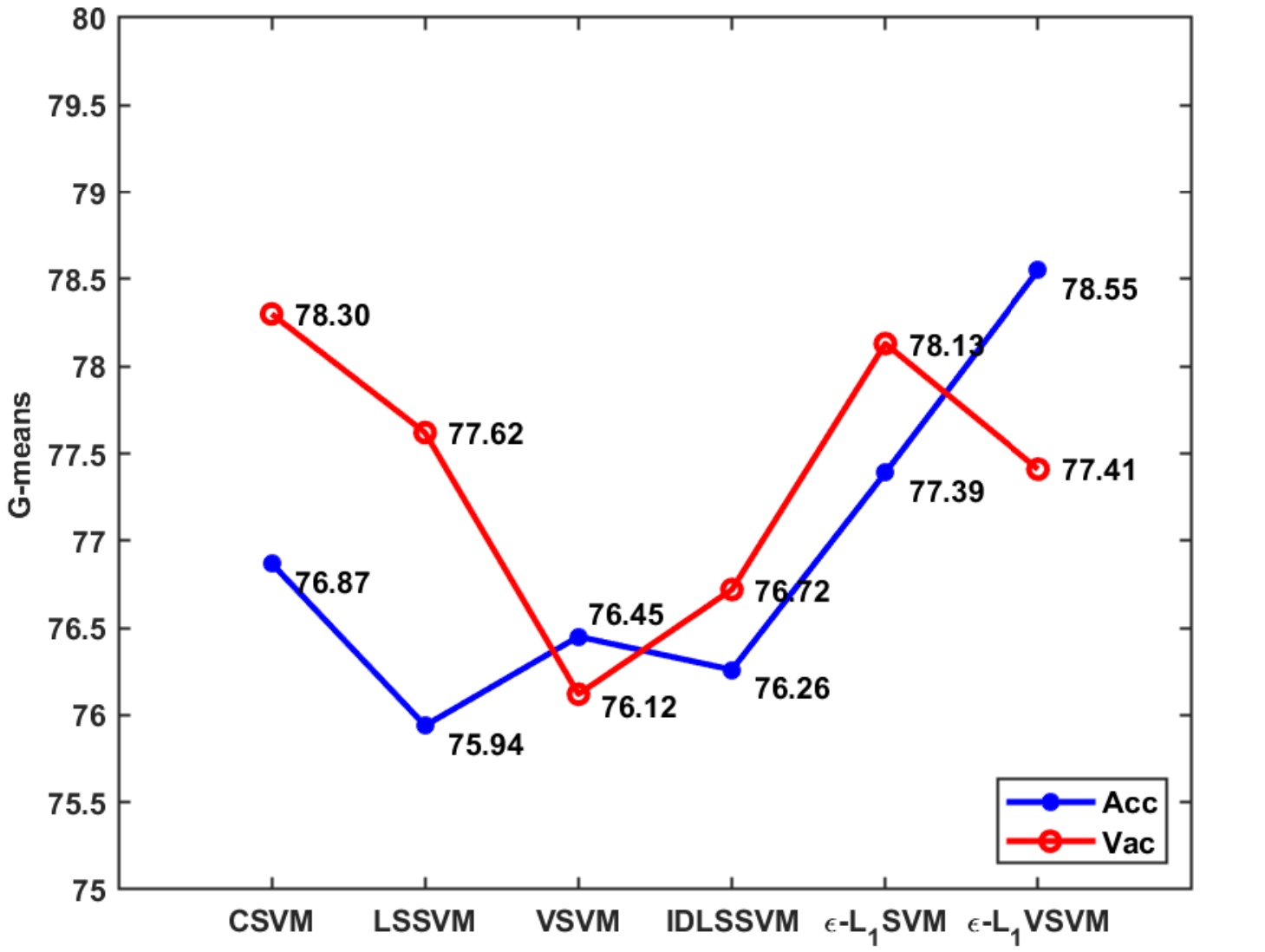}}
    \subfigure[Ecoli]{\includegraphics[width=0.225\textheight]{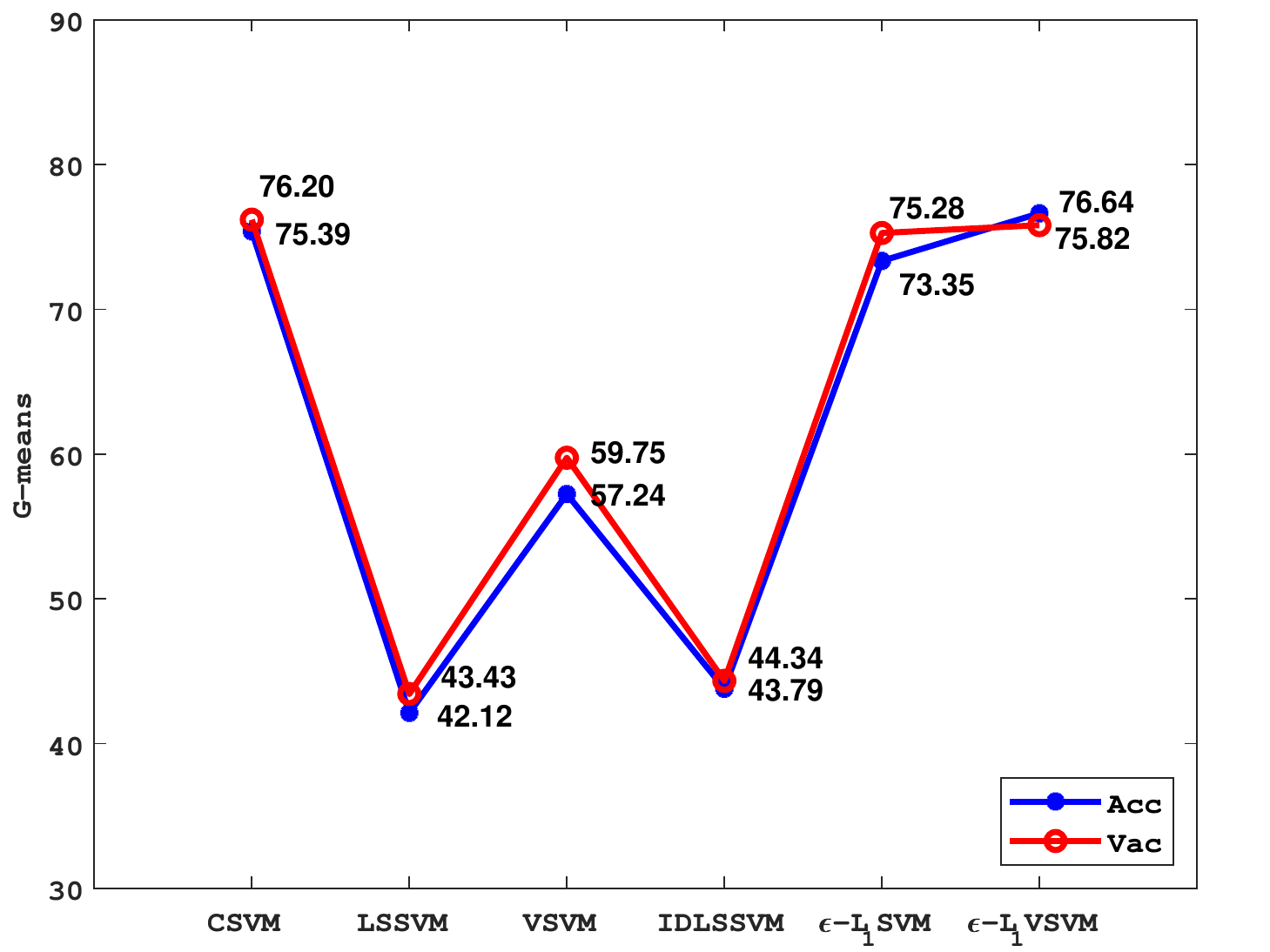}}
    \subfigure[Ecoli0vr]{\includegraphics[width=0.225\textheight]{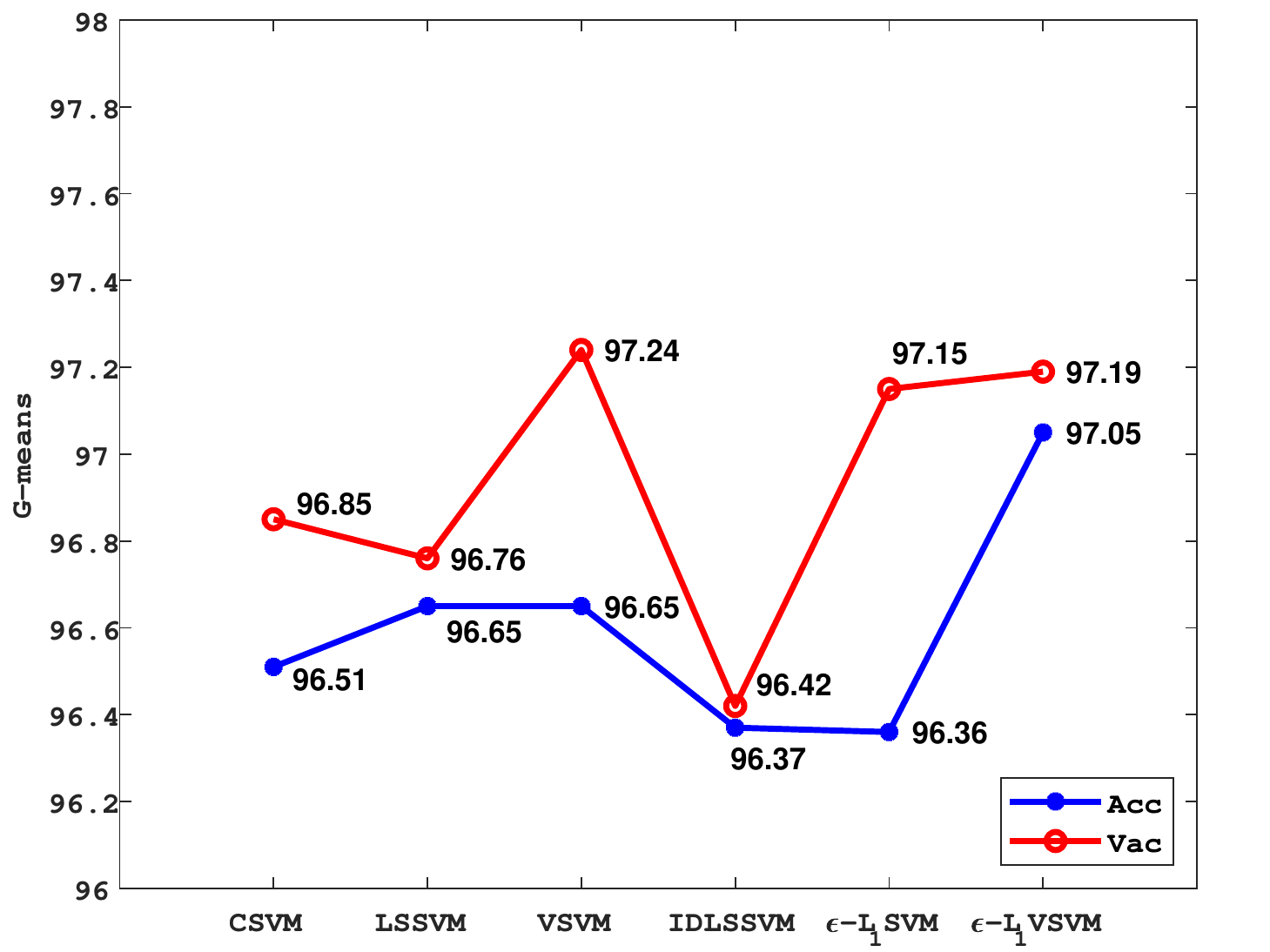}}
\caption{Linear results based on Acc \eqref{Acc} and Vac \eqref{Vac} as evaluation indicators for six classifiers on nine small datasets.}
\label{vac-acc}
\end{figure*}

\begin{figure*}[h]
\setlength{\abovecaptionskip}{0.cm}
\setlength{\belowcaptionskip}{-0.cm}
\centering
    \subfigure[Cleveland-HeartS]{\includegraphics[width=0.225\textheight]{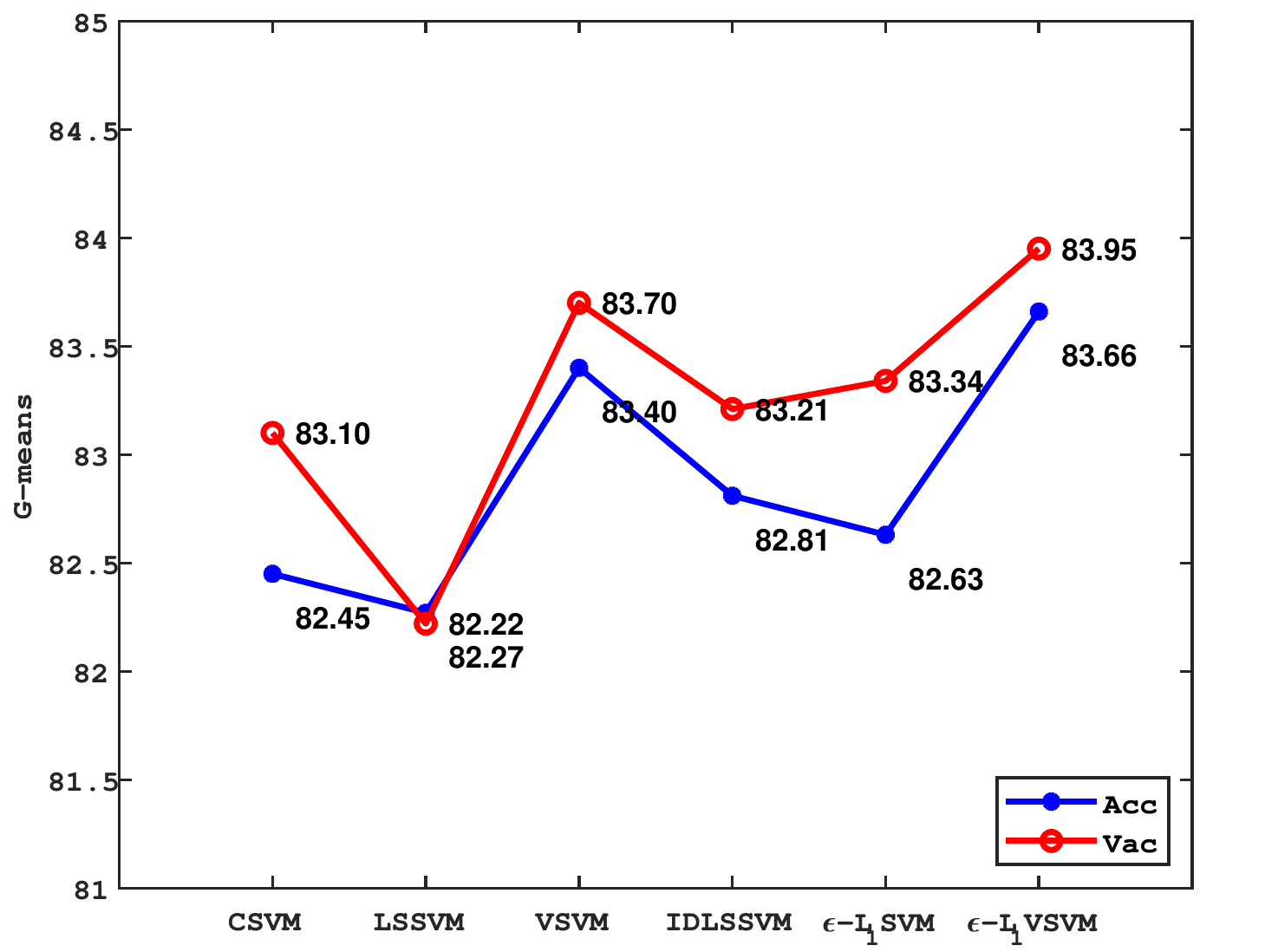}}
    \subfigure[Cleveland0v2]{\includegraphics[width=0.225\textheight]{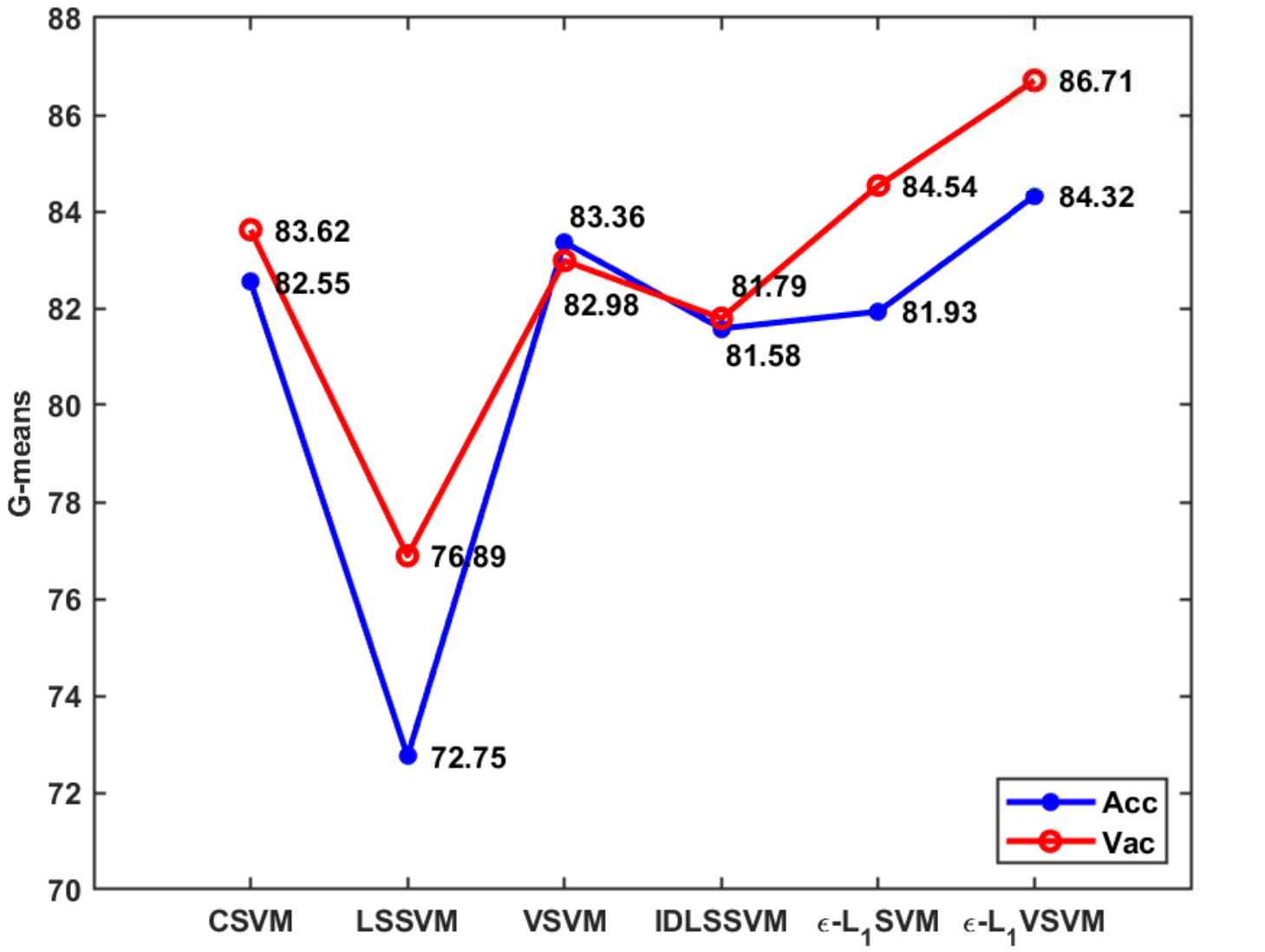}}
    \subfigure[Cleveland0v3]{\includegraphics[width=0.225\textheight]{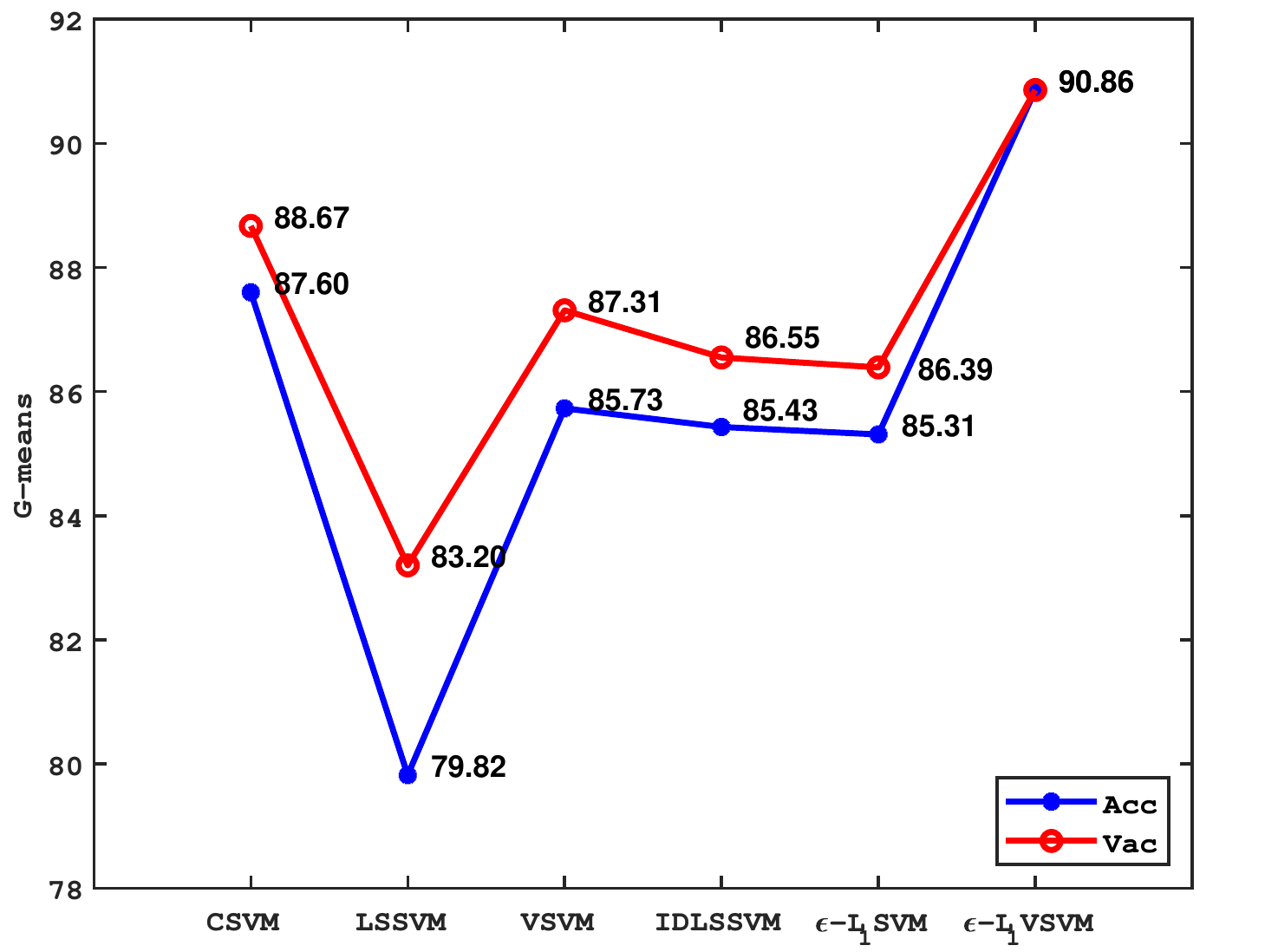}}
    \subfigure[Cleveland0v4]{\includegraphics[width=0.225\textheight]{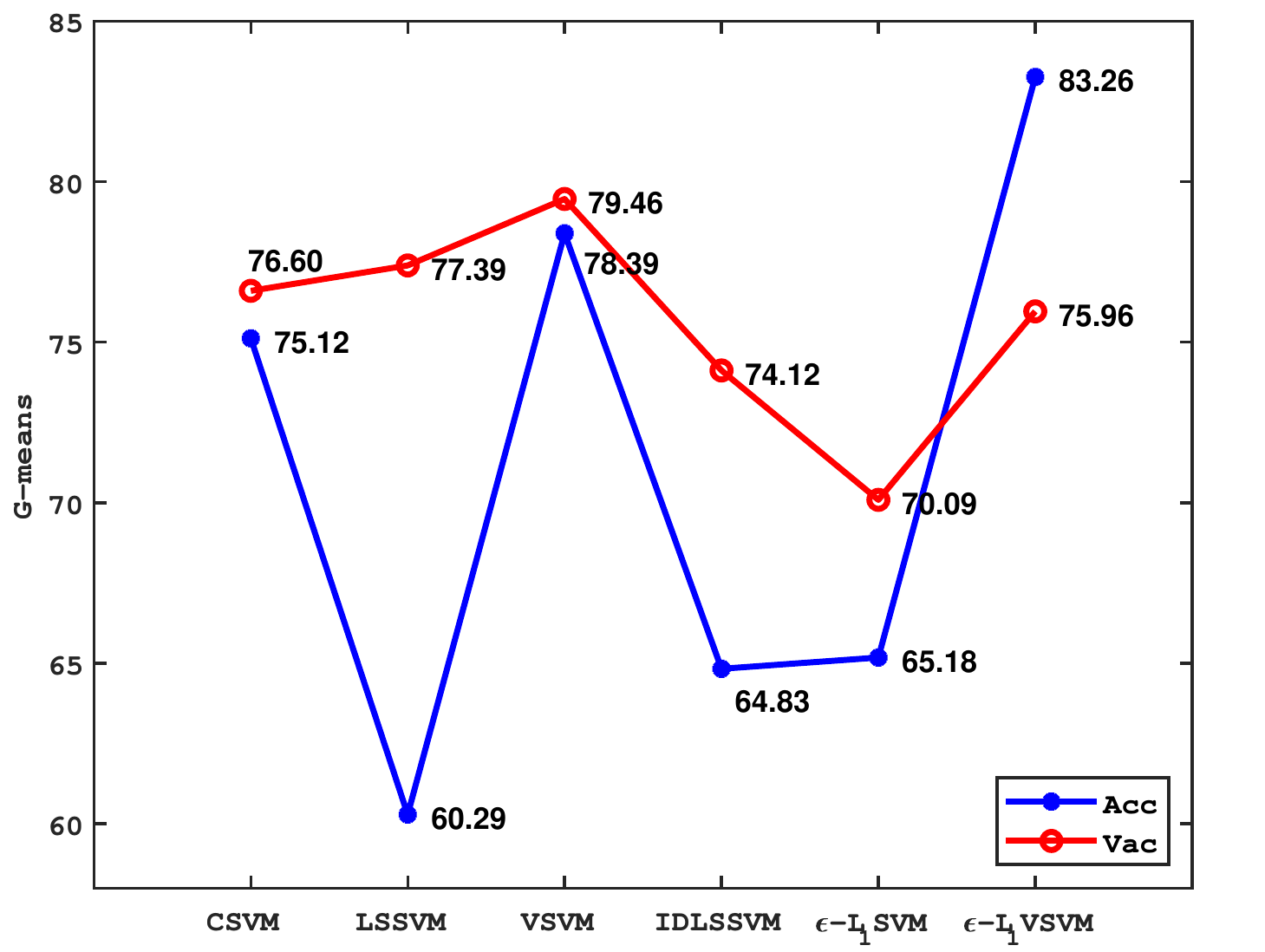}}
    \subfigure[Cleverland0vr]{\includegraphics[width=0.225\textheight]{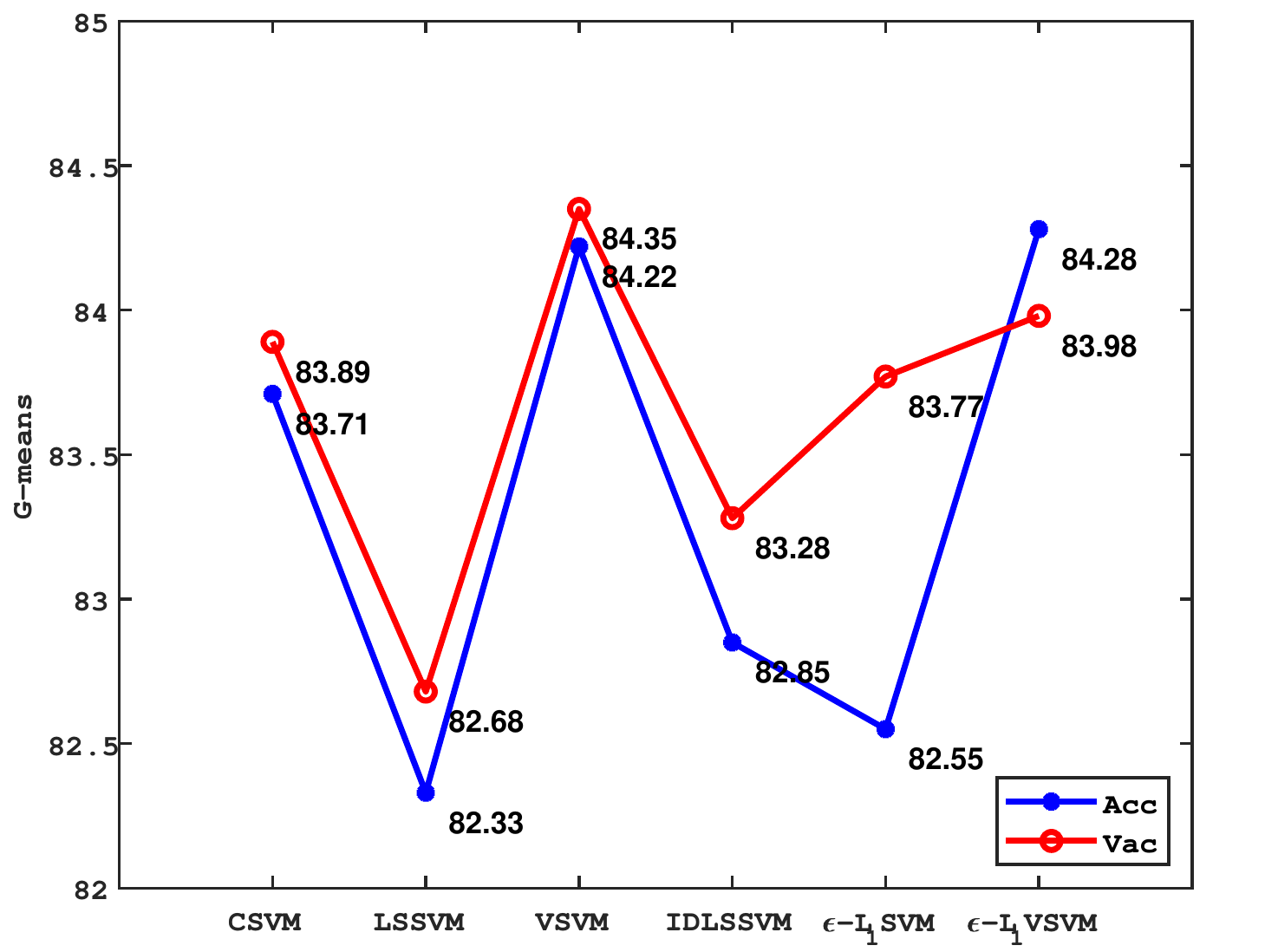}}
    \subfigure[Echo]{\includegraphics[width=0.225\textheight]{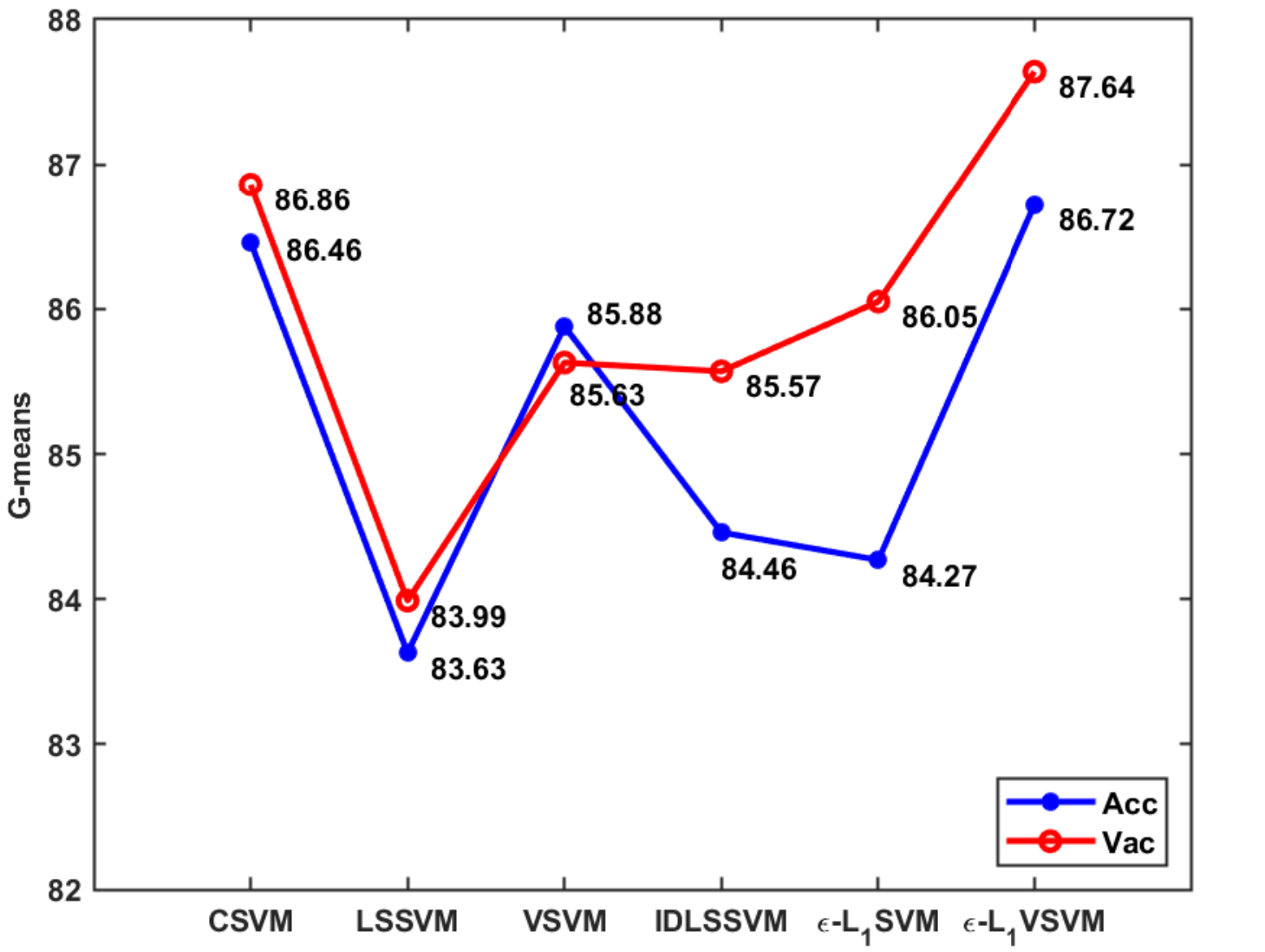}}
    \subfigure[Monks3]{\includegraphics[width=0.225\textheight]{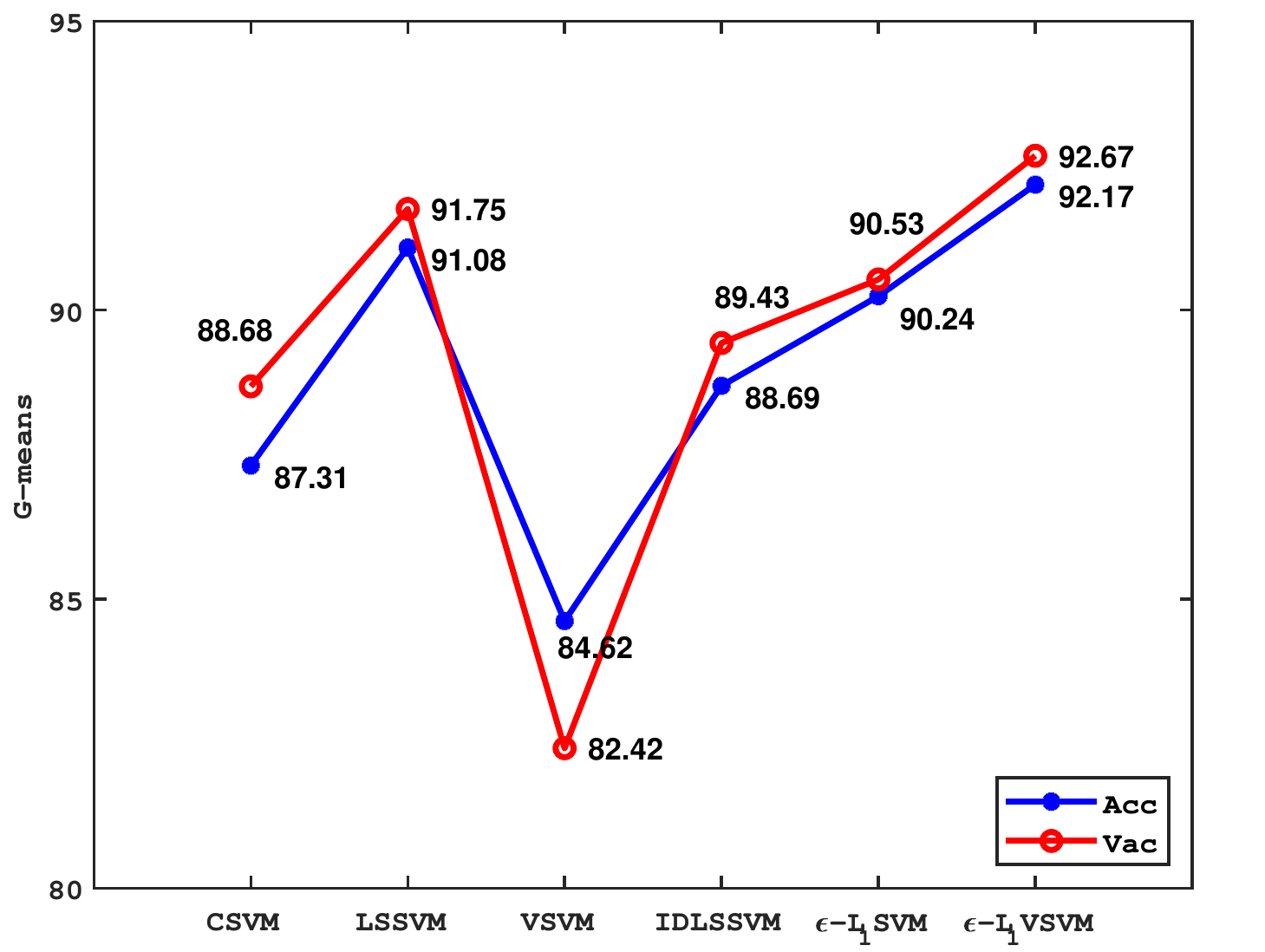}}
    \subfigure[Ecoli]{\includegraphics[width=0.225\textheight]{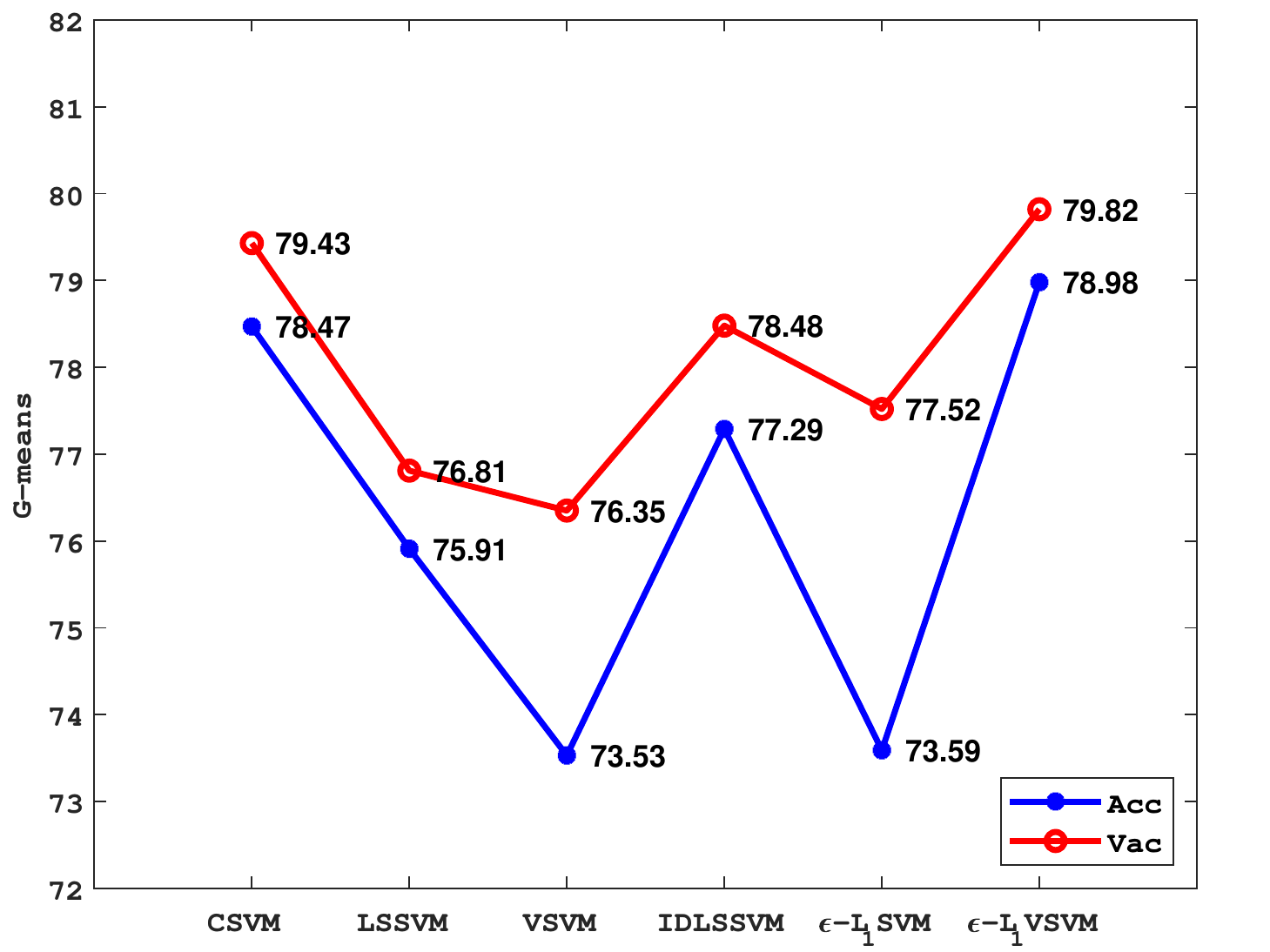}}
    \subfigure[Ecoli0vr]{\includegraphics[width=0.225\textheight]{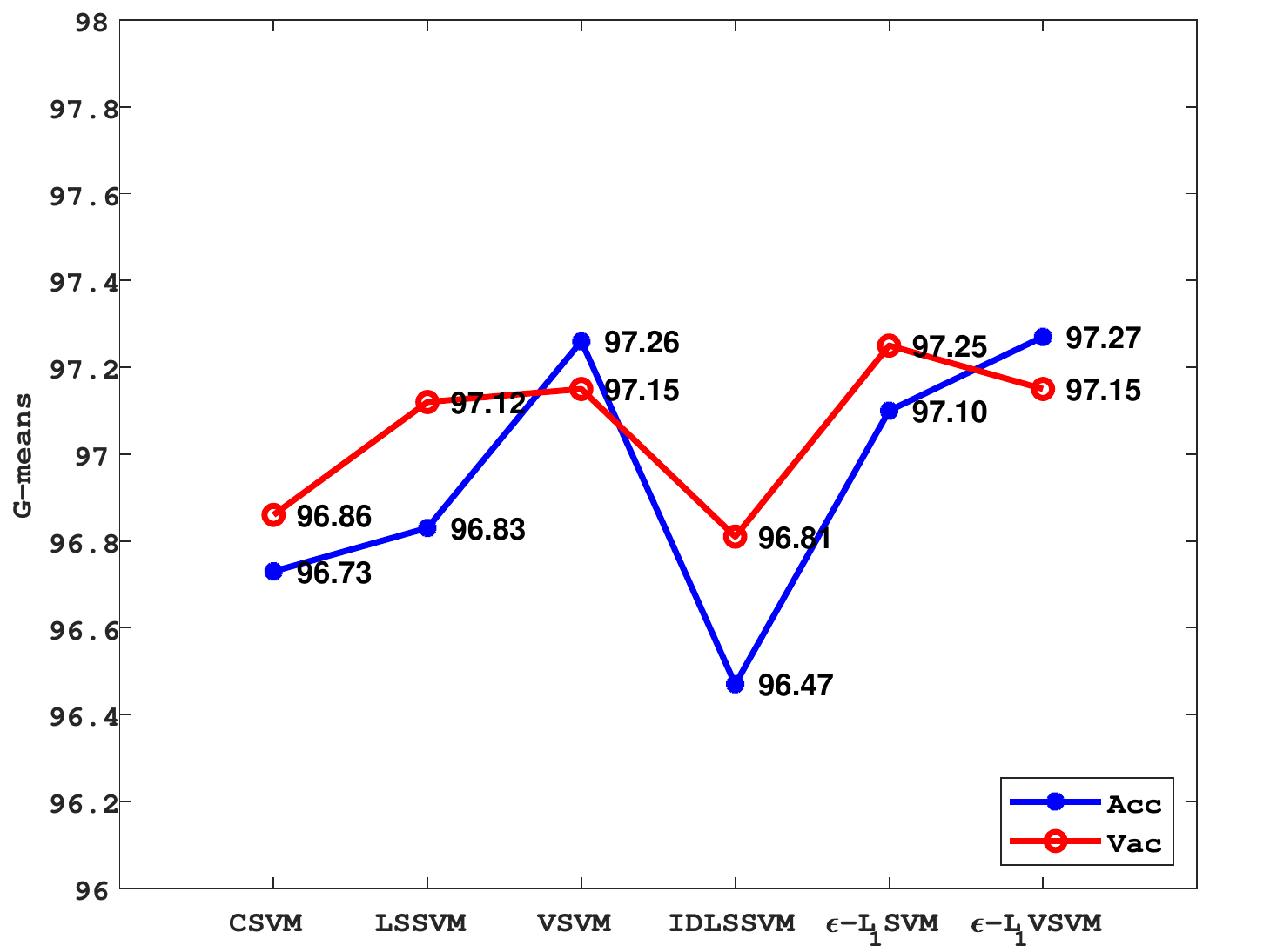}}
\caption{Nonlinear results based on Acc \eqref{Acc} and Vac \eqref{Vac} as evaluation indicators for six classifiers on nine small datasets.}
\label{vac-acc-rbf}
\end{figure*}

Furthermore, in the synthetic datasets experiments, we show the effectiveness of Vac as an evaluation metric. Now, we conduct experiments using Vac on real data in the following. Considering the space of the paper, we select nine smallest datasets as examples for validation since the artificial data experiments reflects the advantage of $\varepsilon$-$L_{1}$VSVM and Vac on small data.
Figure \ref{vac-acc} and Figure \ref{vac-acc-rbf} show the classification results for each model on the selected nine small datasets, respectively, where the red line is a line graph of the results with Vac as the evaluation indicator and the blue is for Acc.

\textbf{Linear results and non-linear results of Vac as an indicator:} From the numerical results labeled in the figure, it is clear that our proposed model is still more advantageous compared to the other models. Whether Acc or Vac is used as the indicator for selecting the parameters (red or blue line), the results of our model are in most cases the peak of the line graph. For the red line, for both linear and nonlinear results, our model is the best in 6 out of 9.

From another perspective, for these nine small datasets, the red line is above the blue one in most cases, which indicates that Vac, as a more rigorous evaluation metric of the same type as Acc but with additional consideration of distribution, can improve the performance of the trained models. Further, for models like CSVM, LSSVM, IDLSSVM and $\varepsilon$-$L_{1}$SVM, the effects shown in Figure \ref{vac-acc} and Figure \ref{vac-acc-rbf} are more significant. However, for $\varepsilon$-$L_{1}$VSVM and VSVM, the performance improvement is not obvious, and even decreases on some datasets.

The following conclusions can be drawn for the results of the above synthetic datasets and benchmark datasets:
\begin{itemize}
\item[i)] Both for synthetic datasets and benchmark datasets, our model shows a superiority on small datasets. For example, $\varepsilon$-$L_1$VSVM  significantly outperforms other classifiers for artificial data of N=100, while the advantage is relatively smaller at N=300. This is because when the sample size is not sufficient, our model can capture the distribution structure of the data and obtain more location information which cannot be directly extracted by classical classifiers, thus classifying more accurately.
\item[ii)] Different versions of v-vectors have a large impact on the performance of the model and evaluation metrics. So we need to choose different v-vector for different data distributions. Two kinds of v-vector have been proposed, and their expressions are \eqref{v1} and \eqref{v2}. When the data distribution is known to be Gaussian, \eqref{v1} is more recommended because its $\mu(x)$ is estimated based on the Gaussian distribution, but when the data distribution is unknown, \eqref{v2} may be better. The follow-up on the calculation of v-vector needs to be studied in depth.
\item[iii)] For models that do not contain data distribution information in the objective function, such as CSVM, LSSVM, IDLSSVM, $\varepsilon$-$L_1$SVM, they are significantly improved the performance of their models when Vac is used as an evaluation metric. Therefore, using Vac as an evaluation metric is another way to add a priori information such as data distribution to the model. For both VSVM and $\varepsilon$-$L_1$VSVM models that inherently have distribution information, Vac gives a relatively much smaller improvement in model performance.
\end{itemize}

\section{Conclusion}
In this paper, we study the expected risk estimation principle and corresponding empirical risk estimation by estimating the conditional probability function of the data via Fredholm equation. The corresponding classification models and the classification evaluation indicators under this paradigm are different from the traditional one.  Experimental results show the effectiveness of the proposed $\varepsilon$-$L_{1}$VSVM and the proposed indicator on validity and interpretability of classification, especially in case of insufficient training data.
It is worth mentioning that the risk measurement space in proposed paradigm is different from the traditional one, so further study the more appropriate loss measurement is our main research directions in the future. In addition,
extend the paradigm by estimating the conditional probability function of the data via Fredholm equation to different statistical inference problems are also promising.

\section*{Acknowledgment}
This work is supported by the Natural Science Foundation
of Hainan Province (No.120RC449), the National Natural
Science Foundation of China (Nos. 12271131, 61866010, 11871183), the
Scientific Research Foundation of Hainan University (No:
kyqd(sk)1804).

\section*{Reference}

\bibliographystyle{elsarticle-num}
\biboptions{numbers,sort&compress}

\bibliography{bibfile}

\end{document}